\documentclass{article} 
\usepackage{iclr2027_conference,times}

\usepackage{amsmath,amsfonts,bm}

\def\eqref#1{equation~\ref{#1}}

\def\1{\bm{1}}

\DeclareMathAlphabet{\mathsfit}{\encodingdefault}{\sfdefault}{m}{sl}
\SetMathAlphabet{\mathsfit}{bold}{\encodingdefault}{\sfdefault}{bx}{n}

\usepackage{makecell}
\usepackage{multirow}
\usepackage{hyperref}
\usepackage{url}
\usepackage{algorithm}
\usepackage{algorithmic}
\usepackage{kotex}
\usepackage{float}
\usepackage{amsmath}
\usepackage{amssymb}
\usepackage{amsthm}

\newtheorem{lemma}{Lemma}[section]
\newtheorem{proposition}{Proposition}[section]
\newtheorem{remark}{Remark}[section]
\newtheorem{corollary}{Corollary}[section]
\usepackage{wrapfig}
\usepackage{booktabs}

\usepackage{etoc}
\usepackage{array}
\usepackage{tabularx}

\usepackage{xcolor}
\definecolor{PredBlue}{RGB}{55,95,170}
\definecolor{ShapPurple}{RGB}{132,72,160}
\definecolor{GapAmber}{RGB}{226,164,55}
\definecolor{InkGray}{RGB}{80,80,80}

\usepackage{tikz-cd}
\tikzset{
  kps node/.style={
    draw=black!70,
    fill=black!3,
    line width=0.4pt,
    rounded corners=2pt,
    minimum width=3.45cm,
    minimum height=1.05cm,
    inner xsep=4pt,
    inner ysep=4pt
  },
  kps tag/.style={
    solid,
    draw=black!28,
    fill=black!3,
    rounded corners=1.7pt,
    inner xsep=3.2pt,
    inner ysep=1.4pt,
    line width=0.35pt,
    font=\scriptsize,
    text=black!80
  },
  kpscolor tag/.style={
    solid,
    draw=PredBlue!65,
    fill=PredBlue!6,
    rounded corners=1.7pt,
    inner xsep=3.2pt,
    inner ysep=1.4pt,
    line width=0.35pt,
    font=\scriptsize,
    text=black!80
  },
  kpscolor2 tag/.style={
    solid,
    draw=GapAmber!65,
    fill=GapAmber!6,
    rounded corners=1.7pt,
    inner xsep=3.2pt,
    inner ysep=1.4pt,
    line width=0.35pt,
    font=\scriptsize,
    text=black!80
  },
  kps empirical tag/.style={
    densely dashed,
    draw=PredBlue!65,
    fill=PredBlue!6,
    rounded corners=1.7pt,
    inner xsep=3.2pt,
    inner ysep=1.4pt,
    line width=0.35pt,
    font=\scriptsize,
    text=black!80
  }
}

\newcommand{\KPSNode}[2]{%
  \begin{gathered}
    \textstyle #1\\[-0.25ex]
    \text{\footnotesize #2}
  \end{gathered}%
}

\usetikzlibrary{
  arrows.meta,
  decorations.pathreplacing
}

\usepackage{pgfplots}
\pgfplotsset{compat=1.18}

\title{On the low rank theory of Data Shapley: a kernel perspective}

\author{Seungcheol Shin\thanks{Equal contribution. Order determined by a coin toss.} \\
Department of Mathematical Sciences\\
Seoul National University\\
Seoul, South Korea \\
\texttt{shinseungcheol@snu.ac.kr} \\
\And
Donghwan Rho${}^{*}$ \\
Samsung Electronics \\
Suwon, South Korea \\
\texttt{donghwan.rho@samsung.com} \\
\AND
Myungjoo Kang \\
  Department of Mathematical Sciences and Research Institute of Mathematics \\
  Interdisciplinary Program in Artificial Intelligence \\
  Interdisciplinary Program in Computational Sciences and Technology \\
  Seoul National University \\
  Seoul, South Korea  \\
\texttt{mkang@snu.ac.kr}
}

\iclrfinalcopy 
\begin{document}

\maketitle

\begin{abstract}
Data Shapley offers a principled framework for data valuation, but remains costly to compute at scale. Although FreeShap avoids repeated model fine-tuning through kernel regression, evaluating many sampled coalitions remains computationally demanding.
Low-rank kernel approximation offers a natural acceleration strategy, but its effect on individual Data Shapley values is poorly understood. We investigate how kernel approximation relates to prediction and Data Shapley. Our analysis shows that kernel approximation error controls errors in individual Data Shapley values, whereas predictive agreement alone does not guarantee Shapley agreement. Together with our experiments, these results indicate that eigendecomposition is more reliable than Nystr\"om approximation for Data Shapley. Guided by this analysis, we propose LRFShap, which accelerates coalition evaluation using shared low-dimensional kernel features. Across three downstream tasks, LRFShap achieves performance comparable to FreeShap at suitable ranks, with speedups of up to $22\times$ on data selection. Our code is available at \url{https://github.com/Donghwan-Rho/lrfshap}.
\end{abstract}

\etocdepthtag.toc{main}

\section{Introduction}
The quality of training data strongly affects the performance of neural networks~\citep{abdin2024phi,penedo2024fineweb}; for instance, repeated training on recursively generated synthetic data can degrade model performance~\citep{shumailov2024ai,alemohammad2024self}. Data valuation is therefore increasingly important, with applications such as data selection and data markets. Data Shapley~\citep{ghorbani2019data,jia2019towards} provides a principled approach to data valuation based on the Shapley value~\citep{shapley1953value}, a classical solution concept from cooperative game theory. It assigns each training sample a value based on its average marginal contribution to a given utility over coalitions of training data.

However, computing Data Shapley for large neural networks is prohibitively expensive. Although truncated Monte Carlo sampling~\citep{ghorbani2019data} mitigates the combinatorial cost, repeatedly training or fine-tuning the model for sampled coalitions remains the primary bottleneck. Methods based on the neural tangent kernel (NTK)~\citep{jacot2018neural} avoid this repeated training~\citep{wu2022davinz,wang2024helpful}. In particular, FreeShap~\citep{wang2024helpful} replaces coalition-wise fine-tuning with kernel regression based on the empirical NTK (eNTK), building on the connection between neural-network training and fixed-kernel learning in the kernel regime~\citep{arora2019exact,lee2019wide,malladi2023kernel}. Nevertheless, FreeShap still requires solving kernel systems for many sampled coalitions. Low-rank kernel approximation is a natural remedy, but how such approximations affect individual Data Shapley values remains poorly understood. We therefore ask:
\begin{center}
\emph{Does preserving predictions under low-rank kernel approximation also preserve \\ individual Data Shapley values?}    
\end{center}

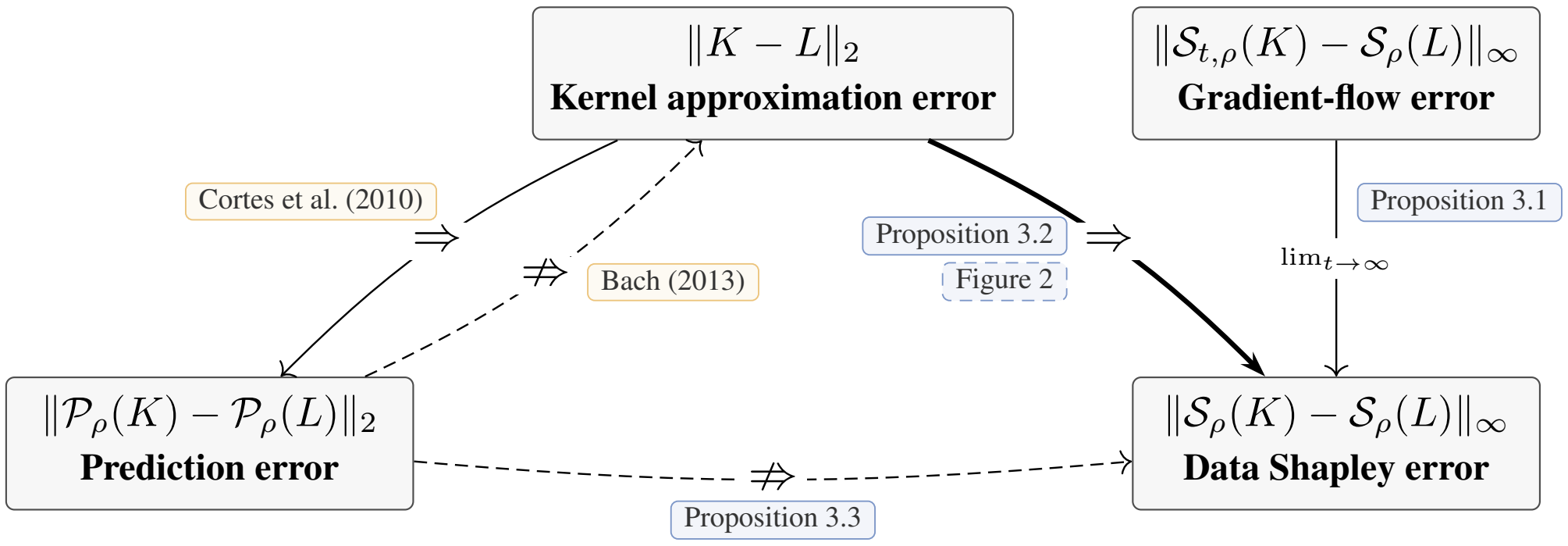
\begin{figure}[htbp]
\centering
\begin{tikzcd}[row sep=2cm,column sep=1cm,every arrow/.append style={line width=0.45pt}]
&
|[kps node]| {
  \KPSNode
    {\lVert K-L\rVert_{2}}
    {\textbf{Kernel approximation error}}
}
\arrow[dl, bend right=10, "\Rightarrow"{description,font=\large},
       "{\text{\scriptsize\citet{cortes2010impact}}}"'{pos=0.5,yshift=+5pt,kpscolor2 tag}]
\arrow[dr, line width=1.2pt,
  -{Stealth[length=2.2mm,width=1.45mm]},bend left=10, "\Rightarrow"{description,font=\large},
       "{\text{\scriptsize Proposition~\ref{prop:low_rank_stab}}}"'{pos=0.5,yshift=+6pt, xshift=-10pt, kpscolor tag}, "{\text{\scriptsize Figure~\ref{fig:rel_error}}}"'{pos=0.5,yshift=-5pt, xshift=-10pt, kps empirical tag}]
&
|[kps node]| {
  \KPSNode
    {\lVert
      \mathcal S_{t,\rho}(K)
      -
      \mathcal S_{\rho}(L)
     \rVert_{\infty}}
    {\textbf{Gradient-flow error}}
}
\arrow[
  d, "\lim_{t\rightarrow \infty}"{description,font=\tiny},
  "{\text{\scriptsize
    Proposition~\ref{prop:proposition_gf_shap_stab}}}"
    {right,kpscolor tag,pos=0.5,xshift=5pt,yshift=13.5pt}
]
\\
|[kps node]| {
  \KPSNode
    {\lVert\mathcal P_{\rho}(K)-\mathcal P_{\rho}(L)\rVert_{2}}
    {\textbf{Prediction error}}
}
\arrow[rr, dashed, bend right = 5,  "\nRightarrow"{description,font=\large},
       "{\text{\scriptsize Proposition~\ref{prop:prediction-shapley-separation}}}"'{pos=0.5,yshift=-5pt,kpscolor tag}]
\arrow[ur, dashed,  bend right=10,  "\nRightarrow"{description,font=\large},
       "{\text{\scriptsize\citet{bach2013sharp}}}"'{pos=0.5,yshift=+3pt, xshift=+10pt, kpscolor2 tag}]
&
&
|[kps node]| {
  \KPSNode
    {\lVert\mathcal S_{\rho}(K)-\mathcal S_{\rho}(L)\rVert_{\infty}}
    {\textbf{Data Shapley error}}
}
\end{tikzcd}
\caption{
\textbf{KPS (kernel--prediction--Shapley) diagram.} Here, \(K\) denotes a kernel
matrix and \(L\) its approximation, while \(\mathcal P\) and
\(\mathcal S\) denote the prediction and Data Shapley maps, respectively.
The symbol \(\Rightarrow\) indicates a theoretical implication.
Previously reported results are highlighted in orange, whereas our results
are highlighted in blue. See Section~\ref{sec:KPS_diagram} for details.}
\label{fig:KPS_Diagram}
\end{figure}

 Figure~\ref{fig:KPS_Diagram} summarizes our answer. Classical results show that a small kernel approximation error controls the resulting prediction error~\citep{cortes2010impact}. Similarly, Proposition~\ref{prop:low_rank_stab} shows that kernel approximation error also controls errors in individual Data Shapley values. For prediction alone, \citet{bach2013sharp} showed that the Nystr\"om method can preserve predictive performance without closely approximating the full kernel matrix. This prediction-oriented conclusion does not extend to Data Shapley. Proposition~\ref{prop:prediction-shapley-separation} establishes that predictive agreement alone does not guarantee agreement in Data Shapley values. Corollary~\ref{cor:nystrom_limitation} makes this separation concrete for uniform Nystr\"om sampling by exhibiting instances in which, with high probability, the full-coalition prediction is preserved exactly, yet individual Data Shapley estimates incur constant-order relative errors.

Guided by this analysis, we propose \underline{L}ow-\underline{R}ank \underline{F}ine-tuning-free \underline{Shap}ley value (\textbf{LRFShap}), which accelerates FreeShap through low-rank kernel approximation. Importantly, the same pattern emerges in a practical pretrained-model fine-tuning setting, where Data Shapley is estimated using the eNTK: the Nystr\"om method can yield unstable estimates, whereas eigendecomposition closely matches FreeShap while achieving a substantial speedup.

Our contributions are as follows:

\begin{itemize}
    \item We establish a stability bound for individual Data Shapley values
    under kernel approximation and show that predictive agreement alone
    need not preserve these values, as summarized in the KPS diagram.
    \item We provide theoretical and empirical evidence that eigendecomposition is better suited to Data Shapley approximation than the Nystr\"om method.
    \item LRFShap achieves performance comparable to FreeShap across three downstream tasks at suitable ranks, with speedups of up to $22\times$ in data selection.
\end{itemize}

\section{Preliminaries}\label{sec:preliminaries}

\paragraph{Notation.}
Let \(N\), \(T\), and \(C\) denote the training-sample, test-sample,
and class-index sets, respectively. Let \(d\) and \(P\) denote the
input and parameter dimensions, respectively. Let
\(D_N:=\{(x_i,y_i)\}_{i\in N}\) and
\(D_T:=\{(x_i,y_i)\}_{i\in T}\) be the training and test datasets,
where \(x_i\in\mathbb R^d\) and
\(y_i\in\{0,1\}^{|C|}\) is the corresponding one-hot encoded class
label. For each \(S\subseteq N\), define
\(D_S:=\{(x_i,y_i)\}_{i\in S}\). 
Given an arbitrary ordering
\(S=\{i_1,\ldots,i_{|S|}\}\), set
$X_S:=$ $[x_{i_1},\ldots,x_{i_{|S|}}]^\top
\in\mathbb R^{|S|\times d}$ and
\(Y_S:=[y_{i_1},\ldots,y_{i_{|S|}}]^\top
\in\{0,1\}^{|S|\times |C|}\).
The choice of ordering is immaterial, as it only determines the row
ordering of these matrices. The test matrices \(X_T\) and \(Y_T\)
are defined analogously.
Unless otherwise specified,
\(k:\mathbb R^d\times\mathbb R^d\to\mathbb R\) denotes an arbitrary
fixed scalar-valued positive-semidefinite kernel, and
\(\mathcal H_k\) denotes its associated reproducing kernel Hilbert
space (RKHS). For index sets \(A,B\subseteq N\cup T\), we write
\(
K_{AB}
:=
\bigl[k(x_i,x_j)\bigr]_{i\in A,j\in B}
\in\mathbb R^{|A|\times |B|}.
\)
When a particular kernel is used, its choice is stated explicitly.

\paragraph{Data Shapley.}
Data Shapley formulates data valuation as a cooperative game in
which each training sample is treated as a player~\citep{ghorbani2019data,jia2019towards}.
In the supervised learning setting considered here, for each coalition
$S\subseteq N$, let $f_S$ denote the model trained on the corresponding
training subset $D_S$. A utility function $U:2^N\to\mathbb{R}$ assigns
to each coalition $S$ a score that measures the performance of $f_S$
on the test dataset $D_T$.
Common utility functions include test accuracy and negative test error,
which we use for empirical evaluation and theoretical analysis,
respectively. The Data Shapley value of a training sample \(i\in N\) is defined by the
following equivalent coalition-based and permutation-based formulas:
\begin{align}
\label{eq:shapley_subset}
\phi_i(U)
:=
\sum_{S\subseteq N\setminus\{i\}}
w_{|N|,|S|}
\left(
U(S\cup\{i\})-U(S)
\right)
=
\frac{1}{|N|!}
\sum_{\pi\in\Pi(N)}
\left(
U(P_i^\pi\cup\{i\})-U(P_i^\pi)
\right),
\end{align}
Here,
\(w_{|N|,|S|}:=
\frac{|S|!\,\bigl(|N|-|S|-1\bigr)!}{|N|!}\),
\(\Pi(N)\) is the set of all permutations of \(N\), and \(P_i^\pi\) is
the set of samples that appear before \(i\) in permutation \(\pi\).
Thus, \(\phi_i(U)\) quantifies the contribution of training sample \(i\)
to the predictive performance measured by \(U\).

Computing Data Shapley  has two main computational bottlenecks:
\begin{enumerate}
    \item \textbf{Combinatorial complexity:}
    computing the exact average in
    \eqref{eq:shapley_subset} requires evaluating all $|N|!$
    permutations of the training samples;
    
    \item \textbf{Model retraining cost:}
    evaluating a marginal contribution
    $U(S\cup\{i\})-U(S)$ generally requires training or retraining
    models on $D_S$ and $D_{S\cup\{i\}}$.
\end{enumerate}

The first bottleneck is commonly mitigated using Monte Carlo sampling
based on \eqref{eq:shapley_subset}, which yields an unbiased
estimator of the Shapley value~\citep{maleki2013bounding}.
In this work, we focus on the second bottleneck, which remains
particularly challenging for large neural networks.

\paragraph{Neural Tangent Kernels for Data Shapley.}

The Neural Tangent Kernel (NTK) provides a theoretical framework for
describing the training dynamics of infinitely wide neural networks
\citep{jacot2018neural}.
Under the kernel regime, neural network training can be characterized as
kernel regression with a fixed NTK \citep{arora2019exact,lee2019wide}.
For finite-width networks, the empirical NTK (eNTK) can be evaluated at
pretrained weights, and prior work shows that the resulting eNTK
regression can approximate prompt-based fine-tuning behavior in language
and vision models\citep{wei2022more,malladi2023kernel}.

Let
\(
F(x;\theta)
=
\bigl(f^1(x;\theta),\ldots,f^{|C|}(x;\theta)\bigr)
\in \mathbb{R}^{|C|}
\)
be a differentiable model parameterized by
\(\theta\in\mathbb{R}^P\), and let \(\theta_0\) denote its pretrained parameter vector.
Following \citet{wei2022more,mohamadi2023fast}, we replace the full multi-output eNTK
with the scalar eNTK associated with the first output logit; see
Appendix~\ref{appendix:NTK_derivation} for further details.
For any pair of inputs $x_1,x_2\in\mathbb{R}^d$,
\begin{equation}
\label{eq:scalar_entk}
\hat{\Theta}_{\theta_0}(x_1,x_2)
:=
\left[
\frac{\partial f^{1}(x_1;\theta_0)}{\partial \theta_0}
\right]
\left[
\frac{\partial f^{1}(x_2;\theta_0)}{\partial \theta_0}
\right]^{\top}.
\end{equation}
For eNTK regression, we specialize the preceding kernel notation to
\(k=\hat{\Theta}_{\theta_0}\).
Given a fixed ridge parameter \(\gamma>0\), for every nonempty coalition
\(S\subseteq N\), we define 
\begin{equation}
\label{eq:single_logit_entk}
f^{\mathrm{eNTK}}_S(X_T)
=
K_{TS}
\bigl(K_{SS}+\gamma I_{|S|}\bigr)^{-1}
Y_S.
\end{equation}
\citet{wang2024helpful} propose Fine-Tuning-Free Shapley Value (FreeShap) to effectively address the model-retraining cost of Data Shapley. For large neural networks, fine-tuning the entire model for each sampled coalition during Monte Carlo estimation is computationally prohibitive. FreeShap replaces these repeated model updates with eNTK regression, so that, once the eNTK matrices are precomputed, coalition utilities can be evaluated without further parameter updates. This enables efficient computation of Data Shapley even for large-scale models.

\paragraph{Best rank-\(r\) and Nystr\"om approximations.}

Let
\(K_{NN}\in\mathbb R^{|N|\times |N|}\)
denote the training kernel matrix. Since \(K_{NN}\) is symmetric
positive semidefinite, write its eigendecomposition as
\(K_{NN}=U\Lambda U^\top\), where
\(\Lambda=\operatorname{diag}(\lambda_1,\ldots,\lambda_{|N|})\) and
\(\lambda_1\geq\cdots\geq\lambda_{|N|}\geq 0\).
Let \(U_r\) and \(\Lambda_r\) collect the leading \(r\) eigenvectors
and eigenvalues, respectively.
By the Eckart--Young theorem, truncation to these components yields a
best rank-\(r\) approximation in spectral norm, with approximation error
\(\lambda_{r+1}\).
We denote this approximation by
\begin{equation}
\label{eq:eig_approx}
K_{NN}^{\mathrm{eig}}
:=
U_r\Lambda_rU_r^\top.
\end{equation}

As an alternative, the Nystr\"om method constructs a low-rank approximation from selected kernel columns, offering potential computational savings over full dense eigendecomposition~\citep{williams2000using,drineas2005nystrom}.
Let \(M\subseteq N\) be a landmark set of size \(m\), and denote the
Moore--Penrose pseudoinverse of \(K_{MM}\) by \(K_{MM}^{\dagger}\).
The standard Nystr\"om approximation is the following matrix of rank
at most \(m\):
\begin{equation}
\label{eq:nys_approx}
K_{NN}^{\mathrm{nys}}
:=
K_{NM}K_{MM}^{\dagger}K_{MN}.
\end{equation}

\section{Low-Rank Kernel Approximation for Data Shapley}
\label{sec:low_rank_data_shapley}

\subsection{Kernel--Prediction--Shapley Analysis}
\label{sec:KPS_diagram}

Throughout this subsection, \(k\), \(\mathcal H_k\), and \(K_{AB}\) retain the generic meanings assigned in the notation above. For the theoretical analysis, we specialize to scalar-response kernel regression rather than multiclass classification. With a slight abuse of notation, \(y_i\) denotes a scalar response satisfying \(y_i\in\mathbb R\) and \(|y_i|\leq 1\) for every \(i\in N\cup T\), while \(Y_S\in\mathbb R^{|S|}\) and \(Y_T\in\mathbb R^{|T|}\) denote the corresponding response vectors. The multiclass convention introduced in Section~\ref{sec:preliminaries} is retained elsewhere. We further assume that \(\kappa:=\sup_{x\in\mathbb R^d}\sqrt{k(x,x)}<\infty\).

We write \(\operatorname{Sym}(|N|)^+\) for the open cone of real
symmetric positive-definite \(|N|\times|N|\) matrices, whose closure
\(\overline{\operatorname{Sym}(|N|)^+}\) is the cone of real symmetric
positive-semidefinite matrices.
We regard \(Y_N\), \(Y_T\), and \(K_{TN}\) as fixed, while \(G\), representing either the exact training kernel matrix or one of its approximations, varies over
\(\overline{\operatorname{Sym}(|N|)^+}\). For \(\rho>0\), \(t\geq0\), and every nonempty \(S\subseteq N\), set
\(A_{S,\rho}(G):=G_{SS}+\rho|S|I_{|S|}\)\footnote{
Here, \(\rho\) denotes a fixed RKHS regularization coefficient under the
averaged empirical loss, yielding the diagonal ridge \(\rho|S|\).
In our implementation, we instead use a coalition-independent ridge
parameter \(\gamma\).}.
Since
\(A_{S,\rho}(G)\succeq\rho|S|I_{|S|}\succ0\) for every
\(G\in\overline{\operatorname{Sym}(|N|)^+}\), define
$$
\mathcal P_{t,\rho}^{S}:
\overline{\operatorname{Sym}(|N|)^+}
\longrightarrow
\mathbb R^{|T|},
\qquad
\mathcal P_{t,\rho}^{S}(G)
:=
K_{TS}
\left(
I_{|S|}
-
e^{-\frac{2t}{|S|}A_{S,\rho}(G)}
\right)
A_{S,\rho}(G)^{-1}Y_S.
$$
For the empty coalition, set
\(\mathcal P_{t,\rho}^{\varnothing}(G):=0\). In particular, when \(G=K_{NN}\),
\(\mathcal P_{t,\rho}^{S}(K_{NN})\) is precisely the prediction at time \(t\) of the zero-initialized gradient-flow
solution to the regularized squared-loss problem in
\(\mathcal H_k\); see
Appendix~\ref{appendix:Solutions by kernel regression}.
We define the stationary prediction is $\mathcal P_\rho^S(G)
:=
\mathcal P_{\infty,\rho}^S(G)
:=
K_{TS}A_{S,\rho}(G)^{-1}Y_S$. 
For each
\(G\in\overline{\operatorname{Sym}(|N|)^+}\), define the utility
function
\[
U_{t,\rho}^G:2^N\to\mathbb R,
\qquad
U_{t,\rho}^G(S)
:=-|T|^{-1}
\lVert Y_T-\mathcal P_{t,\rho}^{S}(G)\rVert_2^2.
\]
Using the Shapley-value definition in
\eqref{eq:shapley_subset}, define
$$
\mathcal S_{t,\rho}:
\overline{\operatorname{Sym}(|N|)^+}
\longrightarrow
\mathbb R^{|N|},
\qquad
\mathcal S_{t,\rho}(G)
:=
\bigl(
\phi_i(U_{t,\rho}^G)
\bigr)_{i\in N}.
$$
Since the Shapley value is a finite linear combination of coalition
utilities, \(\mathcal S_{t,\rho}(G)\) also converges as
\(t\to\infty\). We therefore define the grand-coalition
prediction map and the Data Shapley map by
$$
\mathcal P_\rho(G)
:=
\mathcal P_\rho^N(G),
\qquad
\mathcal S_\rho(G)
:=
\mathcal S_{\infty,\rho}(G).
$$
We show that the Shapley values induced by finite-time kernel
gradient-flow predictors converge to their
infinite-time counterparts. Thus, within the kernel regime, the proposition
justifies replacing finite-time kernel training with its infinite-time
limit when computing Shapley values, providing theoretical support for
the stationary-predictor approximation employed by FreeShap.
\begin{proposition}[Gradient-flow stability]
\label{prop:proposition_gf_shap_stab}
Assume that \(\rho>0\). There exists a constant
\(C_{\kappa,\rho}>0\), depending only on \(\kappa\) and \(\rho\),
such that, for every \(t\geq0\),
\[
\boxed{
\left\|
\mathcal S_{t,\rho}(K_{NN})
-
\mathcal S_{\rho}(K_{NN})
\right\|_{\infty}
\leq
C_{\kappa,\rho}e^{-2\rho t}.
}
\]
\end{proposition}
Under the present assumptions, \(\rho>0\) provides the uniform
spectral gap underlying the dataset-independent exponential rate in
Proposition~\ref{prop:proposition_gf_shap_stab}. In the ridgeless case,
each fixed coalition still converges, but the bound contains
\(\lambda_{\min}^{+}(K_{SS})^{-1}\), which cannot be controlled
uniformly because \(\lambda_{\min}^{+}(K_{SS})\) may be arbitrarily
small. Hence, no dataset-independent Data Shapley stability bound
follows without a uniform positive lower bound; see
Appendices~\ref{appendix:Solutions by kernel regression}
and~\ref{proof:Proof_of_Proposition_1}.

We next control the effect of replacing the exact training
kernel matrix with a positive-semidefinite approximation.
Appendix~\ref{app:theory_gap} empirically examines the
effects of the differences between the setting of
Proposition~\ref{prop:low_rank_stab} and our implementation.

\begin{proposition}[Kernel-approximation stability]
\label{prop:low_rank_stab}
Assume that \(\rho>0\). Let \(K_{NN}\) be the exact training
kernel matrix and let \(L_{NN}\) be a symmetric
positive-semidefinite approximation.
Then there exists a constant
\(\widetilde C_{\kappa,\rho}>0\), depending only on
\(\kappa\) and \(\rho\), such that
\[
\boxed{
\left\|
\mathcal S_{\rho}(K_{NN})
-
\mathcal S_{\rho}(L_{NN})
\right\|_{\infty}
\leq
\widetilde C_{\kappa,\rho}
\frac{1+\log |N|}{|N|}
\lVert K_{NN}-L_{NN}\rVert_2.
}
\]
\end{proposition}


Proposition~\ref{prop:low_rank_stab} provides a common basis
for comparing the approximations in
Equations~\ref{eq:eig_approx} and~\ref{eq:nys_approx}.
For Nystr\"om approximation, suppose that landmark indices
are sampled independently and uniformly with replacement,
retaining repeated indices.
For a comparator rank \(r\leq m\), combining Theorem~1
of \citet{cortes2010impact} with the proposition yields,
with probability at least \(1-\delta\),
\[
\begin{aligned}
&\left\|
\mathcal S_\rho(K_{NN})
-
\mathcal S_\rho(K_{NN}^{\mathrm{nys}})
\right\|_\infty
\leq
\widetilde C_{\kappa,\rho}
\left[
\frac{1+\log|N|}{|N|}
\lambda_{r+1}(K_{NN})
+
\frac{\kappa^2(1+\log|N|)}{\sqrt m}
\left(2+\log\frac{1}{\delta}\right)
\right].
\end{aligned}
\]
With \(\kappa,\rho,\delta\) fixed, the additional sampling
term in the bound above prevents it from certifying
individual Shapley errors at the \(1/|N|\) reference
scale in the low-rank regime \(m<|N|\).
Appendix~\ref{proof: Proof of Proposition 2} discusses
the bounds obtained by applying
Proposition~\ref{prop:low_rank_stab} to Nystr\"om
approximation and eigen truncation.

However, this upper bound alone does not fully characterize
the actual Shapley errors.
For prediction, \citet{bach2013sharp} shows that, under
suitable conditions, Nystr\"om kernel ridge regression
can achieve prediction risk comparable to full-kernel
regression even when the kernel approximation error
\(\|K-L\|_2\) remains relatively large.
It is therefore natural to ask whether the same holds for Data Shapley,
that is, whether an approximation that preserves predictions also
preserves individual Shapley values.

These observations raise two questions: whether agreement
of grand-coalition predictions alone suffices to control
Data Shapley, and whether the limitation of the preceding
upper bound is accompanied by actual Shapley errors.
The next proposition addresses the first question.
For its statement, we restrict \(\mathcal P_\rho\) and
\(\mathcal S_\rho\) to the open cone
\(\operatorname{Sym}(|N|)^+\).
\par

\begin{proposition}[Prediction does not determine Data Shapley]
\label{prop:prediction-shapley-separation}
Let \(G_0\in\operatorname{Sym}(|N|)^+\), and suppose that
\(
\ker D\mathcal P_\rho(G_0)
\not\subseteq
\ker D\mathcal S_\rho(G_0).
\)
Then there do not exist a neighborhood \(\mathcal V\) of
\(G_0\) in \(\operatorname{Sym}(|N|)^+\) and a function
\(\omega:[0,\infty)\to[0,\infty)\) satisfying
\(\omega(0)=0\) such that, for all \(K,L\in\mathcal V\),
\[
\boxed{
\left\|
\mathcal S_\rho(K)
-
\mathcal S_\rho(L)
\right\|_2
\leq
\omega\left(
\left\|
\mathcal P_\rho(K)
-
\mathcal P_\rho(L)
\right\|_2
\right).
}
\]
\end{proposition}

The following corollary addresses the second question
for uniform Nystr\"om sampling by exhibiting actual
relative Shapley errors despite exact preservation
of the grand-coalition prediction.
See Appendix~\ref{proof: Proof of Proposition 3} for the proof of
Proposition~\ref{prop:prediction-shapley-separation},
and Appendix~\ref{proof: nystrom_limitation_cor} for the formal
statement and proof of Corollary~\ref{cor:nystrom_limitation}.

\begin{corollary}[Informal, limitation of uniform Nystr\"om sampling]
\label{cor:nystrom_limitation}
There exist instances for which, with probability
\(1-O(|N|^{-1/3})\), uniform Nystr\"om sampling preserves
the grand-coalition prediction exactly but incurs
constant-order relative Shapley errors on
\(\Theta(|N|^{1/3})\) training samples.
\end{corollary}
\par

\subsection{LRFShap: Low-Rank Data Shapley Computation}
\label{sec:acceleration_low_rank}
We now present LRFShap, a low-rank approximation framework for accelerating kernel-based Data Shapley estimation.
Let \(K_{NN}\) and \(K_{TN}\) denote the full train--train and test--train eNTK matrices, respectively. Let \(\mathcal A\) denote a
low-rank feature construction, and let \(\vartheta\) denote its
construction-specific configuration.
We write
$
\label{eq:lr_feature_construction}
(\Phi_{\mathrm{tr}},\Phi_{\mathrm{te}})
=
\mathcal A(K_{NN},K_{TN};\vartheta),
$
where 
\(\Phi_{\mathrm{tr}}\in\mathbb{R}^{{|N|}\times q}\),
\(\Phi_{\mathrm{te}}\in\mathbb{R}^{|T|\times q}\), and \(q\ll {|N|}\). Here, \(q\) denotes the resulting feature dimension, and both the
meaning of \(\vartheta\) and the value of \(q\) depend on the chosen
construction. The feature matrices are constructed so that
\(K_{NN}\approx
\Phi_{\mathrm{tr}}\Phi_{\mathrm{tr}}^\top\) and
\(K_{TN}\approx\Phi_{\mathrm{te}}\Phi_{\mathrm{tr}}^\top\). We instantiate
\(\mathcal A\) using eigen and Nystr\"om features below.

For a nonempty subset \(S\subseteq N\), let
\(\Phi_S\in\mathbb{R}^{|S|\times q}\) denote the rows of
\(\Phi_{\mathrm{tr}}\) indexed by \(S\). 
Replacing the kernel matrices
in Equation~\ref{eq:single_logit_entk} by their low-rank feature products and applying the push-through identity 
\(
\Phi_S^\top(\Phi_S\Phi_S^\top+\gamma I_{|S|})^{-1}
=
(\Phi_S^\top\Phi_S+\gamma I_q)^{-1}\Phi_S^\top
\) gives
\begin{equation}
\label{eq:lrntk_predictor}
\boxed{
f_S^{\mathrm{LRNTK}}(X_T)
=
\Phi_{\mathrm{te}}
\left(
\Phi_S^\top\Phi_S+\gamma I_q
\right)^{-1}
\Phi_S^\top Y_S.
}
\end{equation}
Thus, LRFShap replaces the \(|S|\times|S|\) kernel system used by
FreeShap with a \(q\times q\) feature-space system.
In the experiments, we evaluate each nonempty coalition using the
accuracy utility
\[
U_{\mathrm{acc}}(S)
:=
\frac{1}{|T|}
\sum_{j\in T}
\mathbf 1\!\left[
\operatorname*{arg\,max}_{c\in C}
\bigl[f_S^{\mathrm{LRNTK}}(X_T)\bigr]_{j,c}
=
\operatorname*{arg\,max}_{c\in C}[y_j]_c
\right],
\qquad
S\neq\emptyset,
\]
and set
\(
U_{\mathrm{acc}}(\emptyset):=0.
\)
The feature matrices
\((\Phi_{\mathrm{tr}},\Phi_{\mathrm{te}})\)
are constructed once and shared across all coalition evaluations.
We then estimate the resulting Shapley values using truncated Monte
Carlo (TMC), as in FreeShap. Algorithm~\ref{alg:lr_freeshap}
summarizes the resulting procedure.

\begin{algorithm}[!ht]
\small
\caption{LRFShap}
\label{alg:lr_freeshap}
\begin{algorithmic}[1]
\REQUIRE
Training set \(D_N\),
test set \(D_T\),
number of MC iterations \(B\),
TMC tolerance \(\tau_{\mathrm{TMC}}>0\),
precomputed eNTK matrices \(K_{NN}\) and \(K_{TN}\),
low-rank feature construction \(\mathcal A\),
construction-specific configuration \(\vartheta\),
regularization \(\gamma>0\)

\ENSURE
Shapley value estimates
\(\phi_1,\ldots,\phi_{|N|}\)

\STATE Initialize \(\phi_i\gets 0\) for every \(i\in N\)

\STATE Construct
\(
(\Phi_{\mathrm{tr}},\Phi_{\mathrm{te}})
\gets
\mathcal A(K_{NN},K_{TN};\vartheta)
\)

\STATE Evaluate \(U_{\mathrm{acc}}(N)\) for the TMC truncation
criterion

\FOR{\(b=1\) to \(B\)}
    \STATE Sample a uniformly random permutation \(\pi^b\) of \(N\)
    \STATE Set
    \(u_0^b\gets U_{\mathrm{acc}}(\emptyset)=0\)

    \FOR{\(p=1\) to \(|N|\)}
        \STATE Set
        \(
        S^{(b,p)}
        \gets
        \{\pi^b[1],\ldots,\pi^b[p]\}
        \)

        \IF{
        \(\left|
        U_{\mathrm{acc}}(N)-u_{p-1}^b
        \right|
        <\tau_{\mathrm{TMC}}\)
        }
            \STATE Set
            \(u_p^b\gets u_{p-1}^b\)

        \ELSE
            \STATE Set
            \(
            \Phi_{S^{(b,p)}}
            \gets
            \Phi_{\mathrm{tr}}[S^{(b,p)},:]
            \)

            \STATE Compute
            \(f_{S^{(b,p)}}^{\mathrm{LRNTK}}(X_T)\)
            using Equation~\ref{eq:lrntk_predictor}

            \STATE Evaluate
            \(
            u_p^b
            \gets
            U_{\mathrm{acc}}(S^{(b,p)})
            \)
        \ENDIF

        \STATE Update
        \(
        \phi_{\pi^b[p]}
        \gets
        \frac{b-1}{b}\phi_{\pi^b[p]}
        +
        \frac{1}{b}
        \left(
        u_p^b-u_{p-1}^b
        \right)
        \)
    \ENDFOR
\ENDFOR
\end{algorithmic}
\end{algorithm}

\paragraph{Eigen and Nystr\"om features.}
We consider two specific choices of the low-rank feature construction
\(\mathcal A\). For eigen features, consider the best rank-\(r\)
approximation in Equation~\ref{eq:eig_approx}, and assume that the
retained eigenvalues are strictly positive. Taking \(\vartheta=r\)
gives \(q=r\), and we define
\begin{equation}
\label{eq:eig_features}
\mathcal A^{\mathrm{eig}}(K_{NN},K_{TN};r)
:=
\left(
U_r\Lambda_r^{1/2},
\,
K_{TN}U_r\Lambda_r^{-1/2}
\right).
\end{equation}
For large kernel matrices, the leading \(r\) eigenvalues can be
approximated using randomized subspace iteration
\citep{halko2011finding}, thereby avoiding a full dense
eigendecomposition.

For Nystr\"om features, consider the approximation in
Equation~\ref{eq:nys_approx}. Write
\(K_{MM}=V_q\Sigma_qV_q^\top\)
for the eigendecomposition of the landmark kernel matrix,
where \(q=\operatorname{rank}(K_{MM})\leq m\) and
\(\Sigma_q\succ 0\). Taking \(\vartheta=M\), we define
\begin{equation}
\label{eq:nys_features}
\mathcal A^{\mathrm{nys}}(K_{NN},K_{TN};M)
:=
\left(
K_{NM}V_q\Sigma_q^{-1/2},
\,
K_{TM}V_q\Sigma_q^{-1/2}
\right).
\end{equation}

\paragraph{Computational complexity.}
For a prefix \(S^{(b,p)}\) of size \(p\), forming
\(\Phi_{S^{(b,p)}}^{\top}\Phi_{S^{(b,p)}}\) requires
\(\mathcal O(pq^2)\) operations, while factoring and solving the
resulting \(q\times q\) system costs \(\mathcal O(q^3)\). Hence, the
coalition-solve cost per prefix is \(\mathcal O(pq^2+q^3)\), and
summing over all prefixes gives
\(\mathcal O(|N|^2q^2+|N|q^3)\) per permutation. Over \(B\) independent
Monte Carlo permutations, the coalition-solve complexity is therefore
\(\mathcal O\!\left(B(|N|^2q^2+|N|q^3)\right)\). This bound excludes the
one-time feature construction and the cost of test prediction and
utility evaluation. Under the same direct-factorization convention,
FreeShap requires \(\mathcal O(B|N|^4)\) operations. Therefore, when
\(q\ll |N|\), LRFShap substantially reduces the dominant
coalition-solve cost. A complete complexity analysis is provided in
Appendix~\ref{app:complexity}.

\section{Experimental results}

\paragraph{Experimental settings}
We present our experiments in a Q\&A format, first examining which decomposition is more suitable for approximating Data Shapley and then evaluating LRFShap on three downstream tasks: data selection, data removal, and wrong-label detection.
We use FreeShap~\citep{wang2024helpful} as the baseline throughout.
Our experiments use BERT and Llama-2 on SST-2, MR, MNLI, QQP, RTE, MRPC, and AG News, and ResNet-18 on CIFAR-10.
Both methods use the same TMC estimator~\citep{ghorbani2019data} and first-logit eNTK approximation, with $\gamma=10^{-2}$.
All results are averaged over three random seeds.
Model identifiers, hyperparameters, hardware specifications, and dataset details are provided in Appendix~\ref{app:exp_details}.

\subsection{Results and Q\&A}\label{sec:results_qna}

\underline{\textbf{Q. Eigen versus Nystr\"om decompositions:
which should we choose?}}

\textbf{A. In our experiments, eigendecomposition provides
more consistent approximations of Data Shapley values.}

\begin{figure}[!ht]
      \centering      \includegraphics[width=0.8\linewidth]{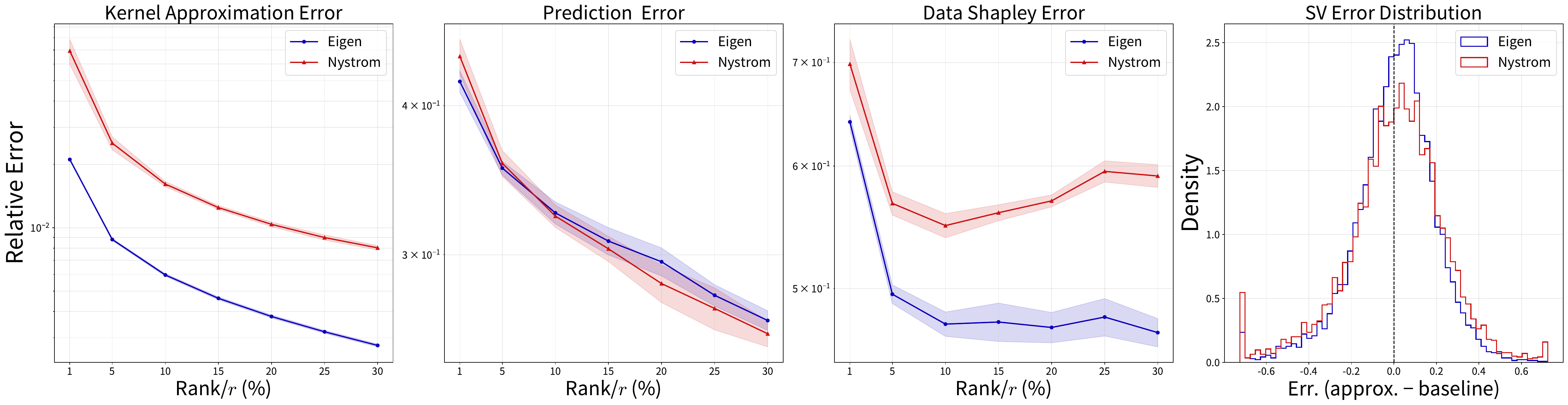}
      \caption{
      Eigen versus Nystr\"om for BERT on MR at $n=2000$: kernel approximation, prediction, and Data Shapley errors across ranks, data Shapley error distributions at rank 600 (30\% of $n$). The prediction error can be low even when the kernel approximation error is not, and a low prediction error does not imply a low Data Shapley error; the eigen decomposition attains higher fidelity. In the error distribution graph, errors are clipped by $[-1, 1]$.}
      \label{fig:rel_error}
\end{figure}

We compare eigendecomposition with uniform Nystr\"om approximation
through the KPS diagram.
Figure~\ref{fig:rel_error} reports kernel approximation, prediction, and
Data Shapley errors for these two methods, together with the
distributions of individual Data Shapley errors.
For the error curves, we use relative Frobenius errors for the
kernel and prediction matrices and relative Euclidean errors
for the Data Shapley vectors, normalizing each discrepancy by
the norm of the corresponding full-kernel baseline.

We observe that both methods achieve similar prediction errors
despite substantial differences in their kernel approximation errors.
For eigendecomposition, improvements in kernel approximation
are accompanied by decreasing Data Shapley errors.
For uniform Nystr\"om approximation, however, the Data Shapley
error can plateau or increase at some ranks even as the kernel
approximation error decreases.
The histograms show that the individual Data Shapley errors
are more concentrated around zero for eigendecomposition.
Its higher Spearman and Pearson correlations further indicate closer agreement with the baseline, see Appendix~\ref{app:kps_error_fidelity}.
Besides the uniform Nystr\"om sampling used above, we also tested the leverage-score Nystr\"om method~\citep{alaoui2015fast} and found that the results were similar.
For all results of this paragraph on all datasets and models, see Appendix~\ref{app:kps_error_fidelity}.
Finally, as an ablation study, we show in Appendix~\ref{app:nystrom_landmark} that the Data Shapley error of the Nystr\"om method heavily depends on the choice of landmarks.
We therefore conclude that eigendecomposition is more suitable for approximating Data Shapley.

\underline{\textbf{Q. Does LRFShap achieve performance comparable to the baseline?}}

\textbf{A. Yes. We demonstrate this across three downstream tasks: data selection, data removal, and wrong-label detection.}

First, we explain how the downstream tasks are conducted.
Given $k\in\{1,\ldots,100\}$, let $\mathcal{T}(k)$ and $\mathcal{B}(k)$ be the index sets of the training data with the top and bottom $k$\% Shapley values, respectively, with $\mathcal{T}(0)=\emptyset$.
In data selection (removal), we compute 
the predictor in Equation~\ref{eq:single_logit_entk} from $D_{\mathcal{T}(k)}$ ($D_{N\setminus\mathcal{T}(k-1)}$) and report its accuracy on the evaluation set.
In wrong label detection, we randomly change the labels of 10\% of the training data, compute Data Shapley on this corrupted dataset, and measure the detection rate, the fraction of the wrongly labeled data contained in $D_{\mathcal{B}(k)}$.
We compute Data Shapley using the validation set of each dataset as the utility set.
For data selection and removal, the accuracy is evaluated on a separate held-out evaluation set, which is taken from the original training split and is disjoint from $D_N$.

\begin{figure}[!ht]
      \centering      \includegraphics[width=0.8\linewidth]{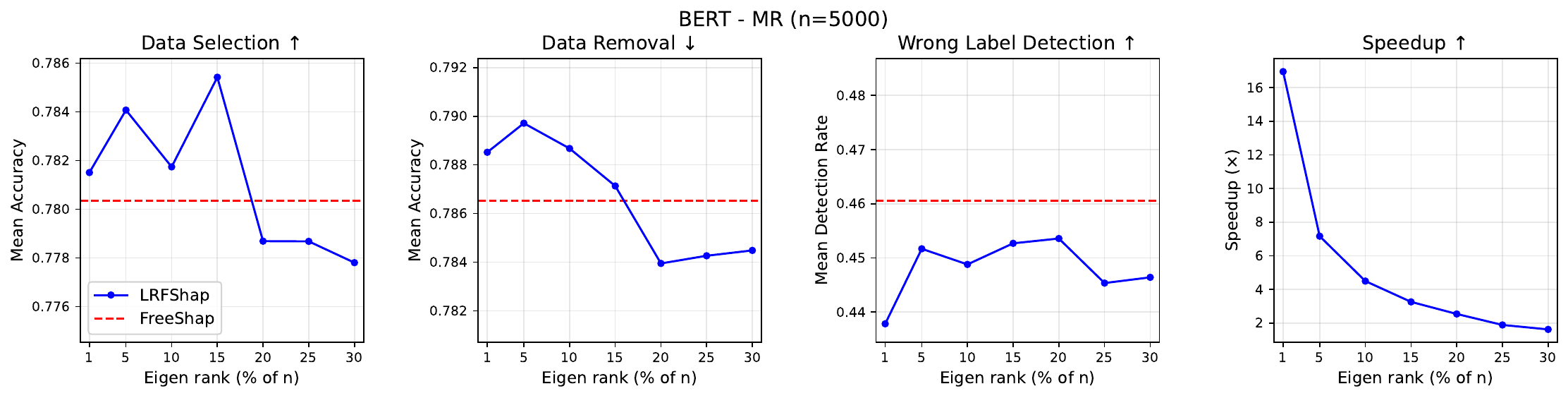}
      \caption{
      Mean accuracy, detection rate, and speedup of BERT on MR as the rank varies. Higher is better for data selection and wrong label detection; lower is better for data removal. LRFShap achieves performance comparable to FreeShap on all three tasks at $r=5\%$ while being up to 7.15$\times$ faster.}
      \label{fig:rank_curves_bert_mr_20}
  \end{figure}

Figure~\ref{fig:rank_curves_bert_mr_20} shows the results for BERT on MR.
Rather than reporting the performance at each individual percentage, we summarize each downstream task by the mean accuracy (mean detection rate for wrong label detection), defined as the performance averaged over $k\in\{1,2,\ldots,20\}$, where $k$ denotes the selection, removal, or inspection percentage of the corresponding task.
Since the most valuable or valueless data matter most in practice, we focus on selection, removal, and inspection percentages up to 20\%.
Tables~\ref{tab:mean_accuracy_selection_bert_20}--\ref{tab:mean_detection_rate_wld_resnet_20} in Appendix~\ref{sec:mean_accuracy_results} report the mean scores for all models and datasets and for figures representing raw accuracy, refer to Appendix~\ref{sec:more_experiments}.

We regard LRFShap as comparable to the baseline if its mean accuracy (mean detection rate) over three seeds is worse than that of the baseline by at most 0.01; that is, at most 0.01 lower in data selection and wrong label detection, where higher is better, and at most 0.01 higher in data removal, where lower is better.
Across all three tasks, LRFShap with a moderate eigen rank is comparable to or better than the baseline in most cases, which indicates that the top eigen-components capture representative features useful for identifying valuable and wrongly labeled data.
Increasing the rank does not consistently improve downstream performance in our experiments.
We guess that excessive information is not helpful for distinguishing such data.

\begin{table}[!ht]
\centering
\small
\setlength{\tabcolsep}{4pt}
\renewcommand{\arraystretch}{1.1}
\caption{The largest speedup LRFShap achieves in data selection while remaining comparable to the baseline. For Llama-2, the speedup is small on a few datasets, such as MR, MRPC, QQP, and SST-2, because the convergence ratio of the baseline is low there (see Figure~\ref{fig:early_stop_all} and the third question).}
\label{tab:speedups}
\begin{tabular}{l cccccccc}
\toprule
Model & \multicolumn{8}{c}{Rank (\%) / \textbf{Speedup} ($\times$)} \\
\cmidrule(lr){1-9}
\multirow{2}{*}{BERT/ResNet-18} & AG News & MNLI & MR & MRPC & QQP & RTE & SST-2 & CIFAR-10 \\
& 1 / \textbf{13.16} & 1 / \textbf{22.32} & 1 / \textbf{16.79} & 1 / \textbf{9.81} & 5 / \textbf{10.77} & 5 / \textbf{2.39} & 1 / \textbf{15.20} & 1 / \textbf{13.23} \\\cmidrule(lr){1-9}
\multirow{2}{*}{Llama-2} & AG News & MNLI & MR & MRPC & QQP & RTE & SST-2 & - \\
& 1 / \textbf{10.90} & 5 / \textbf{4.07} & 1 / \textbf{5.07} & 1 / \textbf{2.10} & 1 / \textbf{1.41} & 1 / \textbf{5.72} & 1 / \textbf{3.75} & - \\
\bottomrule
\end{tabular}
\end{table}

\underline{\textbf{Q. How much faster is LRFShap when the performance is comparable to the baseline?}}

\textbf{A. You can achieve up to 22.32$\times$ while maintaining comparable performance.}

In Section~\ref{sec:acceleration_low_rank}, we saw that LRFShap achieves an algorithmic speedup; however, the speedup is meaningful only if a comparable performance is retained.
Table~\ref{tab:speedups} reports the largest speedup that LRFShap can achieve in data selection while remaining comparable to the baseline, in the sense defined in the previous question.
LRFShap achieves up to a 22.32$\times$ speedup.
Although the same set of ranks is used for all datasets, the speedup varies across datasets; in particular, it is small on MR, MRPC, QQP, and SST-2 for Llama-2, because the TMC iteration of the baseline terminates early there (see the next question).
Tables~\ref{tab:runtime_eigen_vs_freeshap}--\ref{tab:runtime_eigen_vs_freeshap_resnet} in Appendix~\ref{computation_time} report the speedups for all cases, where we can see that, with too high a rank, LRFShap can even be slower than the baseline; this shows the importance of choosing an appropriate rank.

\underline{\textbf{Q. Is the reduced computational complexity the only source of the speedup?}}

\textbf{A. No. The number of coalition evaluations performed before TMC truncation also affects the runtime.}

Both methods use the same fixed number of Monte Carlo permutations and
the same TMC truncation rule.
Within each permutation, coalition evaluations stop once the prefix
utility is within the prescribed tolerance of the full-coalition utility,
and the remaining samples receive zero marginal contributions.
However, FreeShap evaluates utilities using the full kernel, whereas
eigen-based LRFShap uses only the retained eigen-components.
This changes both the prefix utilities and the full-coalition utility
used as the truncation target, so the two methods can truncate at
different coalition sizes.

\begin{wrapfigure}{!r}{0.45\linewidth}
  \centering
  \includegraphics[width=0.95\linewidth]{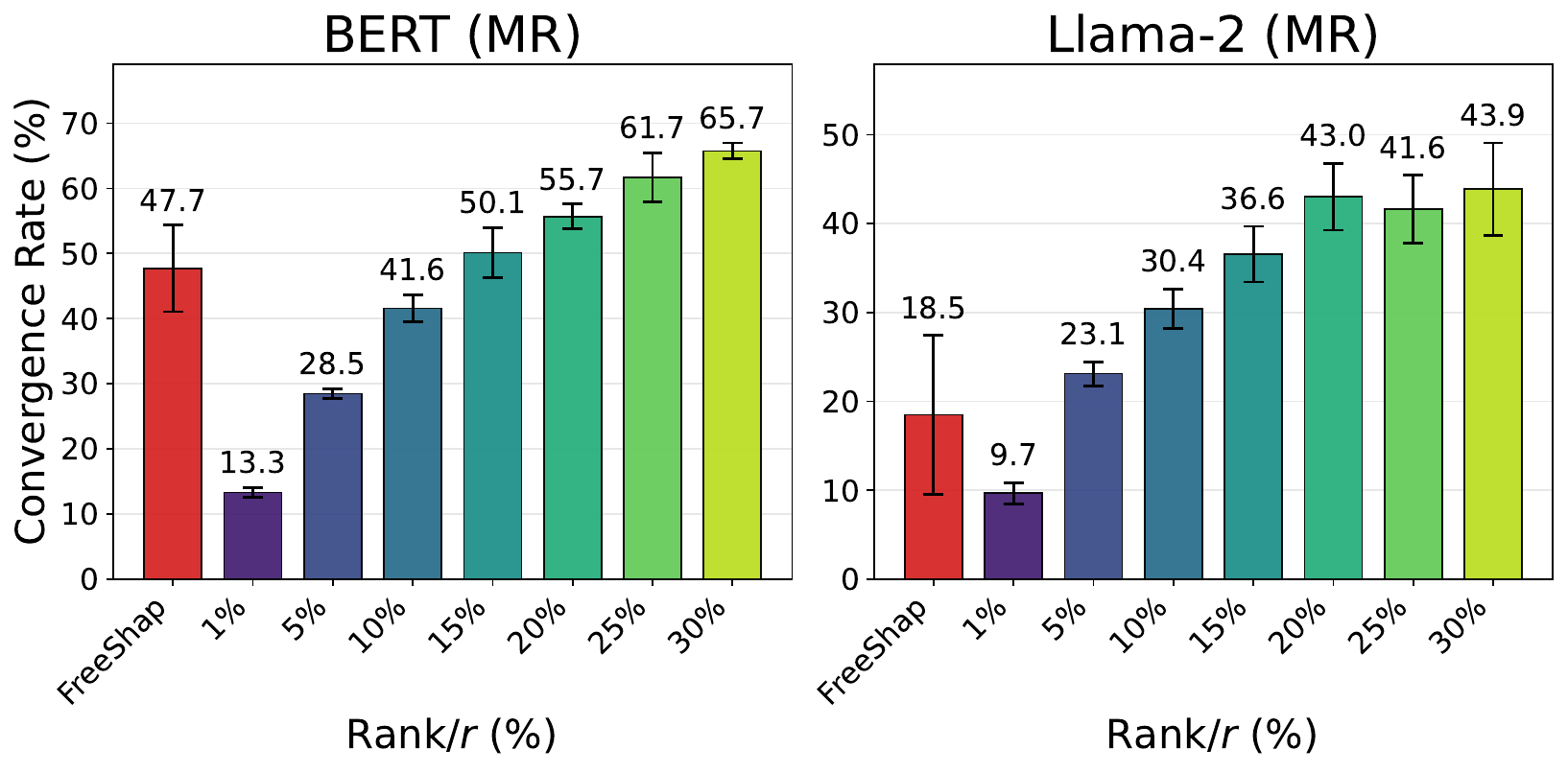}
  \caption{TMC convergence ratio for BERT and Llama-2 on MR.}
  \label{fig:convergence_bar}
\end{wrapfigure}

We quantify this behavior using the \textbf{TMC convergence ratio}
\(R_{\mathrm{TMC}} := \frac{100}{B|N|}\sum_{b=1}^{B}\ell_b\ (\%)\),
where \(B\) is the number of sampled permutations and
\(\ell_b\le|N|\) is the number of nonempty prefixes whose utilities are
evaluated in the \(b\)-th permutation before truncation.
For example, \(R_{\mathrm{TMC}}=10\%\) means that TMC truncates each
permutation after evaluating, on average, only 10\% of its \(|N|\)
prefixes; the number of permutations \(B\) is the same for all methods.

Figure~\ref{fig:convergence_bar} shows that, at the smallest tested rank,
LRFShap truncates at shorter prefixes than FreeShap for both BERT and
Llama-2 on MR.
In these settings, evaluating fewer coalitions provides additional
runtime savings alongside the lower computational cost of each
coalition evaluation.
The effect of low-rank approximation on truncation depends on the model,
dataset, and approximation rank.
Additional results are provided in Appendix~\ref{app:convergence_ratio}.

\underline{\textbf{Q. Does LRFShap work across models and datasets?}}

\textbf{A. Yes. LRFShap can be used on diverse models and datasets.}

Figures~\ref{fig:sel_rem_wld_5000_bert_20}--\ref{fig:resnet_5000} in Appendix~\ref{sec:more_experiments} show the results for BERT, Llama-2, and ResNet-18 on the diverse datasets.
We observe the same tendency as in these figures: with a moderate rank, LRFShap is comparable to the baseline, and with a high rank, it surpasses the baseline in some cases.
These results demonstrate the effectiveness of LRFShap across the evaluated models and datasets, with suitable ranks varying by setting.

\section{Conclusion}\label{sec:conclusion}

We studied how low-rank kernel approximation affects individual Data Shapley values. In regularized kernel regression, we established stability bounds for finite-time training and kernel approximation, while showing that grand-coalition predictive agreement alone does not guarantee preservation of individual data values. Our counterexample for uniform Nystr\"om sampling makes this distinction explicit, and our empirical comparisons show that eigendecomposition provides more faithful Shapley approximations in the settings studied. Guided by this analysis, we proposed LRFShap, which accelerates coalition evaluation by reusing low-dimensional kernel features. Experiments demonstrate its effectiveness for eNTK-based data valuation, where kernel regression serves as a surrogate for model fine-tuning. With suitable ranks, LRFShap maintains performance comparable to FreeShap on data selection, data removal, and wrong-label detection while substantially reducing valuation time. These results emphasize that low-rank approximations for data valuation should be assessed beyond their predictive performance on the full dataset.

\paragraph{Limitations and future work.}

Our stability bound assumes squared-error utilities, a fixed test--train kernel block, and coalition-size-scaled ridge.
The analysis readily extends to joint approximation of both kernel blocks by accounting for their respective errors, whereas accuracy-based guarantees require additional control of classification margins across coalitions.
Future work could develop principled rank-selection methods that account for differences across datasets and valuation tasks, and examine how coalition-size-scaled ridge, fixed ridge, and alternative regularization penalties affect generalization across coalition sizes.
Further extensions include evolving kernels that capture representation changes during fine-tuning, and low-rank approximation in distributional Shapley \citep{ghorbani2020distributional} and MMD-based data distribution valuation \citep{xu2024data}.

\subsection*{AI use statement}

We used generative AI tools to assist with drafting and polishing the manuscript, identifying relevant literature, exploring research ideas, and developing mathematical arguments and proofs. The authors reviewed and revised all AI-assisted content, checked cited sources, and independently verified the mathematical arguments included in the paper. We take responsibility for the final content of this work, including all text, claims, and artifacts produced with the assistance of generative AI.

\subsection*{Reproducibility statement}

We describe the kernel regression setting and the LRFShap algorithm in the main text. The appendix provides the assumptions and proofs for our theoretical results. The experimental setup and supplementary details specify the datasets, pretrained models, kernel construction, approximation ranks, regularization parameters, and evaluation procedures used in our experiments.

\bibliography{bib}

\begin{thebibliography}{52}
\providecommand{\natexlab}[1]{#1}
\providecommand{\url}[1]{\texttt{#1}}
\expandafter\ifx\csname urlstyle\endcsname\relax
  \providecommand{\doi}[1]{doi: #1}\else
  \providecommand{\doi}{doi: \begingroup \urlstyle{rm}\Url}\fi

\bibitem[Abdin et~al.(2024)Abdin, Aneja, Behl, Bubeck, Eldan, Gunasekar, Harrison, Hewett, Javaheripi, Kauffmann, et~al.]{abdin2024phi}
Marah Abdin, Jyoti Aneja, Harkirat Behl, S{\'e}bastien Bubeck, Ronen Eldan, Suriya Gunasekar, Michael Harrison, Russell~J Hewett, Mojan Javaheripi, Piero Kauffmann, et~al.
\newblock Phi-4 technical report.
\newblock \emph{Microsoft tech blog}, 2024.
\newblock URL \url{https://www.microsoft.com/en-us/research/publication/phi-4-technical-report/}.

\bibitem[Alaoui \& Mahoney(2015)Alaoui and Mahoney]{alaoui2015fast}
Ahmed Alaoui and Michael~W. Mahoney.
\newblock Fast randomized kernel ridge regression with statistical guarantees.
\newblock \emph{Advances in Neural Information Processing Systems}, 2015.

\bibitem[Alemohammad et~al.(2024)Alemohammad, Casco-Rodriguez, Luzi, Humayun, Babaei, LeJeune, Siahkoohi, and Baraniuk]{alemohammad2024self}
Sina Alemohammad, Josue Casco-Rodriguez, Lorenzo Luzi, Ahmed~Imtiaz Humayun, Hossein Babaei, Daniel LeJeune, Ali Siahkoohi, and Richard Baraniuk.
\newblock Self-consuming generative models go mad.
\newblock \emph{International Conference on Learning Representations}, 2024.

\bibitem[Arora et~al.(2019)Arora, Du, Hu, Li, Salakhutdinov, and Wang]{arora2019exact}
Sanjeev Arora, Simon~S Du, Wei Hu, Zhiyuan Li, Russ~R Salakhutdinov, and Ruosong Wang.
\newblock On exact computation with an infinitely wide neural net.
\newblock \emph{Advances in Neural Information Processing Systems}, 2019.

\bibitem[Bach(2013)]{bach2013sharp}
Francis Bach.
\newblock Sharp analysis of low-rank kernel matrix approximations.
\newblock \emph{Conference on Learning Theory}, 2013.

\bibitem[Burgess \& Chapman(2021)Burgess and Chapman]{burgess2021approximating}
Mark~Alexander Burgess and Archie~C Chapman.
\newblock Approximating the shapley value using stratified empirical bernstein sampling.
\newblock \emph{International Joint Conferences on Artificial Intelligence}, 2021.

\bibitem[Chen et~al.(2021)Chen, Li, Yu, Wu, and Miao]{chen2021hydra}
Yuanyuan Chen, Boyang Li, Han Yu, Pengcheng Wu, and Chunyan Miao.
\newblock Hydra: Hypergradient data relevance analysis for interpreting deep neural networks.
\newblock \emph{Association for the Advancement of Artificial Intelligence}, 2021.

\bibitem[Cortes et~al.(2010)Cortes, Mohri, and Talwalkar]{cortes2010impact}
Corinna Cortes, Mehryar Mohri, and Ameet Talwalkar.
\newblock On the impact of kernel approximation on learning accuracy.
\newblock \emph{International Conference on Artificial Intelligence and Statistics}, 2010.

\bibitem[Devlin et~al.(2019)Devlin, Chang, Lee, and Toutanova]{devlin2019bert}
Jacob Devlin, Ming-Wei Chang, Kenton Lee, and Kristina Toutanova.
\newblock Bert: Pre-training of deep bidirectional transformers for language understanding.
\newblock \emph{Conference of the North American Chapter of the Association for Computational Linguistics}, 2019.

\bibitem[Dolan \& Brockett(2005)Dolan and Brockett]{mrpc}
William~B. Dolan and Chris Brockett.
\newblock Automatically constructing a corpus of sentential paraphrases.
\newblock \emph{International Workshop on Paraphrasing}, 2005.

\bibitem[Drineas \& Mahoney(2005)Drineas and Mahoney]{drineas2005nystrom}
Petros Drineas and Michael~W Mahoney.
\newblock On the nystr{\"o}m method for approximating a gram matrix for improved kernel-based learning.
\newblock \emph{Journal of Machine Learning Research}, 2005.

\bibitem[Eckart \& Young(1936)Eckart and Young]{eckart1936approximation}
Carl Eckart and Gale Young.
\newblock \emph{The approximation of one matrix by another of lower rank}.
\newblock Psychometrika, 1936.

\bibitem[Ghorbani \& Zou(2019)Ghorbani and Zou]{ghorbani2019data}
Amirata Ghorbani and James Zou.
\newblock Data shapley: Equitable valuation of data for machine learning.
\newblock \emph{International Conference on Machine Learning}, 2019.

\bibitem[Ghorbani et~al.(2020)Ghorbani, Kim, and Zou]{ghorbani2020distributional}
Amirata Ghorbani, Michael Kim, and James Zou.
\newblock A distributional framework for data valuation.
\newblock \emph{International Conference on Machine Learning}, 2020.

\bibitem[Giampiccolo et~al.(2007)Giampiccolo, Magnini, Dagan, and Dolan]{rte}
Danilo Giampiccolo, Bernardo Magnini, Ido Dagan, and Bill Dolan.
\newblock The third {PASCAL} recognizing textual entailment challenge.
\newblock \emph{ACL-PASCAL Workshop on Textual Entailment and Paraphrasing}, 2007.

\bibitem[Haas et~al.(2023)Haas, Holzm{\"u}ller, Luxburg, and Steinwart]{haas2023mind}
Moritz Haas, David Holzm{\"u}ller, Ulrike Luxburg, and Ingo Steinwart.
\newblock Mind the spikes: Benign overfitting of kernels and neural networks in fixed dimension.
\newblock \emph{Advances in Neural Information Processing Systems}, 2023.

\bibitem[Halko et~al.(2011)Halko, Martinsson, and Tropp]{halko2011finding}
Nathan Halko, Per-Gunnar Martinsson, and Joel~A. Tropp.
\newblock Finding structure with randomness: Probabilistic algorithms for constructing approximate matrix decompositions.
\newblock \emph{SIAM Review}, 2011.

\bibitem[He et~al.(2016)He, Zhang, Ren, and Sun]{he2016deep}
Kaiming He, Xiangyu Zhang, Shaoqing Ren, and Jian Sun.
\newblock Deep residual learning for image recognition.
\newblock \emph{IEEE conference on Computer Vision and Pattern Recognition}, 2016.

\bibitem[Jacot et~al.(2018)Jacot, Gabriel, and Hongler]{jacot2018neural}
Arthur Jacot, Franck Gabriel, and Cl{\'e}ment Hongler.
\newblock Neural tangent kernel: convergence and generalization in neural networks.
\newblock \emph{Advances in Neural Information Processing Systems}, 2018.

\bibitem[Jia et~al.(2019)Jia, Dao, Wang, Hubis, Hynes, G{\"u}rel, Li, Zhang, Song, and Spanos]{jia2019towards}
Ruoxi Jia, David Dao, Boxin Wang, Frances~Ann Hubis, Nick Hynes, Nezihe~Merve G{\"u}rel, Bo~Li, Ce~Zhang, Dawn Song, and Costas~J Spanos.
\newblock Towards efficient data valuation based on the shapley value.
\newblock \emph{International Conference on Artificial Intelligence and Statistics}, 2019.

\bibitem[Koh \& Liang(2017)Koh and Liang]{koh2017understanding}
Pang~Wei Koh and Percy Liang.
\newblock Understanding black-box predictions via influence functions.
\newblock \emph{International conference on machine learning}, 2017.

\bibitem[Krizhevsky(2009)]{krizhevsky2009learning}
Alex Krizhevsky.
\newblock Learning multiple layers of features from tiny images.
\newblock \emph{University of Toronto}, 2009.

\bibitem[Kwon \& Zou(2022)Kwon and Zou]{kwon2021beta}
Yongchan Kwon and James Zou.
\newblock Beta shapley: a unified and noise-reduced data valuation framework for machine learning.
\newblock \emph{International Conference on Artificial Intelligence and Statistics}, 2022.

\bibitem[Lee et~al.(2019)Lee, Xiao, Schoenholz, Bahri, Novak, Sohl-Dickstein, and Pennington]{lee2019wide}
Jaehoon Lee, Lechao Xiao, Samuel Schoenholz, Yasaman Bahri, Roman Novak, Jascha Sohl-Dickstein, and Jeffrey Pennington.
\newblock Wide neural networks of any depth evolve as linear models under gradient descent.
\newblock \emph{Advances in Neural Information Processing Systems}, 2019.

\bibitem[Lee(2013)]{lee2013smooth}
John~M Lee.
\newblock \emph{Introduction to smooth manifolds}.
\newblock Springer, 2013.

\bibitem[Li et~al.(2016)Li, Jegelka, and Sra]{li2016fast}
Chengtao Li, Stefanie Jegelka, and Suvrit Sra.
\newblock Fast dpp sampling for nystrom with application to kernel methods.
\newblock \emph{International Conference on Machine Learning}, 2016.

\bibitem[Li \& Yu(2024)Li and Yu]{li2024faster}
Weida Li and Yaoliang Yu.
\newblock Faster approximation of probabilistic and distributional values via least squares.
\newblock \emph{International Conference on Learning Representations}, 2024.

\bibitem[Maleki et~al.(2013)Maleki, Tran-Thanh, Hines, Rahwan, and Rogers]{maleki2013bounding}
Sasan Maleki, Long Tran-Thanh, Greg Hines, Talal Rahwan, and Alex Rogers.
\newblock Bounding the estimation error of sampling-based shapley value approximation.
\newblock \emph{arXiv:1306.4265}, 2013.

\bibitem[Malladi et~al.(2023)Malladi, Wettig, Yu, Chen, and Arora]{malladi2023kernel}
Sadhika Malladi, Alexander Wettig, Dingli Yu, Danqi Chen, and Sanjeev Arora.
\newblock A kernel-based view of language model fine-tuning.
\newblock \emph{International Conference on Machine Learning}, 2023.

\bibitem[Mitchell et~al.(2022)Mitchell, Cooper, Frank, and Holmes]{mitchell2022sampling}
Rory Mitchell, Joshua Cooper, Eibe Frank, and Geoffrey Holmes.
\newblock Sampling permutations for {S}hapley value estimation.
\newblock \emph{Journal of Machine Learning Research}, 2022.

\bibitem[Mohamadi et~al.(2023)Mohamadi, Bae, and Sutherland]{mohamadi2023fast}
Mohamad~Amin Mohamadi, Wonho Bae, and Danica~J Sutherland.
\newblock A fast, well-founded approximation to the empirical neural tangent kernel.
\newblock \emph{International Conference on Machine Learning}, 2023.

\bibitem[Musco \& Musco(2017)Musco and Musco]{musco2017recursive}
Cameron Musco and Christopher Musco.
\newblock Recursive sampling for the {N}ystr{\"o}m method.
\newblock \emph{Advances in Neural Information Processing Systems}, 2017.

\bibitem[Okhrati \& Lipani(2021)Okhrati and Lipani]{okhrati2021multilinear}
Ramin Okhrati and Aldo Lipani.
\newblock A multilinear sampling algorithm to estimate shapley values.
\newblock \emph{International Conference on Pattern Recognition}, 2021.

\bibitem[Pang \& Lee(2005)Pang and Lee]{pang2005seeing}
Bo~Pang and Lillian Lee.
\newblock Seeing stars: Exploiting class relationships for sentiment categorization with respect to rating scales.
\newblock \emph{Annual Meeting of the Association for Computational Linguistics}, 2005.

\bibitem[Penedo et~al.(2024)Penedo, Kydl{\'\i}{\v{c}}ek, Lozhkov, Mitchell, Raffel, Von~Werra, Wolf, et~al.]{penedo2024fineweb}
Guilherme Penedo, Hynek Kydl{\'\i}{\v{c}}ek, Anton Lozhkov, Margaret Mitchell, Colin Raffel, Leandro Von~Werra, Thomas Wolf, et~al.
\newblock The fineweb datasets: Decanting the web for the finest text data at scale.
\newblock \emph{Advances in Neural Information Processing Systems}, 2024.

\bibitem[Pruthi et~al.(2020)Pruthi, Liu, Kale, and Sundararajan]{pruthi2020estimating}
Garima Pruthi, Frederick Liu, Satyen Kale, and Mukund Sundararajan.
\newblock Estimating training data influence by tracing gradient descent.
\newblock \emph{Advances in neural information processing systems}, 2020.

\bibitem[Shapley(1953)]{shapley1953value}
Lloyd~S. Shapley.
\newblock \emph{A value for n-person games}.
\newblock Princeton University Press, 1953.

\bibitem[Shumailov et~al.(2024)Shumailov, Shumaylov, Zhao, Papernot, Anderson, and Gal]{shumailov2024ai}
Ilia Shumailov, Zakhar Shumaylov, Yiren Zhao, Nicolas Papernot, Ross Anderson, and Yarin Gal.
\newblock Ai models collapse when trained on recursively generated data.
\newblock \emph{Nature}, 2024.

\bibitem[Socher et~al.(2013)Socher, Perelygin, Wu, Chuang, Manning, Ng, and Potts]{sst-2}
Richard Socher, Alex Perelygin, Jean~Y Wu, Jason Chuang, Christopher~D Manning, Andrew~Y Ng, and Christopher Potts.
\newblock Recursive deep models for semantic compositionality over a sentiment treebank.
\newblock \emph{Conference on Empirical Methods in Natural Language Processing}, 2013.

\bibitem[Touvron et~al.(2023)Touvron, Martin, Stone, Albert, Almahairi, Babaei, Bashlykov, Batra, Bhargava, Bhosale, et~al.]{touvron2023llama}
Hugo Touvron, Louis Martin, Kevin Stone, Peter Albert, Amjad Almahairi, Yasmine Babaei, Nikolay Bashlykov, Soumya Batra, Prajjwal Bhargava, Shruti Bhosale, et~al.
\newblock Llama 2: Open foundation and fine-tuned chat models.
\newblock \emph{arXiv:2307.09288}, 2023.

\bibitem[Wang et~al.(2019)Wang, Singh, Michael, Hill, Levy, and Bowman]{wang2019glue}
Alex Wang, Amanpreet Singh, Julian Michael, Felix Hill, Omer Levy, and Samuel~R. Bowman.
\newblock {GLUE}: A multi-task benchmark and analysis platform for natural language understanding.
\newblock \emph{International Conference on Learning Representations}, 2019.

\bibitem[Wang et~al.(2025)Wang, Mittal, Song, and Jia]{wang2025data}
Jiachen~Tianhao Wang, Prateek Mittal, Dawn Song, and Ruoxi Jia.
\newblock Data shapley in one training run.
\newblock \emph{International Conference on Learning Representations}, 2025.

\bibitem[Wang et~al.(2024)Wang, Lin, Qiao, Foo, and Low]{wang2024helpful}
Jingtan Wang, Xiaoqiang Lin, Rui Qiao, Chuan-Sheng Foo, and Bryan Kian~Hsiang Low.
\newblock Helpful or harmful data? fine-tuning-free shapley attribution for explaining language model predictions.
\newblock \emph{International Conference on Machine Learning}, 2024.

\bibitem[Wang \& Zhang(2013)Wang and Zhang]{wang2013improving}
Shusen Wang and Zhihua Zhang.
\newblock Improving cur matrix decomposition and the nystr{\"o}m approximation via adaptive sampling.
\newblock \emph{Journal of Machine Learning Research}, 2013.

\bibitem[Wei et~al.(2022)Wei, Hu, and Steinhardt]{wei2022more}
Alexander Wei, Wei Hu, and Jacob Steinhardt.
\newblock More than a toy: Random matrix models predict how real-world neural representations generalize.
\newblock \emph{International Conference on Machine Learning}, 2022.

\bibitem[Williams et~al.(2018)Williams, Nangia, and Bowman]{mnli}
Adina Williams, Nikita Nangia, and Samuel~R Bowman.
\newblock A broad-coverage challenge corpus for sentence understanding through inference.
\newblock \emph{Conference of the North American Chapter of the Association for Computational Linguistics}, 2018.

\bibitem[Williams \& Seeger(2000)Williams and Seeger]{williams2000using}
Christopher K.~I. Williams and Matthias Seeger.
\newblock Using the {N}ystr{\"o}m method to speed up kernel machines.
\newblock \emph{Advances in Neural Information Processing Systems}, 2000.

\bibitem[Wu et~al.(2022)Wu, Shu, and Low]{wu2022davinz}
Zhaoxuan Wu, Yao Shu, and Bryan Kian~Hsiang Low.
\newblock Davinz: Data valuation using deep neural networks at initialization.
\newblock \emph{International Conference on Machine Learning}, 2022.

\bibitem[Xu et~al.(2024)Xu, Wang, Foo, Low, and Fanti]{xu2024data}
Xinyi Xu, Shuaiqi Wang, Chuan-Sheng Foo, Bryan~K Low, and Giulia Fanti.
\newblock Data distribution valuation.
\newblock \emph{Advances in Neural Information Processing Systems}, 2024.

\bibitem[Yang et~al.(2012)Yang, Li, Mahdavi, Jin, and Zhou]{yang2012nystrom}
Tianbao Yang, Yu-Feng Li, Mehrdad Mahdavi, Rong Jin, and Zhi-Hua Zhou.
\newblock Nystr{\"o}m method vs random fourier features: A theoretical and empirical comparison.
\newblock \emph{Advances in neural information processing systems}, 2012.

\bibitem[Zhang et~al.(2008)Zhang, Tsang, and Kwok]{zhang2008improved}
Kai Zhang, Ivor~W Tsang, and James~T Kwok.
\newblock Improved nystr{\"o}m low-rank approximation and error analysis.
\newblock \emph{International Conference on Machine Learning}, 2008.

\bibitem[Zhang et~al.(2015)Zhang, Zhao, and LeCun]{zhang2015character}
Xiang Zhang, Junbo Zhao, and Yann LeCun.
\newblock Character-level convolutional networks for text classification.
\newblock \emph{Advances in Neural Information Processing Systems}, 2015.

\end{thebibliography}
\bibliographystyle{iclr2027_conference}

\clearpage
\appendix

\etocdepthtag.toc{appendix}

\etocsettagdepth{main}{none}
\etocsettagdepth{appendix}{subsection}
\etocsetnexttocdepth{subsection}

\renewcommand{\contentsname}{Appendix Contents}
\tableofcontents

\clearpage
\section{Related work}\label{appendix:related_work}

\paragraph{Reducing the combinatorial cost of Data Shapley.}
Data Shapley~\citep{ghorbani2019data,jia2019towards} formulates
data valuation as a cooperative game, treating each training sample
as a player and using the Shapley value to quantify its contribution.
To mitigate the combinatorial cost of Shapley value computation,
existing methods improve estimation efficiency through
group-testing-based estimation~\citep{jia2019towards},
multilinear-extension-based sampling~\citep{okhrati2021multilinear},
adaptive stratified sampling~\citep{burgess2021approximating},
and structured permutation sampling~\citep{mitchell2022sampling}.
\citet{li2024faster} develop least-squares-based estimators
for probabilistic values, including the Shapley value,
providing another approach to efficient data valuation.

\paragraph{Reducing model retraining costs.}
Classical leave-one-out (LOO) attribution evaluates marginal
contributions to the remaining training set, whereas Data Shapley
averages them across all subsets of that set.
LOO scores can have low signal-to-noise ratios and unstable
signs under dataset resampling~\citep{kwon2021beta,wang2024helpful}.
Influence functions~\citep{koh2017understanding} approximate LOO attribution. Dynamic methods such as
TracIn~\citep{pruthi2020estimating} and
HyDRA~\citep{chen2021hydra} estimate data influence along
training trajectories, while In-Run Data
Shapley~\citep{wang2025data} defines local Shapley games
within a single training run.
Among NTK-based approaches, DAVINZ~\citep{wu2022davinz}
uses a generalization bound incorporating distributional
differences as its utility.
FreeShap~\citep{wang2024helpful} approximates coalition-wise
fine-tuning through empirical NTK regression and evaluates
utility using predictive accuracy.
We adopt FreeShap as our baseline because its kernel regression
formulation facilitates low-rank analysis of Data Shapley
and supports attribution for large pretrained models.

\paragraph{Low-rank approximation of kernel matrices.}
Classical approaches include truncated
eigendecomposition~\citep{eckart1936approximation} and
Nystr\"om approximation~\citep{williams2000using,drineas2005nystrom}.
CUR decomposition generalizes the Nystr\"om framework to arbitrary
matrices through column and row selection; see
\citet{wang2013improving} for an overview of these methods
and their variants.
Landmark-selection strategies include clustering-based
selection~\citep{zhang2008improved},
ridge leverage-score sampling~\citep{alaoui2015fast,musco2017recursive},
and determinantal point process sampling~\citep{li2016fast}. The relationship between kernel approximation and predictive
performance has also been studied.
\citet{yang2012nystrom} compare Nystr\"om and random Fourier
features, showing that Nystr\"om can achieve sharper
generalization bounds under suitable spectral conditions.
\citet{bach2013sharp} directly analyzes the fixed-design
prediction risk of Nystr\"om kernel ridge regression,
obtaining relative-risk guarantees without relying on
a small kernel-matrix approximation error.
Our work investigates how kernel approximation affects
individual Data Shapley values and whether predictive
agreement suffices to preserve these values.

\clearpage
\section{Neural Tangent Kernel}
\label{appendix:NTK_derivation}

We consider a parametric model
\(
f(x;\theta): \mathbb{R}^d \to \mathbb{R}^{|C|}
\)
that is differentiable with respect to the parameter vector
\(\theta \in \mathbb{R}^P\).
Let \(\theta_0 \in \mathbb{R}^P\) denote an initialization of the model parameters.
The empirical neural tangent kernel (eNTK) associated with the model at \(\theta_0\)
is defined, for any pair of inputs \(x_1, x_2 \in \mathbb{R}^d\), as
\begin{align}
\label{eq:eNTK}
\Theta_{\theta_0}(x_1,x_2)
:=
\underbrace{
\left[
\frac{\partial f(x_1;\theta_0)}{\partial \theta_0}
\right]
}_{{|C|}\times P}
\underbrace{
\left[
\frac{\partial f(x_2;\theta_0)}{\partial \theta_0}
\right]^{\top}
}_{P\times {|C|}}
\;\in\;
\mathbb{R}^{{|C|}\times {|C|}}.
\end{align}
Here,
\(
\frac{\partial f(x;\theta_0)}{\partial \theta_0} \in \mathbb{R}^{{|C|}\times P}
\)
denotes the Jacobian of the model output with respect to the flattened parameter vector \(\theta_0\). 

In the \emph{kernel regime}, the network parameters remain sufficiently
close to their initialization throughout training, so that the network
is well approximated by its first-order linearization
\(f(x;\theta_t)\approx f(x;\theta_0)
+\frac{\partial f(x;\theta_0)}{\partial\theta_0}
(\theta_t-\theta_0)\) for every \(t\geq0\). At the same time, the
empirical neural tangent kernel remains approximately constant along
the training trajectory:
\(\Theta_{\theta_t}(x_1,x_2)\approx
\Theta_{\theta_0}(x_1,x_2)\) for every
\(x_1,x_2\in\mathbb R^d\). Consequently, the evolution of the network
outputs is governed by the fixed kernel determined at initialization. Under the kernel approximation, parameter-space gradient flow reduces
to kernel gradient flow in function space with respect to the fixed
eNTK \(\Theta_{\theta_0}\). For the squared loss, its infinite-time
limit corresponds to kernel regression associated with
\(\Theta_{\theta_0}\).

Let the training dataset be denoted by $D_S$ with index set 
$S := \{1,\ldots,|S|\}$.
We define the input matrix as
\(
X_S := [x_1,\ldots,x_{|S|}]^\top
\)
and the corresponding one-hot encoded label matrix as
\(
Y_S :=
[y_1,\dots,y_{|S|}]^\top
\in
\{0,1\}^{|S|\times {|C|}}.
\) We further use $\operatorname{vec}(Y_S)\in\{0,1\}^{|S|{|C|}}$ to denote the vectorized
form of the label matrix.

Given a test input $x_t$, the prediction of the eNTK regression model is given by
\begin{equation}
f^{\mathrm{eNTK}}_S(x_t)
=
\underbrace{f(x_{t};\theta_0)}_{{|C|}\times 1}
+
\underbrace{K(x_t,X_S)^\top}_{{|C|}\times |S|{|C|}}
\underbrace{K(X_S,X_S)^{\dagger}}_{|S|{|C|}\times |S|{|C|}}
\underbrace{\operatorname{vec}(Y_S - f(X_{S};\theta_0) )}_{|S|{|C|}\times 1} ,
\label{eq:entk_regression}
\end{equation}
where
$
K(x_t,X_S)
=
\bigl[\Theta_{\theta_0}(x_t,x_j)\bigr]_{j=1}^{|S|}
$ and $
K(X_S,X_S)
=
\bigl[\Theta_{\theta_0}(x_i,x_j)\bigr]_{i,j=1}^{|S|}
.
$

We make two remarks regarding the practical use of the empirical neural tangent
kernel.

First, one may ask whether it is possible to use pretrained weights, rather than
random initialization, as the reference parameter $\theta_0$ for defining the
empirical neural tangent kernel.
In the context of natural language processing, this question was investigated by
\cite{malladi2023kernel}, while related results for computer vision tasks were
reported in \cite{wei2022more}.

Second, directly inverting the kernel matrix $K(X_S,X_S)$ in
\eqref{eq:entk_regression} requires $\mathcal{O}((|S|{|C|})^3)$ computational cost,
which becomes prohibitive when the number of classes $|C|$ is large.
To mitigate this class-dependent inefficiency, various approximation techniques
have been proposed, including the first-logit approximation.

From a theoretical perspective, \cite{mohamadi2023fast} provides a principled justification for such approximations. Specifically, they consider a multi-output model
$f(x;\theta) = [f^{1}(x;\theta),\ldots,f^{|C|}(x;\theta)]$ and introduces a
pseudo neural tangent kernel (pNTK) defined by
\begin{align}
    \hat{\Theta}_{\theta_0}(x_1,x_2)
    :=
    \underbrace{\left[
    \frac{\partial}{\partial \theta_0}
    \left(
    \frac{1}{\sqrt{{|C|}}}
    \sum_{i=1}^{{|C|}} f^{i}(x_1;\theta_0)
    \right)\right]
    }_{1\times P}
    \underbrace{\left[
    \frac{\partial}{\partial \theta_0}
    \left(
    \frac{1}{\sqrt{{|C|}}}
    \sum_{i=1}^{{|C|}} f^{i}(x_2;\theta_0)
    \right)\right]^{\top}
    }_{P\times 1}.
\end{align}

Using this kernel, they show that the regression predictor in \eqref{eq:entk_regression} admits the following closed-form approximation:
\begin{align}
    f^{\mathrm{pNTK}}_S(x_t)
    =
    \underbrace{f(x_t;\theta_0)}_{1\times {|C|}}
    +
    \underbrace{\hat{K}(x_t,X_S)^\top}_{1\times |S|}
    \underbrace{\hat{K}(X_S,X_S)^{\dagger}}_{|S|\times |S|}
    \underbrace{(Y_S-f(X_{S};\theta_0))}_{|S|\times {|C|}},
    \label{eq:pntk_regression}
\end{align}
where
$
\hat{K}(x_t,X_S)
=
\bigl[\hat{\Theta}_{\theta_0}(x_t,x_j)\bigr]_{j=1}^{|S|}
\in \mathbb{R}^{|S|}$ and $
\hat{K}(X_S,X_S)
=
\bigl[\hat{\Theta}_{\theta_0}(x_i,x_j)\bigr]_{i,j=1}^{|S|}
\in \mathbb{R}^{|S|\times |S|}.
$

Importantly, this approximation has been shown empirically to remain effective
when $\theta_0$ is chosen as pretrained weights, rather than random initialization. To facilitate subsequent analysis, we rewrite \eqref{eq:pntk_regression}
in vectorized form over a test set $X_T$ and introduce an explicit ridge
regularization parameter $\gamma>0$. This yields
\begin{align}
    F^{\mathrm{pNTK}}_S(X_{T})
    =
    \underbrace{f(X_{T};\theta_0)}_{|T|\times {|C|}}
    +
    \underbrace{\hat{K}(X_{T},X_S)^\top}_{|T|\times |S|}
    \underbrace{(\hat{K}(X_S,X_S)+\gamma I_{|S|})^{-1}}_{|S|\times |S|}
    \underbrace{(Y_S-f(X_{S};\theta_0))}_{|S|\times {|C|}}.
    \label{eq:pntk_regression2 vector}
\end{align}
Define the residual targets
$\tilde{Y}_S := Y_S - f(X_S;\theta_0)$
and the residual predictor
$\tilde{F}^{\mathrm{pNTK}}_S(X_T)
:= F^{\mathrm{pNTK}}_S(X_T) - f(X_T;\theta_0)$.
Then \eqref{eq:pntk_regression2 vector} can be equivalently expressed as
\begin{align}
    \tilde{F}^{\mathrm{pNTK}}_S(X_T)
    =
    k(X_T,X_S)^\top
    \bigl(k(X_S,X_S)+\gamma I_{|S|}\bigr)^{-1}
    \tilde{Y}_S,
    \label{eq:pntk_regression_centered}
\end{align}
where we write $k=\hat{K}$ for simplicity.

At independent Gaussian initialization, the first-logit and sum-of-logits scalarizations agree in distribution. Consequently, the random-initialization approximation guarantee of \citet{mohamadi2023fast} applies to both constructions. In this work, however, we use the first-logit kernel evaluated at pretrained parameters; we do not assert equality between the two kernels for a fixed pretrained network.

\section{Solutions by kernel regression}
\label{appendix:Solutions by kernel regression}

Let
\(
k:\mathbb{R}^d \times \mathbb{R}^d \to \mathbb{R}
\)
be a positive semi-definite kernel function, and let
$\mathcal{H}_k$ denote the associated reproducing kernel
Hilbert space (RKHS).
Given a training dataset $D_S = \{(x_i,y_i)\}_{i=1}^{|S|}$ with $x_i \in \mathbb{R}^d$ and $y_i \in \mathbb{R}$, we consider the following empirical risk functional:
\begin{align}
\mathcal{L}_{\rho}^{S}(f)
:=
\frac{1}{|S|}\sum_{i=1}^{|S|}\bigl(f(x_i)-y_i\bigr)^2
+
\rho\|f\|_{\mathcal{H}_k}^{2}
\label{eq:L_risk_object}
\end{align}
where $\rho \ge 0$ is a regularization parameter.

Let \( f_{t,\rho}^{S} \) denote the function obtained by running gradient flow
on the regularized objective \( \mathcal L_\rho^{S} \) in the RKHS \( \mathcal H_k \), initialized at \( f_0 = 0 \) and evolved up to time \( t \in [0,\infty) \)
with a fixed regularization parameter \( \rho > 0 \).
According to Appendix C of \cite{haas2023mind},
the solution admits the explicit representation
\begin{align}
f_{t,\rho}^{S}(x)
=
k(x,X_{S})\Bigl(
I_{|S|} - e^{-\frac{2}{|S|}t\bigl(k(X_{S},X_{S})+\rho |S| I_{|S|}\bigr)}
\Bigr)
\bigl(k(X_{S},X_{S})+\rho |S| I_{|S|}\bigr)^{-1}Y_{S},
\label{eq:gf_closed_form_appendix}
\end{align}
where \( k(x,X_{S}) := [k(x,x_i)]_{i\in S} \) and
\( k(X_{S},X_{S}) := [k(x_i,x_j)]_{i,j\in S} \) denotes the kernel matrix.

This expression provides a unified perspective on several classical kernel estimators:
\begin{enumerate}
    \item kernel ridge regression in the limit \( t \to \infty \);
    \item gradient flow on the unregularized objective as \( \rho \downarrow 0 \);
    \item ridgeless kernel regression
    $f_{\infty,0}(x) = k(x,X_{S})\,k(X_{S},X_{S})^{\dagger} y$
    in the joint limit \( \rho \to 0 \) and \( t \to \infty \).
\end{enumerate}

\subsection{Spectral analysis for the finite-time error of gradient flow}
\label{appendix:Spectral analysis for the finite-time error of gradient flow}
Our goal is to analyze the finite-time error
\(
\|f_{\infty,\rho}^{S}-f_{t,\rho}^{S}\|_{\mathcal H_k}^2
\)
of the gradient flow solution for each subset \(S\subseteq N\).
To this end, we express $f_{t,\rho}^{S}(x)
=
k(x,X_S)\,\alpha_{t,\rho}^{S}$
where the coefficient vector \(\alpha_{t,\rho}^{S}\in\mathbb{R}^{|S|}\) is given by
\begin{equation}
\alpha_{t,\rho}^{S}
=
\Bigl(
I_{|S|}
-
e^{-\frac{2}{|S|}t\bigl(k(X_{S},X_{S})+\rho |S| I_{|S|}\bigr)}
\Bigr)
\bigl(k(X_{S},X_{S})+\rho |S| I_{|S|}\bigr)^{-1}Y_{S}.
\label{eq:alpha_t}
\end{equation}
Then, the limiting coefficient vector is
\(
\alpha_{\infty,\rho}^{S}
=
\bigl(k(X_{S},X_{S})+\rho |S| I_{|S|}\bigr)^{-1} Y_S,
\)
and the corresponding finite-time residual is
\(
\Delta\alpha_{\rho}^{S}
:=
\alpha_{t,\rho}^{S}-\alpha_{\infty,\rho}^{S}
=
-
e^{-\frac{2}{|S|}t\bigl(k(X_{S},X_{S})+\rho |S| I_{|S|}\bigr)}
\bigl(k(X_{S},X_{S})+\rho |S| I_{|S|}\bigr)^{-1}Y_S .\)
From this, since
$
f_{\infty,\rho}^{S}-f_{t,\rho}^{S}
=
k(\cdot,X_S)\,\Delta\alpha_{\rho}^{S},
$
the RKHS norm satisfies
\begin{align*}
    \|f_{\infty,\rho}^{S}-f_{t,\rho}^{S}\|_{\mathcal H}^2
    =
    (\Delta\alpha_{\rho}^{S})^\top k(X_{S},X_{S}) (\Delta\alpha_{\rho}^{S}).
\end{align*}
Using the eigen decomposition we have $k(X_{S},X_{S}) = V_{S}\Lambda_{S} V_{S}^{\top}$ where $\Lambda_{S} = \mathrm{diag}(\lambda_1^{S},\dots,\lambda_{|S|}^{S})$ is diagonal and $V_{S}\in \mathbb{R}^{|S|\times |S|}$ is orthonormal. Then we can represent the residual as
\begin{align*}
    \Delta\alpha_{\rho}^{S} =  V_{S}e^{-\frac{2}{|S|}t(\Lambda_{S}+\rho |S|I_{|S|})}(\Lambda_{S}+\rho |S|I_{|S|})^{-1}V_{S}^{\top}Y_S.
\end{align*}
Defining rotated coordinates
$\tilde{Y}_{S}:=V_{S}^{\top}Y_{S}$,
a direct computation yields the exact spectral identity
\begin{align}
    \|f_{\infty,\rho}^{S}-f_{t,\rho}^{S}\|_{\mathcal H}^2
    &=
    (V_{S}^{\top}\Delta\alpha_{\rho}^{S})^\top \Lambda_{S} (V_{S}^{\top} \Delta\alpha_{\rho}^{S})
    \\
    &=
    \sum_{i=1}^{|S|}\lambda_{i}^{|S|}(V_{S}^{\top} \Delta\alpha_{\rho}^{S})_{i}^{2}
    =
    \sum_{i=1}^{\lvert S \rvert}
    \lambda_i^{S}
    \left(
    \frac{e^{-\frac{2t}{\lvert S|}(\lambda_i^{S}+\rho \lvert S|)}}{\lambda_i^{S}+\rho \lvert S|}
    \right)^2
    (\tilde{Y}_{S,i})^2.
    \label{eq:spectral_error}
\end{align}

\paragraph{Case 1: $\rho = 0$ (ridgeless).}
When \(\rho=0\), only the strictly positive eigenvalues of $k(X_{S},X_{S})$ contribute:
\begin{align*}
\|f_{\infty,0}^{S}-f_{t,0}^{S}\|_{\mathcal H_k}^2
=
\sum_{i:\lambda_i^{S}>0}
\frac{
e^{-\frac{4t}{|S|}\lambda_i^{S}}
}{
\lambda_i^{S}
}
(\tilde{Y}_{S,i})^2.
\end{align*}
Since $e^{-x}\le \frac{1}{x}$ for $x>0$, we have
\begin{align}\label{eq:ridgeless_rkhsnormbound}
    \|f_{\infty,0}^{S}-f_{t,0}^{S}\|_{\mathcal H_k}^2
    \le
    \frac{|S|}{4t}\lVert k(X_{S},X_{S})^\dagger Y_{S}\rVert_{2}^{2}
    \le
    \frac{|S|\lVert Y_{S}\rVert_{2}^{2}}{4t(\lambda_{\min}^{+}(k(X_{S},X_{S})))^{2}},
\end{align}
where $\lambda_{\min}^{+}(k(X_{S},X_{S})):=\min\{\lambda_{i}^{S}:\lambda_{i}^{S}>0\}$.
\paragraph{Case 2: $\rho > 0$ (regularized).}
When \(\rho>0\), we have
\(
\lambda_i^{S}+\rho |S|
\ge
\rho |S|>0.
\)
Hence from \eqref{eq:spectral_error}, we have
\begin{align}
    \|f_{\infty,\rho}^{S}-f_{t,\rho}^{S}\|_{\mathcal H_k}^2
    \le
    e^{-4t\rho}
    \sum_{i=1}^{|S|}
    \frac{(\tilde{Y}_{S,i})^2}{\lambda_i^{S}+\rho |S|}
    =
    Y_{S}^{\top}(k(X_{S},X_{S})+\rho|S|I_{|S|})^{-1}Y_{S}
    \le
    \frac{e^{-4t\rho}}{\rho |S|}\lVert Y_{S}\rVert_{2}^{2}.
    \label{eq:rho>0 spectral bound}
\end{align}

\clearpage
\section{Theoretical understanding}\label{sec:theory_proof}

This appendix follows the scalar-response setting of Section~\ref{sec:KPS_diagram},
with notation summarized in Table~\ref{tab:appendix-c-notation}.
The notation below is recalled only for convenience; no new assumptions
are introduced.
When the training kernel matrix varies, \(Y_N\), \(Y_T\), and
\(K_{TN}\) remain fixed. Thus, in the kernel-approximation analysis,
only the train--train kernel matrix is perturbed, while the
test--train kernel blocks remain unchanged.

\begin{table}[htbp]
\centering
\caption{Notation used throughout Appendix~C.}
\label{tab:appendix-c-notation}
\small
\renewcommand{\arraystretch}{1.12}
\setlength{\tabcolsep}{5pt}

\begin{tabularx}{\linewidth}{
    @{}
    >{\raggedright\arraybackslash}p{0.31\linewidth}
    >{\raggedright\arraybackslash}X
    @{}
}
\toprule
Symbol & Meaning in Appendix~C \\
\midrule

\(N,T,S\)
&
Nonempty finite training- and test-index sets and a coalition
\(S\subseteq N\).
\\

\(D_S,X_S,Y_S\)
&
The coalition dataset, its input matrix, and its scalar-response
vector. In particular,
\(Y_S\in\mathbb R^{|S|}\).
\\

\(y_i,\kappa\)
&
Scalar responses satisfying
\(\lvert y_i\rvert\leq 1\), and the kernel bound
\(\kappa:=\sup_{x\in\mathbb R^d}\sqrt{k(x,x)}<\infty\).
\\

\(k,\mathcal H_k,K_{AB}\)
&
A fixed scalar-valued positive-semidefinite kernel, its RKHS, and
the kernel matrix indexed by
\(A,B\subseteq N\cup T\).
\\

\(K_{NN},L_{NN},G\)
&
The exact training kernel matrix, its positive-semidefinite
approximation, and a generic training-matrix argument, respectively.
\\

\(\rho,t,A_{S,\rho}(G)\)
&
The theoretical regularization parameter, gradient-flow time, and
the regularized coalition matrix
\(A_{S,\rho}(G):=G_{SS}+\rho|S|I_{|S|}\).
The main propositions assume \(\rho>0\); the case \(\rho=0\) is
considered only where explicitly stated.
\\

\(\mathcal P_{t,\rho}^{S}(G),
  \mathcal P_{\rho}^{S}(G),
  \mathcal P_{\rho}(G)\)
&
The finite-time coalition prediction, stationary coalition
prediction, and grand-coalition prediction, respectively, with
\(\mathcal P_{\rho}^{S}(G):=\mathcal P_{\infty,\rho}^{S}(G)\), 
\(\mathcal P_{\rho}(G):=\mathcal P_{\rho}^{N}(G)\).
\\

\(U_{t,\rho}^{G}(S)\)
&
The negative squared-test-error utility
\(
U_{t,\rho}^{G}(S)
:=
-|T|^{-1}
\lVert
Y_T-\mathcal P_{t,\rho}^{S}(G)
\rVert_2^2
\).
\\

\(\mathcal S_{t,\rho}(G),
  \mathcal S_{\rho}(G)\)
&
The finite-time and stationary Data Shapley maps, with
\(
\mathcal S_{t,\rho}(G)
:=
(\phi_i(U_{t,\rho}^{G}))_{i\in N}
\)
and
\(
\mathcal S_{\rho}
:=
\mathcal S_{\infty,\rho}
\).
\\

\bottomrule
\end{tabularx}
\end{table}

For the empty coalition, we use the conventions
$
\mathcal P_{t,\rho}^{\emptyset}(G)
=
\mathcal P_{\rho}^{\emptyset}(G)
=
0,
$
and
$
U_{t,\rho}^{G}(\emptyset)
=
U_{\infty,\rho}^{G}(\emptyset)
=
-\frac{1}{|T|}
\lVert Y_T\rVert_2^2.
$
The implementation parameter \(\gamma\) is a coalition-independent
ridge parameter and is distinct from the size-normalized theoretical
parameter \(\rho\). Symbols used only within one proof are introduced
locally when they first appear.

\begin{lemma}[Size-dependent perturbation bound for Shapley value]
\label{lemma: Size-dependent perturbation bound for Shapley value}
Let $N=\{1,\dots,n\}$ be the player set, and let 
$U,V:2^{N}\to\mathbb{R}$ be two cooperative games. Assume that there exists a function 
$g:\{0,1,\dots,n\}\to\mathbb{R}_{\ge 0}$
such that for every $l=0,\dots,n$ and every 
$S\subseteq N$ with $|S|=l$,
$
|U(S)-V(S)| \le g(l).
$ 
Then for every $i\in N$,
\begin{equation*}
    |\phi_i(U)-\phi_i(V)|
    \le
    \frac{1}{n}
    \Bigl(
    g(0)+g(n)+2\sum_{l=1}^{n-1} g(l)
    \Bigr).
\end{equation*}
\end{lemma}
The above result follows directly from the size-grouped representation of the Shapley value~\eqref{eq:shapley_subset} together with the additivity axiom of the Shapley value. Moreover, if $g(0)=0$, we obtain the following simplified bound:
$
|\phi_i(U)-\phi_i(V)|
\le
\frac{2}{n}
\sum_{l=1}^{n} g(l).
$
In particular, if the perturbation bound does not depend on the coalition size, 
that is, if there exists a constant $\varepsilon \ge 0$ such that 
$|U(S)-V(S)| \le \varepsilon$ for all $S \subseteq N$, 
then taking $g(l) \equiv \varepsilon$ yields
$
|\phi_i(U)-\phi_i(V)|
\le
2\varepsilon,
$
which coincides with the bound obtained by directly controlling the supremum $\sup_{S \subseteq N} |U(S)-V(S)|$.

\subsection{Proof of Proposition~\ref{prop:proposition_gf_shap_stab}}
\label{proof:Proof_of_Proposition_1}

\begin{proof}
Fix a nonempty coalition \(S\subseteq N\).
Let \(f_{t,\rho}^{S}\in\mathcal H_k\) denote the zero-initialized
gradient-flow solution associated with \(D_S\), and let
\(f_{\infty,\rho}^{S}\) denote its stationary limit. By the definitions
in Appendix~\ref{appendix:Solutions by kernel regression},
\(\mathcal P_{t,\rho}^{S}(K_{NN})
=
f_{t,\rho}^{S}(X_T)\) and
\(\mathcal P_{\rho}^{S}(K_{NN})
=
f_{\infty,\rho}^{S}(X_T)\).

For every \(f\in\mathcal H_k\), the reproducing property and the
definition of \(\kappa\) give
\[
\lVert f(X_T)\rVert_2^2
=
\sum_{i\in T}|f(x_i)|^2
\leq
\lVert f\rVert_{\mathcal H_k}^2
\operatorname{tr}(K_{TT})
\leq
\kappa^2|T|
\lVert f\rVert_{\mathcal H_k}^2.
\]
By Proposition~2 of \citet{haas2023mind},
\(\lVert f_{t,\rho}^{S}\rVert_{\mathcal H_k}
\leq
\lVert f_{\infty,\rho}^{S}\rVert_{\mathcal H_k}
\leq
\lVert f_{\infty,0}^{S}\rVert_{\mathcal H_k}
\).
Since \(f_{\infty,\rho}^{S}\) minimizes the regularized risk in
Equation~\eqref{eq:L_risk_object}, comparison with the zero function
gives
\(\rho\lVert f_{\infty,\rho}^{S}\rVert_{\mathcal H_k}^2
\leq
\mathcal L_{\rho}^{S}(f_{\infty,\rho}^{S})
\leq
\mathcal L_{\rho}^{S}(0)
=
|S|^{-1}\lVert Y_S\rVert_2^2
\leq 1\).
Consequently,
\(\lVert f_{t,\rho}^{S}\rVert_{\mathcal H_k}
\leq
\lVert f_{\infty,\rho}^{S}\rVert_{\mathcal H_k}
\leq
\rho^{-1/2}\).

Moreover, the spectral decay estimate in
Equation~\eqref{eq:rho>0 spectral bound} and the bound
\(\lVert Y_S\rVert_2\leq\sqrt{|S|}\) imply
\(\lVert f_{t,\rho}^{S}-f_{\infty,\rho}^{S}\rVert_{\mathcal H_k}
\leq
\rho^{-1/2}e^{-2\rho t}\).
Therefore,
\[
\begin{aligned}
&
\left|
U_{t,\rho}^{K_{NN}}(S)
-
U_{\infty,\rho}^{K_{NN}}(S)
\right|
\\
&\leq
\frac{1}{|T|}
\left(
\left\lVert
Y_T-\mathcal P_{t,\rho}^{S}(K_{NN})
\right\rVert_2
+
\left\lVert
Y_T-\mathcal P_{\rho}^{S}(K_{NN})
\right\rVert_2
\right)
\left\lVert
\mathcal P_{t,\rho}^{S}(K_{NN})
-
\mathcal P_{\rho}^{S}(K_{NN})
\right\rVert_2
\\
&\leq
2\kappa
\left(
1+
\kappa
\lVert f_{\infty,\rho}^{S}\rVert_{\mathcal H_k}
\right)
\left\lVert
f_{t,\rho}^{S}
-
f_{\infty,\rho}^{S}
\right\rVert_{\mathcal H_k}
\\
&\leq
\frac{2\kappa}{\sqrt{\rho}}
\left(
1+\frac{\kappa}{\sqrt{\rho}}
\right)
e^{-2\rho t}
=:
\varepsilon_t.
\end{aligned}
\]
Together with the standing convention for the empty coalition, the
preceding estimate shows that the utility difference is
bounded by \(g(|S|)\), where \(g(0):=0\) and
\(g(\ell):=\varepsilon_t\) for \(1\leq\ell\leq|N|\).
Applying
Lemma~\ref{lemma: Size-dependent perturbation bound for Shapley value}
gives
\(\left|
\phi_i(U_{t,\rho}^{K_{NN}})
-
\phi_i(U_{\infty,\rho}^{K_{NN}})
\right|
\leq
2\varepsilon_t\)
for every \(i\in N\).
Taking the maximum over \(i\in N\), we obtain
\[
\left\lVert
\mathcal S_{t,\rho}(K_{NN})
-
\mathcal S_{\rho}(K_{NN})
\right\rVert_\infty
\leq
C_{\kappa,\rho}e^{-2\rho t},
\qquad
C_{\kappa,\rho}
:=
\frac{4\kappa}{\sqrt{\rho}}
\left(
1+\frac{\kappa}{\sqrt{\rho}}
\right).
\]
\end{proof}
\paragraph{Ridgeless setting.}
When \(\rho=0\), the preceding utility estimate together with
\eqref{eq:ridgeless_rkhsnormbound} yields, for every nonempty coalition
\(S\subseteq N\) and every \(t>0\),
\begin{align*}
\left\lvert
U_{t,0}^{K_{NN}}(S)
-
U_{\infty,0}^{K_{NN}}(S)
\right\rvert
\le
\kappa
\left(
1+\kappa
\left\lVert f_{\infty,0}^{S}\right\rVert_{\mathcal H_k}
\right)
\frac{
\sqrt{|S|}\,\left\lVert Y_S\right\rVert_2
}{
\sqrt{t}\,
\lambda_{\min}^{+}(K_{SS})
}.
\end{align*}
However,
\(\lambda_{\min}^{+}(K_{SS})\) can be arbitrarily small as the dataset
and coalition vary. Consequently, because Data Shapley depends on all
coalitions \(S\subseteq N\), the coalition-wise estimate above does not
yield a uniform stability bound for the ridgeless
gradient flow without an additional spectral-gap assumption.
\subsection{Proof of Proposition~\ref{prop:low_rank_stab}}
\label{proof: Proof of Proposition 2}

\begin{proof}
Fix a nonempty coalition \(S\subseteq N\).
Recall that, for any training-matrix \(G\),
\(A_{S,\rho}(G)=G_{SS}+\rho|S|I_{|S|}\),
\(\mathcal P_\rho^S(G)
=
K_{TS}A_{S,\rho}(G)^{-1}Y_S\), and
\(U_{\infty,\rho}^{G}(S)
=
-|T|^{-1}
\lVert Y_T-\mathcal P_{\rho}^{S}(G)\rVert_2^2\).

Since \(K_{NN}\) and \(L_{NN}\) are positive semidefinite, for every
\(G\in\{K_{NN},L_{NN}\}\), we have
\(A_{S,\rho}(G)\succeq\rho|S|I_{|S|}\succ0\), and hence
\(\lVert A_{S,\rho}(G)^{-1}\rVert_2\leq(\rho|S|)^{-1}\).
By the definition of \(\kappa\),
\(\lVert K_{TS}\rVert_2
\leq\lVert K_{TS}\rVert_{\mathrm F}
\leq\kappa^2\sqrt{|T||S|}\).
Together with
\(\lVert Y_T\rVert_2\leq\sqrt{|T|}\) and
\(\lVert Y_S\rVert_2\leq\sqrt{|S|}\), this gives, for
\(G\in\{K_{NN},L_{NN}\}\),
\[
\left\lVert\mathcal P_\rho^S(G)\right\rVert_2
\leq
\frac{\kappa^2}{\rho}\sqrt{|T|},
\qquad
\left\lVert Y_T-\mathcal P_\rho^S(G)\right\rVert_2
\leq
\left(1+\frac{\kappa^2}{\rho}\right)\sqrt{|T|}.
\]
We first bound the prediction difference. Since \(K_{SS}-L_{SS}\) is a principal submatrix of
\(K_{NN}-L_{NN}\), we have
\(\lVert K_{SS}-L_{SS}\rVert_2
\leq
\lVert K_{NN}-L_{NN}\rVert_2\).
Combining this observation with the resolvent identity
\(A^{-1}-B^{-1}=A^{-1}(B-A)B^{-1}\) and the preceding bounds, we obtain
\[
\begin{aligned}
&
\left\lVert
\mathcal P_\rho^S(K_{NN})
-
\mathcal P_\rho^S(L_{NN})
\right\rVert_2
\\
&\leq
\lVert K_{TS}\rVert_2
\lVert A_{S,\rho}(K_{NN})^{-1}\rVert_2
\lVert K_{SS}-L_{SS}\rVert_2
\lVert A_{S,\rho}(L_{NN})^{-1}\rVert_2
\lVert Y_S\rVert_2
\\
&\leq
\frac{\kappa^2\sqrt{|T|}}{\rho^2|S|}
\lVert K_{NN}-L_{NN}\rVert_2.
\end{aligned}
\]
We next bound the corresponding utility difference. Applying
\(\lvert\lVert a\rVert_2^2-\lVert b\rVert_2^2\rvert
\leq
(\lVert a\rVert_2+\lVert b\rVert_2)\lVert a-b\rVert_2\)
and using the preceding prediction and residual bounds yields
\[
\begin{aligned}
&
\left|
U_{\infty,\rho}^{K_{NN}}(S)
-
U_{\infty,\rho}^{L_{NN}}(S)
\right|
\\
&\leq
\frac{1}{|T|}
\left(
\left\lVert
Y_T-\mathcal P_\rho^S(K_{NN})
\right\rVert_2
+
\left\lVert
Y_T-\mathcal P_\rho^S(L_{NN})
\right\rVert_2
\right)
\left\lVert
\mathcal P_\rho^S(K_{NN})
-
\mathcal P_\rho^S(L_{NN})
\right\rVert_2
\\
&\leq
2\kappa^2\rho^{-2}
\left(1+\frac{\kappa^2}{\rho}\right)
\frac{\lVert K_{NN}-L_{NN}\rVert_2}{|S|}.
\end{aligned}
\]
For \(S=\emptyset\), the utility difference is zero by the standing
convention. Applying
Lemma~\ref{lemma: Size-dependent perturbation bound for Shapley value}
to the preceding estimate and using
\(\sum_{\ell=1}^{|N|}\ell^{-1}\leq1+\log|N|\), we obtain, for every
\(i\in N\),
\[
\begin{aligned}
&
\left|
\phi_i\!\left(U_{\infty,\rho}^{K_{NN}}\right)
-
\phi_i\!\left(U_{\infty,\rho}^{L_{NN}}\right)
\right|
\\
&\leq
\frac{4\kappa^2}{\rho^2|N|}
\left(1+\frac{\kappa^2}{\rho}\right)
\lVert K_{NN}-L_{NN}\rVert_2
\sum_{\ell=1}^{|N|}\frac{1}{\ell}
\\
&\leq
4\kappa^2\rho^{-2}
\left(1+\frac{\kappa^2}{\rho}\right)
\frac{1+\log|N|}{|N|}
\lVert K_{NN}-L_{NN}\rVert_2.
\end{aligned}
\]
Taking the maximum over \(i\in N\) proves the claimed bound, with
\(\widetilde C_{\kappa,\rho}
=
4\kappa^2\rho^{-2}(1+\kappa^2/\rho)\).
\end{proof}

\paragraph{Scale invariance.}
The right-hand side of the bound in
Proposition~\ref{prop:low_rank_stab} is invariant under the simultaneous
rescaling \(K_{NN}\mapsto cK_{NN}\),
\(L_{NN}\mapsto cL_{NN}\), and \(\rho\mapsto c\rho\) for \(c>0\).
Indeed, the corresponding kernel rescaling sends \(\kappa\) to
\(\sqrt{c}\,\kappa\), so that
\(\widetilde C_{\sqrt{c}\kappa,c\rho}
=
c^{-1}\widetilde C_{\kappa,\rho}\), while
\(\lVert cK_{NN}-cL_{NN}\rVert_2
=
c\lVert K_{NN}-L_{NN}\rVert_2\).

\paragraph{Best rank-\(r\) approximation.}
For the best rank-\(r\) approximation \(K_{NN}^{\mathrm{eig}}\)
introduced in \eqref{eq:eig_approx}
\(\lVert K_{NN}-K_{NN}^{\mathrm{eig}}\rVert_2
=
\lambda_{r+1}(K_{NN})\). Hence, if
\(\lambda_j(K_{NN})\leq C_{\lambda}j^{-\alpha}\) for some
\(C_{\lambda}>0\) and \(\alpha>0\), Proposition~\ref{prop:low_rank_stab}
gives
\[
\begin{aligned}
\left\lVert
\mathcal S_\rho(K_{NN})
-
\mathcal S_\rho(K_{NN}^{\mathrm{eig}})
\right\rVert_\infty
\leq
\widetilde C_{\kappa,\rho}
\frac{1+\log|N|}{|N|}
\lambda_{r+1}(K_{NN})
\leq
C_{\lambda}\widetilde C_{\kappa,\rho}
\frac{1+\log|N|}{|N|}
(r+1)^{-\alpha}.
\end{aligned}
\]
By the efficiency axiom, the average signed Shapley value is of order
\(1/|N|\) whenever
\(U_{\infty,\rho}^{K_{NN}}(N)
-
U_{\infty,\rho}^{K_{NN}}(\emptyset)=O(1)\).
To make the preceding approximation error no larger than this scale, it
suffices to choose \(r\) such that
\[
r
\geq
\left[
C_{\lambda}\widetilde C_{\kappa,\rho}
(1+\log|N|)
\right]^{1/\alpha}
-1.
\]
Using the explicit expression for
\(\widetilde C_{\kappa,\rho}\), when
\(\rho\ll\kappa^2\), it suffices to take
\[
r
\gtrsim
\left(
C_{\lambda}\kappa^4\rho^{-3}
(1+\log|N|)
\right)^{1/\alpha}.
\]
Thus, faster eigenvalue decay reduces the sufficient rank, whereas
weaker regularization increases it.

\paragraph{Uniform Nystr\"om approximation.}
For the following analysis, we impose an additional sampling assumption
on the Nystr\"om approximation \(K_{NN}^{\mathrm{nys}}\) introduced in
\eqref{eq:nys_approx}: \(M\) is formed by sampling \(m\) landmark
indices independently and uniformly with replacement from \(N\), with
repeated indices retained, as in Theorem~1 of
\citet{cortes2010impact}. For \(r\leq\operatorname{rank}(K_{MM})\), let
\((K_{MM})_r\) be the best rank-\(r\) approximation of \(K_{MM}\) and
define the truncated approximation
\(K_{NN}^{\mathrm{nys},r}
:=
K_{NM}(K_{MM})_r^\dagger K_{MN}\).
Since
\(K_{NN}^{\mathrm{nys},r}
\preceq
K_{NN}^{\mathrm{nys}}
\preceq
K_{NN}\), Theorem~1 of \citet{cortes2010impact} gives, with probability
at least \(1-\delta\),
\[
\begin{aligned}
\left\lVert
K_{NN}-K_{NN}^{\mathrm{nys}}
\right\rVert_2
\leq
\left\lVert
K_{NN}-K_{NN}^{\mathrm{nys},r}
\right\rVert_2
\leq
\lambda_{r+1}(K_{NN})
+
\frac{|N|\kappa^2}{\sqrt m}
\left(
2+\log\frac{1}{\delta}
\right).
\end{aligned}
\]
Combining this estimate with Proposition~\ref{prop:low_rank_stab}
yields, with the same probability,
\[
\begin{aligned}
\left\lVert
\mathcal S_\rho(K_{NN})
-
\mathcal S_\rho(K_{NN}^{\mathrm{nys}})
\right\rVert_\infty
\leq
\widetilde C_{\kappa,\rho}
\left[
\frac{1+\log|N|}{|N|}
\lambda_{r+1}(K_{NN})
+
\frac{\kappa^2(1+\log|N|)}{\sqrt m}
\left(
2+\log\frac{1}{\delta}
\right)
\right].
\end{aligned}
\]
For fixed \(\kappa\), \(\rho\), and \(\delta\), the sampling contribution
in this upper bound is of order
\((1+\log|N|)/\sqrt m\). Requiring this contribution to be of order
\(1/|N|\) would require
\(m\gtrsim |N|^2(1+\log|N|)^2\), which is incompatible with the intended
low-rank regime \(m<|N|\). Thus, this particular uniform-sampling bound
does not certify individual Shapley accuracy at the \(1/|N|\) scale.

\subsection{Proof of Proposition~\ref{prop:prediction-shapley-separation}}
\label{proof: Proof of Proposition 3}

\paragraph{Open-domain extension of the prediction and Data Shapley maps.}
We use the scalar-response notation of
Section~\ref{sec:KPS_diagram}. Throughout this subsection,
\(Y_N\), \(Y_T\), \(K_{TN}\), and \(\rho>0\) are fixed, while \(G\)
varies in \(\operatorname{Sym}(|N|)\). Here,
\(\operatorname{Sym}(|N|)^+\) denotes the open cone of real symmetric
positive-definite matrices, and
\(\overline{\operatorname{Sym}(|N|)^+}\) denotes its closure, namely,
the cone of real symmetric positive-semidefinite matrices.
To differentiate the coalition prediction and Data Shapley maps,
define the open set
\(
\mathcal D_\rho
:=
\left\{
G\in\operatorname{Sym}(|N|):
G+\rho I_{|N|}
\in
\operatorname{Sym}(|N|)^+
\right\}.
\)
Since \(\mathcal D_\rho\) and \(\operatorname{Sym}(|N|)^+\) are open
subsets of \(\operatorname{Sym}(|N|)\), we have
\[
T_G\mathcal D_\rho
=
\operatorname{Sym}(|N|),
\qquad
T_G\operatorname{Sym}(|N|)^+
=
\operatorname{Sym}(|N|).
\]
Moreover, since \(\rho>0\), every
\(G\in\overline{\operatorname{Sym}(|N|)^+}\) satisfies
\(G+\rho I_{|N|}\succ0\), and hence
\(\overline{\operatorname{Sym}(|N|)^+}\subseteq\mathcal D_\rho\).
Therefore, for every nonempty \(S\subseteq N\), the coalition
prediction map defined in Section~\ref{sec:KPS_diagram} extends to
\(\mathcal D_\rho\) by
\[
A_{S,\rho}(G)
:=
G_{SS}+\rho|S|I_{|S|},
\qquad
\mathcal P_\rho^S(G)
:=
K_{TS}A_{S,\rho}(G)^{-1}Y_S,
\]
with \(\mathcal P_\rho^\varnothing(G):=0\).
For every \(G\in\mathcal D_\rho\) and every nonempty
\(S\subseteq N\),
\[
A_{S,\rho}(G)
=
(G+\rho I_{|N|})_{SS}
+
\rho(|S|-1)I_{|S|}
\succ0,
\]
because every principal submatrix of
\(G+\rho I_{|N|}\succ0\) is positive definite. Hence
\(A_{S,\rho}(G)^{-1}\) exists, and the extended coalition prediction
map is well-defined.
Using these prediction maps, define
\[
\mathcal P_\rho(G)
:=
\mathcal P_\rho^N(G),
\qquad
U_{\infty,\rho}^G(S)
:=
-\frac{1}{|T|}
\left\|
Y_T-\mathcal P_\rho^S(G)
\right\|_2^2,
\qquad
\mathcal S_\rho(G)
:=
\bigl(\phi_i(U_\rho^G)\bigr)_{i\in N}.
\]
These definitions agree with the corresponding maps from
Section~\ref{sec:KPS_diagram} on
\(\overline{\operatorname{Sym}(|N|)^+}\). Matrices in
\(\mathcal D_\rho\setminus
\overline{\operatorname{Sym}(|N|)^+}\)
are used only to provide an open extension on which Fréchet
differentiation can be performed. The differentials computed below have the types
\[
D\mathcal P_\rho(G):
\operatorname{Sym}(|N|)
\longrightarrow
\mathbb R^{|T|},
\qquad
D\mathcal S_\rho(G):
\operatorname{Sym}(|N|)
\longrightarrow
\mathbb R^{|N|}.
\]

\begin{lemma}[Smoothness of prediction and Data Shapley]
\label{lem:prediction-shapley-smoothness}
For every nonempty \(S\subseteq N\), the maps
\(G\mapsto\mathcal P_\rho^S(G)\) and
\(G\mapsto U_\rho^G(S)\) are \(C^\infty\) on
\(\mathcal D_\rho\). More explicitly, for \(G\in\mathcal D_\rho\),
\(H\in T_G\mathcal D_\rho=\operatorname{Sym}(|N|)\), and every
nonempty \(S\subseteq N\), one has
\[
\begin{aligned}
D\mathcal P_\rho^S(G)[H]
&=
-
K_{TS}
A_{S,\rho}(G)^{-1}
H_{SS}
A_{S,\rho}(G)^{-1}
Y_S,\\
D_GU_{\infty,\rho}^G(S)[H]
&=
-\frac{2}{|T|}
\left\langle
Y_T-\mathcal P_\rho^S(G),
K_{TS}
A_{S,\rho}(G)^{-1}
H_{SS}
A_{S,\rho}(G)^{-1}
Y_S
\right\rangle.
\end{aligned}
\]
Consequently, \(\mathcal P_\rho\) and \(\mathcal S_\rho\) are
\(C^\infty\) on \(\mathcal D_\rho\).
\end{lemma}
\begin{proof}
For every nonempty \(S\subseteq N\), the map
\(G\mapsto A_{S,\rho}(G)\) is affine with derivative
\(H\mapsto H_{SS}\), and it takes values in
\(\operatorname{Sym}(|S|)^+\). Matrix inversion is \(C^\infty\) on
the invertible matrices and satisfies
\(D(A^{-1})[B]=-A^{-1}BA^{-1}\). The first formula therefore follows
from the chain rule.

Applying the chain rule to
\(U_{\infty,\rho}^G(S)
=-|T|^{-1}\lVert Y_T-\mathcal P_\rho^S(G)\rVert_2^2\)
gives
\(D_GU_{\infty,\rho}^G(S)[H]
=2|T|^{-1}
\langle Y_T-\mathcal P_\rho^S(G),
D\mathcal P_\rho^S(G)[H]\rangle\).
Substituting the prediction derivative gives the second formula.
Finally, \(\mathcal P_\rho=\mathcal P_\rho^N\), and every component
of \(\mathcal S_\rho\) is a finite linear combination of coalition
utilities. Hence both maps are \(C^\infty\).
\end{proof}

\begin{lemma}[Constant-rank geometry of the prediction fibers]
\label{lem:constant-rank-prediction-fiber}
Define
\(\Lambda:\operatorname{Sym}(|N|)\to\mathbb R^{|T|}\) by
\(\Lambda(B):=K_{TN}BY_N\). Then
\[
\operatorname{rank}D\mathcal P_\rho(G)
=
\operatorname{rank}\Lambda
\qquad
\text{for every }G\in\mathcal D_\rho.
\]
Consequently, \(\mathcal P_\rho\) has constant rank on
\(\mathcal D_\rho\).
For every \(G_0\in\operatorname{Sym}(|N|)^+\), the prediction fiber
\[
\mathcal M_0
:=
\left\{
G\in\operatorname{Sym}(|N|)^+:
\mathcal P_\rho(G)=\mathcal P_\rho(G_0)
\right\}
\]
is a properly embedded smooth submanifold of
\(\operatorname{Sym}(|N|)^+\), and
\[
T_{G_0}\mathcal M_0
=
\ker D\mathcal P_\rho(G_0)
\subseteq
T_{G_0}\operatorname{Sym}(|N|)^+
=
\operatorname{Sym}(|N|).
\]
Consequently, for every
\(H\in\ker D\mathcal P_\rho(G_0)\), there exist
\(\varepsilon>0\) and a smooth curve
\(\gamma:(-\varepsilon,\varepsilon)\to\mathcal M_0\) satisfying
\(\gamma(0)=G_0\) and \(\gamma'(0)=H\).
\end{lemma}

\begin{proof}
For \(G\in\mathcal D_\rho\), set
\(Q_G:=(G+\rho|N|I_{|N|})^{-1}\) and define
\(\Gamma_G:\operatorname{Sym}(|N|)
\to\operatorname{Sym}(|N|)\) by
\(\Gamma_G(H):=Q_GHQ_G\).
Since \(Q_G\) is symmetric and invertible, \(\Gamma_G\) is a linear
automorphism with inverse
\(\Gamma_G^{-1}(B)=Q_G^{-1}BQ_G^{-1}\). Moreover, we have
\(
D\mathcal P_\rho(G)
=
-\Lambda\circ\Gamma_G.
\)
Composition with a linear automorphism does not change rank, proving
the constant-rank assertion.

Since \(\operatorname{Sym}(|N|)^+\) is open in
\(\operatorname{Sym}(|N|)\), the differential of the restriction
\(\mathcal P_\rho|_{\operatorname{Sym}(|N|)^+}\) agrees with
\(D\mathcal P_\rho(G)\) under the identification
\(T_G\operatorname{Sym}(|N|)^+
=\operatorname{Sym}(|N|)\). Hence this restriction also has constant
rank. By the constant-rank level-set theorem of
\citet[Theorem~5.12]{lee2013smooth},
\(\mathcal M_0\) is a properly embedded submanifold of
\(\operatorname{Sym}(|N|)^+\), whose codimension equals
\(\operatorname{rank}\Lambda\).

Let \(\iota:\mathcal M_0\hookrightarrow
\operatorname{Sym}(|N|)^+\) denote the inclusion. Since
\(\mathcal P_\rho\circ\iota\) is constant, the chain rule gives
\(T_{G_0}\mathcal M_0
\subseteq\ker D\mathcal P_\rho(G_0)\).
The constant-rank level-set theorem and rank--nullity give
\(\dim T_{G_0}\mathcal M_0
=\dim\operatorname{Sym}(|N|)
-\operatorname{rank}\Lambda
=\dim\ker D\mathcal P_\rho(G_0)\).
The inclusion is therefore an equality. The curve assertion follows
from the standard characterization of tangent vectors as velocities
of smooth curves.
\end{proof}

We now combine the preceding two lemmas to prove
Proposition~\ref{prop:prediction-shapley-separation}.

\begin{proof}
Suppose, toward a contradiction, that there exist a neighborhood
\(\mathcal V\) of \(G_0\) in \(\operatorname{Sym}(|N|)^+\) and a
function \(\omega:[0,\infty)\to[0,\infty)\) satisfying
\(\omega(0)=0\) such that the asserted bound holds for every
\(K,L\in\mathcal V\).

The kernel condition gives
\(H\in T_{G_0}\operatorname{Sym}(|N|)^+
=\operatorname{Sym}(|N|)\) such that
\(D\mathcal P_\rho(G_0)[H]=0\) but
\(D\mathcal S_\rho(G_0)[H]\neq0\).
By Lemma~\ref{lem:constant-rank-prediction-fiber}, there exist
\(\varepsilon>0\) and a smooth curve
\(\gamma:(-\varepsilon,\varepsilon)\to\mathcal M_0\) such that
\(\gamma(0)=G_0\), \(\gamma'(0)=H\), and
\(
\mathcal P_\rho(\gamma(t))
=
\mathcal P_\rho(G_0)
\)
for every \(t\in(-\varepsilon,\varepsilon)\).

Fréchet differentiability of \(\mathcal S_\rho\) gives
\[
\mathcal S_\rho(\gamma(t))
=
\mathcal S_\rho(G_0)
+
tD\mathcal S_\rho(G_0)[H]
+
o(|t|).
\]
Since \(D\mathcal S_\rho(G_0)[H]\neq0\), this expansion implies
\(
\mathcal S_\rho(\gamma(t))
\neq
\mathcal S_\rho(G_0)
\)
for every sufficiently small nonzero \(t\).
Since \(\gamma(t)\to G_0\) and \(\mathcal V\) is a neighborhood of
\(G_0\), we may choose such a \(t\neq0\) with
\(\gamma(t)\in\mathcal V\). Taking
$
K=G_0$ and $
L=\gamma(t)$
in the assumed bound gives
\[
\begin{aligned}
0
<
\left\|
\mathcal S_\rho(\gamma(t))
-
\mathcal S_\rho(G_0)
\right\|_2
\leq
\omega\left(
\left\|
\mathcal P_\rho(\gamma(t))
-
\mathcal P_\rho(G_0)
\right\|_2
\right)
=
\omega(0)
=
0,
\end{aligned}
\]
which is impossible.
\end{proof}

\begin{remark}[Equivalent Jacobian rank condition]
Let \(d:=|N|(|N|+1)/2\), and let
\(\operatorname{svec}:\operatorname{Sym}(|N|)\to\mathbb R^d\)
denote the linear isomorphism obtained by stacking the upper-triangular
entries, including the diagonal, of a symmetric matrix in a fixed order.
Let
\(J_{\mathcal P_\rho}(G_0)\in\mathbb R^{|T|\times d}\) and
\(J_{\mathcal S_\rho}(G_0)\in\mathbb R^{|N|\times d}\)
be the corresponding Jacobian matrices. Then
\[
\ker D\mathcal P_\rho(G_0)
\not\subseteq
\ker D\mathcal S_\rho(G_0)
\quad\Longleftrightarrow\quad
\operatorname{rank}
\begin{bmatrix}
J_{\mathcal P_\rho}(G_0)\\
J_{\mathcal S_\rho}(G_0)
\end{bmatrix}
>
\operatorname{rank}J_{\mathcal P_\rho}(G_0).
\]
Thus the hypothesis of
Proposition~\ref{prop:prediction-shapley-separation} can be checked
through a Jacobian rank comparison without explicitly constructing
\(\ker D\mathcal P_\rho(G_0)\).
\end{remark}

\subsection{proof of Corollary~\ref{cor:nystrom_limitation}}
\label{proof: nystrom_limitation_cor}

The upper-bound analysis in Appendix~\ref{proof: Proof of Proposition 2} highlights limitations in certifying the accuracy of Data Shapley approximation via uniform Nyström sampling. \citet{wang2013improving} establish lower bounds on Nyström matrix approximation errors through explicit constructions. Motivated by their approach, we present an explicit example demonstrating how uniform Nyström sampling can fail to preserve individual Data Shapley values.

Fix \(\rho>0\). For all sufficiently large \(|N|\), partition
\(N=A\sqcup B\), where
\(|B|:=2\lfloor |N|^{1/3}/2\rfloor\) and
\(|A|:=|N|-|B|\). Further partition \(B=B_+\sqcup B_-\), where
\(|B_+|=|B_-|=|B|/2\). Let
\(J_k:=\mathbf 1_k\mathbf 1_k^\top\), and define
\[
K_{NN}
:=
\begin{pmatrix}
J_{|A|} & 0\\
0 & J_{|B|}
\end{pmatrix},
\qquad
K_{TN}
:=
\begin{pmatrix}
\mathbf 0_{|A|}^\top & \mathbf 1_{|B|}^\top\\
\mathbf 0_{|A|}^\top & -\mathbf 1_{|B|}^\top
\end{pmatrix}.
\]
Set \(y_i=0\) for \(i\in A\), \(y_i=1\) for \(i\in B_+\), and
\(y_i=-1\) for \(i\in B_-\), and let \(Y_T=(1,1)^\top\). Thus,
\(\sum_{i\in B}y_i=0\) and \(|y_i|\leq1\) for every \(i\in N\).
Let \(K_{NN}^{\mathrm{eig},2}\) denote the eigen approximation
in \eqref{eq:eig_approx} specialized to \(r=2\). Let
\(K_{NN}^{\mathrm{nys}}\) be the Nystr\"om approximation in
\eqref{eq:nys_approx}, using \(|B|\) landmark indices sampled
independently and uniformly with replacement from \(N\). 
Throughout, \(K_{TN}\), \(Y_N\), and \(Y_T\) are held fixed. Thus,
for
\(G\in
\{K_{NN},K_{NN}^{\mathrm{eig},2},K_{NN}^{\mathrm{nys}}\}\),
the quantities \(\mathcal P_\rho(G)\) and \(\mathcal S_\rho(G)\)
vary only through the training--training kernel block \(G\).

This construction is realized by a bounded two-cluster feature map.
Indeed, for orthonormal vectors \(e_A,e_B\), take
\(\psi(x_i)=e_A\) for \(i\in A\),
\(\psi(x_i)=e_B\) for \(i\in B\),
\(\psi(x_{t_+})=e_B\), and
\(\psi(x_{t_-})=-e_B\). Hence the bounded-kernel assumption holds
uniformly with \(\kappa=1\).
Moreover, each admissible training block \(G\) above is positive
semidefinite. Therefore, every nonempty coalition \(S\subseteq N\)
satisfies
\[
G_{SS}+\rho|S|I_{|S|}
\succeq
\rho|S|I_{|S|}
\succ0.
\]
\begin{proposition}[Exact eigen recovery and uniform Nystr\"om failure]
\label{prop:fixed-cross-exact-rank-counterexample}
Under the preceding setting, the rank-two eigen approximation recovers
\(K_{NN}\) exactly and hence preserves its Data Shapley values:
\[
K_{NN}^{\mathrm{eig},2}
=
K_{NN},
\qquad
\mathcal S_\rho(K_{NN}^{\mathrm{eig},2})
=
\mathcal S_\rho(K_{NN}).
\]
For every \(i\in B\),
\(
\left|
[\mathcal S_\rho(K_{NN})]_i
\right|
=
\Theta_\rho(|N|^{-1}).
\)
Moreover, with probability \(1-O(|N|^{-1/3})\), the Nystr\"om
approximation preserves the grand-coalition prediction,
\(
\mathcal P_\rho(K_{NN}^{\mathrm{nys}})
=
\mathcal P_\rho(K_{NN})
=
0,
\)
but, simultaneously for every \(i\in B\),
\[
\left|
[\mathcal S_\rho(K_{NN}^{\mathrm{nys}})]_i
-
[\mathcal S_\rho(K_{NN})]_i
\right|
=
\Theta_\rho(|N|^{-1}).
\]
Thus, the rank-two eigen approximation preserves
\(\mathcal S_\rho(K_{NN})\) exactly, whereas, with high probability,
uniform Nystr\"om preserves the grand-coalition prediction but incurs
a Shapley error on \(B\) of the same order as the corresponding
coordinates of \(\mathcal S_\rho(K_{NN})\).
\end{proposition}

\begin{proof}
Define \(u_A:=\mathbf 1_A/\sqrt{|A|}\) and
\(u_B:=\mathbf 1_B/\sqrt{|B|}\). Since \(A\cap B=\emptyset\), these
vectors are orthonormal, and
\[
K_{NN}
=
|A|u_Au_A^\top+|B|u_Bu_B^\top.
\]
This is the eigendecomposition of \(K_{NN}\). Its only nonzero
eigenvalues are \(|A|\) and \(|B|\), while every vector orthogonal to
\(u_A\) and \(u_B\) belongs to the kernel of \(K_{NN}\). Hence
\(K_{NN}\) has rank two and
\(K_{NN}^{\mathrm{eig},2}=K_{NN}\).
The rows of \(K_{TN}\) are \(\mathbf 1_B^\top\) and
\(-\mathbf 1_B^\top\), and therefore belong to
\(\operatorname{span}\{u_A,u_B\}\). Thus, if \(U_2\) contains the two
nonzero eigenvectors, then \(K_{TN}U_2U_2^\top=K_{TN}\).
Consequently, the rank-two eigen construction preserves both
\(K_{NN}\) and \(K_{TN}\) exactly.

Now consider the event \(\mathcal E_N:=\{M\subseteq A\}\), where
multiplicities are retained. Since the \(|M|=|B|\) landmarks are
sampled independently and uniformly with replacement,
\[
\mathbb P(\mathcal E_N)
=
\left(1-\frac{|B|}{|N|}\right)^{|B|}
\geq
1-\frac{|B|^2}{|N|}
=
1-O(|N|^{-1/3}).
\]
On \(\mathcal E_N\), every sampled column of \(K_{NN}\) equals
\(\mathbf 1_A\). Hence
\(K_{NM}=\mathbf 1_A\mathbf 1_{|M|}^\top\),
\(K_{MM}=J_{|M|}\), and
\(J_{|M|}^{\dagger}=|M|^{-2}J_{|M|}\). It follows that
\[
K_{NN}^{\mathrm{nys}}
=
\mathbf 1_A\mathbf 1_A^\top
=
\begin{pmatrix}
J_{|A|}&0\\
0&0_{|B|}
\end{pmatrix},
\qquad
\left\|K_{NN}-K_{NN}^{\mathrm{nys}}\right\|_2
=
|B|
=
\Theta(|N|^{1/3}).
\]

Fix a nonempty coalition \(S\subseteq N\), order its indices so that
those in \(S\cap A\) precede those in \(S\cap B\), and define
\(z_S:=\sum_{j\in S\cap B}y_j\). Under this ordering,
\[
K_{SS}
=
\begin{pmatrix}
J_{|S\cap A|}&0\\
0&J_{|S\cap B|}
\end{pmatrix},
\quad
Y_S
=
\begin{pmatrix}
\mathbf 0_{|S\cap A|}\\
Y_{S\cap B}
\end{pmatrix},
\quad
K_{TS}
=
\begin{pmatrix}
\mathbf 0^\top&\mathbf 1^\top\\
\mathbf 0^\top&-\mathbf 1^\top
\end{pmatrix}.
\]
The regularized coalition matrix is block diagonal. Since \(Y_S\)
vanishes on \(S\cap A\), only its \(B\)-block contributes to the
prediction. If \(S\cap B\neq\emptyset\), then
\(\mathbf 1_{|S\cap B|}^\top\) is a left eigenvector of
\(J_{|S\cap B|}+\rho|S|I_{|S\cap B|}\) with eigenvalue
\(|S\cap B|+\rho|S|\). Therefore,
\(\mathbf 1_{|S\cap B|}^\top
(J_{|S\cap B|}+\rho|S|I_{|S\cap B|})^{-1}
=
\mathbf 1_{|S\cap B|}^\top/(|S\cap B|+\rho|S|)\).
Using
\(\mathbf 1_{|S\cap B|}^\top Y_{S\cap B}=z_S\), we obtain
\[
\mathcal P_\rho^S(K_{NN})
=
\begin{pmatrix}p_S\\-p_S\end{pmatrix},
\quad
p_S
=
\frac{z_S}{|S\cap B|+\rho|S|},
\quad
\mathcal P_\rho^S(K_{NN}^{\mathrm{nys}})
=
\begin{pmatrix}q_S\\-q_S\end{pmatrix},
\quad
q_S
=
\frac{z_S}{\rho|S|}.
\]
Indeed, on \(\mathcal E_N\), the \(B\)-block of
\((K_{NN}^{\mathrm{nys}})_{SS}\) is zero, so its regularized
\(B\)-block is \(\rho|S|I_{|S\cap B|}\). Here \(K_{TS}\) is retained
without approximation. If \(S\cap B=\emptyset\), then \(z_S=0\), so
both formulas still give the zero prediction.
Since \(Y_T=(1,1)^\top\) and \(|T|=2\), the two utilities are
\(U_{\infty,\rho}^{K_{NN}}(S)=-1-p_S^2\) and
\(U_{\infty,\rho}^{K_{NN}^{\mathrm{nys}}}(S)=-1-q_S^2\).
The balanced-label condition gives \(z_N=0\). Hence
\[
\mathcal P_\rho(K_{NN})
=
\mathcal P_\rho(K_{NN}^{\mathrm{nys}})
=
0,
\qquad
U_{\infty,\rho}^{K_{NN}}(N)
=
U_{\infty,\rho}^{K_{NN}^{\mathrm{nys}}}(N)
=
U_{\infty,\rho}^{K_{NN}}(\emptyset)
=
U_{\infty,\rho}^{K_{NN}^{\mathrm{nys}}}(\emptyset)
=
-1.
\]
Fix \(i\in A\), and let \(S\subseteq N\setminus\{i\}\) be its
predecessor coalition. Define
\(\Delta_i^K(S):=
U_{\infty,\rho}^{K_{NN}}(S\cup\{i\})
-U_{\infty,\rho}^{K_{NN}}(S)\) and
\(\Delta_i^{\mathrm{nys}}(S):=
U_{\infty,\rho}^{K_{NN}^{\mathrm{nys}}}(S\cup\{i\})
-U_{\infty,\rho}^{K_{NN}^{\mathrm{nys}}}(S)\).
If \(S=\emptyset\), then both marginal contributions are zero.

Suppose that \(S\neq\emptyset\). Adding \(i\in A\) leaves \(z_S\) and
\(|S\cap B|\) unchanged and replaces \(|S|\) by \(|S|+1\). Define
\(g_\rho(u):=u^{-2}-(u+\rho)^{-2}\) for \(u>0\). Then
\[
\begin{aligned}
\Delta_i^K(S)
&=
z_S^2
\left[
\frac{1}{\left(|S\cap B|+\rho|S|\right)^2}
-
\frac{1}{\left(|S\cap B|+\rho|S|+\rho\right)^2}
\right]
=
z_S^2g_\rho\!\left(|S\cap B|+\rho|S|\right),\\
\Delta_i^{\mathrm{nys}}(S)
&=
z_S^2
\left[
\frac{1}{(\rho|S|)^2}
-
\frac{1}{(\rho|S|+\rho)^2}
\right]
=
z_S^2g_\rho(\rho|S|),\\
\Delta_i^{\mathrm{nys}}(S)-\Delta_i^K(S)
&=
z_S^2
\left[
g_\rho(\rho|S|)
-
g_\rho\!\left(|S\cap B|+\rho|S|\right)
\right]
\geq0.
\end{aligned}
\]
The last inequality follows because
\(g_\rho'(u)=-2u^{-3}+2(u+\rho)^{-3}<0\).

Let \(\delta_A\) denote the common Nystr\"om-minus-exact Shapley
difference for samples in \(A\). In the permutation representation,
the event that \(S\) consists of one sample from \(B\) has probability
\(|B|/[|N|(|N|-1)]\). On this event,
\(|S|=|S\cap B|=1\) and \(z_S^2=1\), giving the required lower bound.

For the upper bound, conditional on \(|S|=k\), the set \(S\) is
uniformly distributed over the \(k\)-subsets of
\(N\setminus\{i\}\). Since \(i\in A\), we have
\(\sum_{j\in N\setminus\{i\}}y_j=0\),
\(\sum_{j\in N\setminus\{i\}}y_j^2=|B|\), and
\(\sum_{j\neq\ell}y_jy_\ell=-|B|\). Expanding \(z_S^2\) and using the
corresponding inclusion probabilities gives
\[
\mathbb E\!\left[z_S^2\,\middle|\,|S|=k\right]
=
\frac{|B|k(|N|-1-k)}
{(|N|-1)(|N|-2)}
\leq
\frac{|B|k}{|N|-2},
\qquad
g_\rho(\rho k)
\leq
\frac{2}{\rho^2k^3}.
\]
Since the predecessor cardinality is uniform on
\(\{0,\ldots,|N|-1\}\), these estimates imply
\[
\begin{aligned}
\frac{|B|}{|N|(|N|-1)}
\left[g_\rho(\rho)-g_\rho(1+\rho)\right]
\leq
\delta_A
\leq
\frac{2|B|}{\rho^2|N|(|N|-2)}
\sum_{k=1}^{|N|-1}\frac{1}{k^2}
\leq
\frac{\pi^2|B|}
{3\rho^2|N|(|N|-2)}.
\end{aligned}
\]
Therefore,
\(\delta_A=\Theta_\rho(|B|/|N|^2)\).

All samples in \(A\) have the same Shapley value. Samples in \(B\)
also have the same Shapley value: exchanging \(B_+\) and \(B_-\)
changes \(z_S\) to \(-z_S\), while both utilities depend on \(z_S\)
only through \(z_S^2\).

Let \(\delta_B\) denote the common Nystr\"om-minus-exact Shapley
difference for samples in \(B\). Since the two games have the same
empty- and grand-coalition utilities, Shapley efficiency gives
\[
\begin{gathered}
|A|\delta_A+|B|\delta_B=0,\\
\frac{|A|}{|N|(|N|-1)}
\left[g_\rho(\rho)-g_\rho(1+\rho)\right]
\leq
|\delta_B|
\leq
\frac{\pi^2|A|}
{3\rho^2|N|(|N|-2)},
\qquad
|\delta_B|=\Theta_\rho(|N|^{-1}).
\end{gathered}
\]

It remains to determine the exact Shapley values on \(B\). Let
\(\alpha_A\) denote the common exact Shapley value of the samples in
\(A\). Applying the same singleton lower bound and moment upper bound
to \(\Delta_i^K(S)\) gives
\[
\frac{|B|}{|N|(|N|-1)}g_\rho(1+\rho)
\leq
\alpha_A
\leq
\frac{\pi^2|B|}
{3\rho^2|N|(|N|-2)},
\qquad
\alpha_A
=
\Theta_\rho\!\left(\frac{|B|}{|N|^2}\right).
\]
Let \(\alpha_B\) denote the common exact Shapley value of the samples
in \(B\). Shapley efficiency for the exact game gives
\[
|A|\alpha_A+|B|\alpha_B=0,
\qquad
\alpha_B
=
-\frac{|A|}{|B|}\alpha_A
=
-\Theta_\rho(|N|^{-1}).
\]
Finally, \(K_{NN}^{\mathrm{eig},2}=K_{NN}\), so the eigen
approximation has zero Data Shapley error. On \(\mathcal E_N\),
\begin{align*}
[\mathcal S_\rho(K_{NN})]_i
=
-\Theta_\rho(|N|^{-1}),
\quad
\left|
[\mathcal S_\rho(K_{NN}^{\mathrm{nys}})]_i
-
[\mathcal S_\rho(K_{NN})]_i
\right|
=
\Theta_\rho(|N|^{-1})
\quad (i\in B).
\end{align*}
This completes the proof.
\end{proof}

\clearpage
\section{Empirical Analysis of the Gap Between
Proposition~\ref{prop:low_rank_stab} and the Implemented
Predictor}\label{app:theory_gap}

Proposition~\ref{prop:low_rank_stab} bounds the change of Data Shapley
values when the training kernel is replaced by an
approximation. The setting of the proposition is not the setting of the
predictor implemented in LRFShap. There are
three differences: (i) the proposition approximates only the training
block $K_{SS}$ and keeps the test--train block $K_{TS}$ exact, whereas
LRFShap approximates both blocks
through the low-rank features $\Phi$; (ii) the ridge in the
proposition is $\rho|S|$, which comes from a fixed
regularization coefficient under the averaged empirical loss,
whereas the implementation uses a coalition-independent ridge
$\gamma$; (iii) the utility in the proposition is the negative
squared test error of a scalar-response regression, whereas
the implementation uses test accuracy. We do not extend the
proposition to the implemented predictor here. Instead, we
introduce the three differences one at a time and measure how
the Data Shapley error changes at each step. The purpose is to
see which of the assumptions matters in practice, not to show
that the bound holds in the implemented setting.

\paragraph{Setup.}
We use the eNTK of \texttt{bert-base-uncased} on SST-2 and MR,
and draw a random subset of $n=500$ training samples and
$|T|=500$ test samples. We write $\phi$ for the exact Data
Shapley values of the full-kernel game and $\phi_r$ for those
of the game with the rank-$r$ eigen-truncation
$K^{\mathrm{eig}}_{NN}$, $r\in\{1,5,10,15,20,25,30\}\%$ of
$n$. Neither vector can be computed exactly, so we use Monte
Carlo estimates $\widehat\phi$ and $\widehat\phi_r$ over
$B=300$ random permutations. The same permutations are used
for both games and no truncation is applied, so the estimate
of the difference, $\widehat\phi_r-\widehat\phi$, is a paired
estimate: for each permutation we take the difference of the
two marginal contributions and average over permutations. This
estimate is unbiased for $\phi_r-\phi=\mathcal{S}(K^{\mathrm{
eig}}_{NN})-\mathcal{S}(K_{NN})$, the quantity bounded in
Proposition~\ref{prop:low_rank_stab}, and the pairing removes
the variation that the two games share. Monte Carlo
uncertainty from the finite number of permutations remains,
and the reported $\|\widehat\phi_r-\widehat\phi\|_\infty$ is
not an unbiased estimate of $\|\phi_r-\phi\|_\infty$. We
report $\|\widehat\phi_r-\widehat\phi\|_\infty$ and the
Spearman rank correlation between $\widehat\phi_r$ and
$\widehat\phi$ over the $n$ training samples. The
regularization parameter is $\rho=10^{-2}$ for
the size-scaled ridge. For the two variants with a fixed ridge
we report $\gamma=10^{-2}$ and $\gamma=10^{-1}$.
Starting from the setting of the proposition, we switch on the
three differences cumulatively
(Table~\ref{tab:theory_gap_variants}); the last variant is the
predictor used by LRFShap.

\begin{table}[!ht]
\centering
\small
\begin{tabular}{llll}
\toprule
Variant & Approximated blocks & Ridge & Utility \\
\midrule
Prop.~\ref{prop:low_rank_stab} setting & $K_{SS}$ & $\rho|S|$
& $-$MSE, $y\in\{-1,+1\}$ \\
$+$ approximate $K_{TS}$ & $K_{SS}$, $K_{TS}$ & $\rho|S|$ &
$-$MSE, $y\in\{-1,+1\}$ \\
$+$ fixed ridge & $K_{SS}$, $K_{TS}$ & $\gamma$ & $-$MSE,
$y\in\{-1,+1\}$ \\
$+$ accuracy utility (LRFShap) & $K_{SS}$, $K_{TS}$ & $\gamma$
& accuracy, one-hot $Y$ \\
\bottomrule
\end{tabular}
\caption{Variants used in Appendix~\ref{app:theory_gap}. Each
row adds one difference between the setting of
Proposition~\ref{prop:low_rank_stab} and the implemented
predictor; the last row is the predictor used by LRFShap.}
\label{tab:theory_gap_variants}
\end{table}

\begin{figure}[!ht]
\centering
\includegraphics[width=\textwidth]{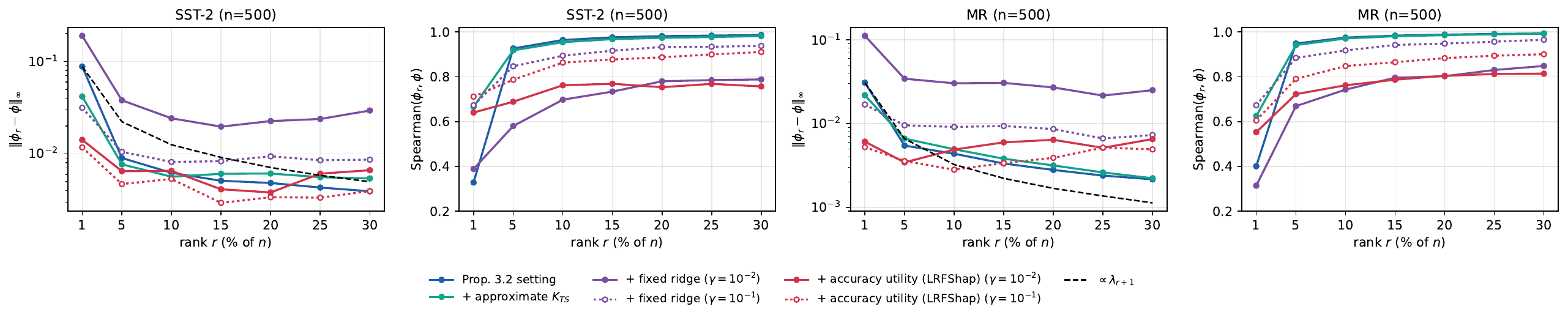}
\caption{Paired Monte Carlo estimate of the Data Shapley
difference between the rank-$r$ eigen-truncated game and the
full-kernel game, using the same $300$ permutations, on SST-2
(left two panels) and MR (right two panels) with $n=500$. Left
of each pair: $\|\widehat\phi_r-\widehat\phi\|_\infty$ on a
log scale. The dashed line is a reference curve proportional
to $\lambda_{r+1}$, scaled to the first point of the
Proposition~\ref{prop:low_rank_stab} setting; it shows the
shape of the spectral tail and is not a quantitative bound.
Right of each pair: Spearman rank correlation between
$\widehat\phi_r$ and $\widehat\phi$. For the
fixed-ridge variants, solid lines with filled markers use
$\gamma=10^{-2}$ and dotted
lines with open markers use $\gamma=10^{-1}$.}
\label{fig:theory_gap}
\end{figure}

\begin{table}[!ht]
\centering
\small
\setlength{\tabcolsep}{5pt}
\begin{tabular}{l cccc cccc}
\toprule
& \multicolumn{4}{c}{SST-2} & \multicolumn{4}{c}{MR} \\
\cmidrule(lr){2-5} \cmidrule(lr){6-9}
Variant & 5\% & 10\% & 20\% & 30\% & 5\% & 10\% & 20\% & 30\%
\\
\midrule
Prop.~\ref{prop:low_rank_stab} setting & 0.93 & 0.96 & 0.98 &
0.99 & 0.95 & 0.97 & 0.99 & 0.99 \\
$+$ approximate $K_{TS}$ & 0.92 & 0.95 & 0.97 & 0.98 & 0.94 &
0.97 & 0.99 & 0.99 \\
$+$ fixed ridge ($\gamma=10^{-2}$) & 0.58 & 0.70 & 0.78 &
0.79 & 0.67 & 0.74 &
0.80 & 0.85 \\
{$+$ fixed ridge ($\gamma=10^{-1}$)} &
{0.85} & {0.89} & {0.93} &
{0.94} & {0.88} & {0.92} &
{0.95} & {0.97} \\
$+$ accuracy utility (LRFShap) {($\gamma=10^{-2}$)} & 0.69
& 0.76 & 0.75 & 0.76 &
0.72 & 0.76 & 0.80 & 0.81 \\
{$+$ accuracy utility (LRFShap) ($\gamma=10^{-1}$)} &
{0.79} & {0.86} & {0.89} &
{0.91} &
{0.79} & {0.85} & {0.88} &
{0.90} \\
\bottomrule
\end{tabular}
\caption{Spearman rank correlation between $\widehat\phi_r$
and $\widehat\phi$ for each variant and rank ($\%$ of $n$),
$n=500$, $300$ shared permutations. Both vectors are Monte
Carlo estimates. For the fixed-ridge variants,
the value of $\gamma$ is given in parentheses.}
\label{tab:theory_gap_spearman}
\end{table}

\paragraph{Rank dependence in the setting of the proposition.}
In the setting of Proposition~\ref{prop:low_rank_stab}, the
estimated error decreases with the rank. Up to rank about
$10\%$ its decrease follows the shape of $\lambda_{r+1}$
(dashed line in Figure~\ref{fig:theory_gap}); beyond that the
curve flattens, which we attribute to the resolution of the
Monte Carlo estimate rather than to the approximation. The
rank correlation with the full-kernel estimate is $0.93$
(SST-2) and $0.95$ (MR) at rank $5\%$ and $0.99$ at rank
$30\%$ on both datasets. We note that this compares the
observed rank dependence with the decay of the spectral tail;
we did not evaluate the constant $\widetilde{C}_{\kappa,\rho}$
of the bound, so we make no claim about how tight the bound
is numerically.

\paragraph{Approximating the test--train block.}
Approximating $K_{TS}$ in addition to $K_{SS}$ changes the
reported quantities little on these two datasets. For ranks of
$5\%$ and above, the Spearman correlations of the two
variants differ by at most $0.02$ and the $\ell_\infty$ curves
are close; at rank $1\%$ the two $\ell_\infty$ errors do
differ visibly (Figure~\ref{fig:theory_gap}). A plausible
reason for the small effect at moderate ranks is that the
low-rank test--train block equals $K_{TN}U_rU_r^\top$
restricted to $S$, so its error is driven by the same tail of
the training spectrum. We only compare each variant against
the full-kernel estimate here; we did not measure the distance
between the two low-rank estimates themselves.

\paragraph{Fixed ridge.}
Replacing $\rho|S|$ by a fixed $\gamma$ is the one change with
a clear effect under the squared-error utility. With
$\gamma=10^{-2}$, the
$\ell_\infty$ error stops decreasing with $\lambda_{r+1}$, and
the rank correlation plateaus at $0.79$ (SST-2) and $0.85$
(MR). A fixed ridge corresponds to an effective coefficient
$\gamma/|S|$ under the averaged loss, which becomes small for
large coalitions; this is consistent with the constant
$\widetilde{C}_{\kappa,\rho}$ growing as $\rho$ decreases.
In line with this explanation, a larger fixed
ridge reduces the gap. With $\gamma=10^{-1}$, the rank
correlation rises to $0.94$ (SST-2) and $0.97$ (MR) at rank
$30\%$, close to the setting of the proposition, and the
$\ell_\infty$ error decreases by a factor of about two to four
at ranks of $5\%$ and above, although it does not follow the
decay of $\lambda_{r+1}$. These results suggest that, on these
two datasets, the gap in rank agreement observed with the
smaller fixed ridge is largely a matter of regularization
strength and can be substantially reduced through the choice
of $\gamma$.

\paragraph{Accuracy utility.}
Using accuracy as the utility, as in LRFShap, yields rank
correlations between the low-rank and full-kernel estimates
of $0.76$ (SST-2) and $0.81$ (MR) at rank $30\%$ for
$\gamma=10^{-2}$. The correlations remain roughly constant
at ranks of $10\%$ and above, while the $\ell_\infty$ error
is of a similar magnitude to that in the setting of the
proposition. With $\gamma=10^{-1}$, the rank correlation
continues to increase with the rank and reaches $0.91$
(SST-2) and $0.90$ (MR) at rank $30\%$, while the
$\ell_\infty$ error remains of a similar magnitude.
At the same fixed ridge, the rank correlations under
accuracy and negative squared-test-error utilities are
relatively close. These results suggest that, in the
experimental settings considered, rank agreement is more
sensitive to the choice of the fixed ridge parameter than
to the choice between these two utilities.

\paragraph{Summary.}
On two eNTK datasets, we went from the setting of
Proposition~\ref{prop:low_rank_stab} to the implemented
predictor in three steps. Approximating the test--train block
changed the reported fidelity measures little at moderate
ranks. Replacing the size-scaled ridge by a fixed one clearly
changed the sensitivity of the squared-error game to the
approximation. However, a larger fixed ridge brought the rank
agreement close to that observed in the setting of the
proposition. These results suggest that, when replacing
coalition-size-scaled regularization with a fixed ridge,
choosing an appropriate regularization strength can
substantially reduce the resulting loss of rank agreement.
With the accuracy utility, a larger fixed ridge likewise improved rank agreement, while switching between the two utilities at the same fixed ridge had a smaller effect.
This section examines the experimental role of the
Proposition~\ref{prop:low_rank_stab} assumptions; it does not extend the theoretical
guarantee to the implemented predictor.


\subsection{Implications for extending the theoretical analysis}
\label{app:theory_gap_extensions}

The three differences examined above have distinct implications
for extending Proposition~\ref{prop:low_rank_stab}.
Approximating the test--train block introduces an additional
perturbation term, changing the utility requires control of
classification margins, and changing the ridge rule affects
the dependence of the perturbation bounds on coalition size.

\paragraph{Approximating the test--train kernel block.}

Proposition~\ref{prop:low_rank_stab} keeps the test--train
block fixed, following the setting of
\citet{cortes2010impact}, to isolate the effect of
training-kernel approximation on individual Data Shapley
values. This choice does not preclude an analysis in which
both blocks are approximated.

For a nonempty coalition, define
\[
\begin{aligned}
A_S
&:=K_{SS}+\rho|S|I,
&
\widetilde A_S
&:=L_{SS}+\rho|S|I,
\\
F_S
&:=K_{TS}A_S^{-1}Y_S,
&
\widetilde F_S
&:=L_{TS}\widetilde A_S^{-1}Y_S.
\end{aligned}
\]
The inverse-difference identity gives
\[
\begin{aligned}
F_S-\widetilde F_S
={}&
(K_{TS}-L_{TS})\widetilde A_S^{-1}Y_S
+
K_{TS}A_S^{-1}
(L_{SS}-K_{SS})
\widetilde A_S^{-1}Y_S.
\end{aligned}
\]
The first term accounts for the test--train approximation,
while the second is the training-kernel perturbation term
used in the proof of Proposition~\ref{prop:low_rank_stab}.
Taking norms therefore yields a prediction-error bound
involving both approximation errors. The same decomposition
applies column by column when the responses are represented
by a one-hot label matrix.

Such an extension requires additional control of the
test--train approximation error. In particular, the new
term need not have the same dependence on coalition size
as the training-kernel term, so the sample-size dependence
of Proposition~\ref{prop:low_rank_stab} does not follow
automatically. The empirical comparisons above complement
this observation: they measure the fidelity of each variant
relative to its corresponding full-kernel reference, rather
than directly bounding the distance between the two
low-rank variants.

\paragraph{From squared-error utility to accuracy utility.}

Accuracy requires an additional argument because
classification correctness can change discontinuously
under perturbations of the prediction scores.
For multiclass score matrices, define the correct-class
margin and the maximum score error by
\[
\begin{aligned}
m_S(t)
:=
(F_S)_{t,c_t^\star}
-
\max_{c\ne c_t^\star}(F_S)_{t,c},
\qquad
\varepsilon_S
:=
\max_{t\in T,\;c\in C}
\left|(F_S-\widetilde F_S)_{t,c}\right|,
\end{aligned}
\]
where the starred class index denotes the correct class.
Define the proportion of test samples with small absolute
margins as
\[
Q_S(a)
:=
\frac{1}{|T|}
\sum_{t\in T}
\mathbf 1\{|m_S(t)|\le a\},
\qquad a\ge0.
\]
Since the maximum of the incorrect-class scores changes
by at most the maximum score error, we have
\[
|m_S(t)-\widetilde m_S(t)|
\le 2\varepsilon_S.
\]
Thus, classification correctness is preserved whenever
the original absolute margin exceeds twice the score
error. With a common deterministic tie-breaking rule,
this gives
\[
\left|
U_{\mathrm{acc}}(S)
-
\widetilde U_{\mathrm{acc}}(S)
\right|
\le Q_S(2\varepsilon_S).
\]

One sufficient additional assumption is a uniform linear
bound on the small-margin proportions:
\[
Q_S(a)\le C_{\mathrm{mar}}a,
\qquad
a\ge0,\quad
\varnothing\ne S\subseteq N,
\]
with a common finite positive constant. Under this condition,
\[
\left|
U_{\mathrm{acc}}(S)
-
\widetilde U_{\mathrm{acc}}(S)
\right|
\le 2C_{\mathrm{mar}}\varepsilon_S.
\]
Combining this inequality with a coalition-wise score-error
bound and the Shapley perturbation argument in
Lemma~\ref{lemma: Size-dependent perturbation bound for Shapley value}
yields a conditional accuracy-based Shapley bound, provided
that the two games have the same empty-coalition utility.

For exact Shapley values, the margin condition must hold
over all nonempty coalitions, not only the grand coalition.
Moreover, the resulting bound depends on the common margin
constant. The current kernel assumptions do not provide
quantitative control of this constant or establish the
uniform margin condition. Consequently, this argument
identifies sufficient additional assumptions without
establishing a theoretical guarantee for the
accuracy-based Shapley values computed by LRFShap.

\paragraph{Coalition-size-scaled ridge versus fixed ridge.}

The ridge rule changes the relative weight of the
average squared loss and the RKHS norm penalty.
The theoretical setting minimizes
\[
\frac{1}{|S|}
\sum_{i\in S}(f(x_i)-y_i)^2
+
\rho\|f\|_{\mathcal H_k}^2,
\]
which yields a matrix ridge proportional to coalition size.
In contrast, a fixed matrix ridge corresponds to
\[
\frac{1}{|S|}
\sum_{i\in S}(f(x_i)-y_i)^2
+
\frac{\gamma}{|S|}
\|f\|_{\mathcal H_k}^2.
\]
Thus, the theoretical setting keeps the regularization
weight relative to the average loss constant, whereas
the fixed-ridge setting reduces that weight as the
coalition grows.

This distinction enters the perturbation analysis through
the inverse bounds
\[
\left\|
(K_{SS}+\rho|S|I)^{-1}
\right\|_2
\le \frac{1}{\rho|S|},
\qquad
\left\|
(K_{SS}+\gamma I)^{-1}
\right\|_2
\le \frac{1}{\gamma}.
\]
The first bound is used in
Proposition~\ref{prop:low_rank_stab} to obtain prediction
and utility perturbation bounds that decrease inversely
with coalition size for a fixed training-kernel
approximation error. Aggregating these bounds over
coalition sizes gives the stated Shapley-error bound.
A positive fixed ridge also permits perturbation analysis,
but the second inverse bound does not provide the same
coalition-size decay. Consequently, analogous bounds
need not retain the sample-size dependence of
Proposition~\ref{prop:low_rank_stab}.

The implementation follows FreeShap in using a fixed
ridge so that the comparison evaluates low-rank
approximation under the same regularization setting.
The experiments above examine the regularization change
separately. In particular,
Table~\ref{tab:theory_gap_spearman} shows that increasing
the fixed ridge improves agreement with the corresponding
full-kernel Shapley estimates in these experiments.
This supports the role of regularization strength in
approximation stability, but does not establish that
coalition-size-scaled regularization is always preferable
to a fixed ridge.


\clearpage
\section{Computational complexity of FreeShap and LRFShap}
\label{app:complexity}

Table~\ref{tab:computation_complexity} summarizes the computational
costs of FreeShap and LRFShap. We denote the training-set size by
$|N|$, the coalition size by $|S|$, the number of classes by $|C|$,
the test-set size by $|T|$, and the number of Monte Carlo
permutations by $B$. The feature dimension is $q=r$ for eigen
features. For Nystr\"om features, $q$ is the number of retained
positive eigenvalues of the landmark matrix $W$, with $q\le m$,
where $m$ is the number of landmarks. We assume
$|C|\le q\le |N|$ when simplifying the bounds, and use the
coalition-independent ridge parameter $\gamma>0$.

Both implementations share the class-independent factorization
or inverse across all label columns. In particular, the FreeShap
baseline computes the factorization or inverse of
$K_{SS}+\gamma I_{|S|}$ once and reuses it for all $|C|$ classes.
LRFShap similarly shares its system matrix across classes.
Consequently, factorization and inverse-update costs do not
carry an additional factor of $|C|$. Label-dependent solves,
matrix products, and test predictions are counted separately.

The table reports linear-algebra costs once the required kernel
entries are available. The main text focuses on coalition-solve
costs, whereas the table also includes label-dependent
computations and test predictions.

\begin{table}[ht]
  \scriptsize
  \setlength{\tabcolsep}{3pt}
  \centering
  \caption{
    Computational complexity of FreeShap and LRFShap with
    factorization and inverse reuse across classes.
    Preparation costs are incurred once; other costs are per
    evaluated coalition. For direct solves, LRFShap uses the
    $q\times q$ feature-space system when $|S|\ge q$ and the
    $|S|\times |S|$ kernel-space system when $|S|<q$.
    Incremental inverse updates use blockwise inversion for
    FreeShap and Sherman--Morrison updates in feature space
    for LRFShap.
  }
  \label{tab:computation_complexity}
  \begin{tabular}{@{}lll@{}}
    \toprule
    Operation & FreeShap & LRFShap \\
    \midrule

    \multicolumn{3}{@{}l}{\emph{One-time preparation}} \\
    Eigen features
      & ---
      & $\mathcal O(|N|^3+|T||N|q)$ \\
    Nystr\"om features
      & ---
      & $\mathcal O(m^3+(|N|+|T|)mq)$ \\

    \midrule
    \multicolumn{3}{@{}l}{
      \emph{Per coalition, direct, $|S|\ge q$}
    } \\
    Matrix construction and factorization
      & $\mathcal O(|S|^3)$
      & $\mathcal O(|S|q^2+q^3)$ \\
    Label solves and predictions
      & $\mathcal O(|C||S|(|S|+|T|))$
      & $\mathcal O(|C|q(|S|+q+|T|))$ \\
    Total
      & $\mathcal O(|S|^3+|C||S|(|S|+|T|))$
      & $\mathcal O(|S|q^2+q^3+|C||T|q)$ \\

    \midrule
    \multicolumn{3}{@{}l}{
      \emph{Per coalition, direct, $|S|<q$}
    } \\
    Total
      & $\mathcal O(|S|^3+|C||S|(|S|+|T|))$
      & $\mathcal O(|S|q(|S|+|T|))$ \\

    \midrule
    \multicolumn{3}{@{}l}{
      \emph{Per coalition, incremental}
    } \\
    Shared inverse update
      & $\mathcal O(|S|^2)$
      & $\mathcal O(q^2)$ \\
    Label products and predictions
      & $\mathcal O(|C||S|(|S|+|T|))$
      & $\mathcal O(|C|q(|S|+q+|T|))$ \\
    Total
      & $\mathcal O(|C||S|(|S|+|T|))$
      & $\mathcal O(|C|q(|S|+q+|T|))$ \\

    \bottomrule
  \end{tabular}
\end{table}

For a direct feature-space solve, LRFShap forms
$G_S=\Phi_S^\top\Phi_S+\gamma I_q$ and
$R_S=\Phi_S^\top Y_S$, solves $G_S A_S=R_S$, and computes
$\Phi_{\mathrm{te}}A_S$.
For $|S|<q$, the table assumes that LRFShap explicitly forms
$\widetilde K_{SS}=\Phi_S\Phi_S^\top$ and
$\widetilde K_{TS}=\Phi_{\mathrm{te}}\Phi_S^\top$ before solving
the smaller kernel-space system.
The incremental label-product bounds allow these products to
be recomputed at each evaluated prefix; sharing the inverse
does not eliminate their dependence on $|C|$.

The comparisons below sum over all $|N|$ nonempty prefixes
of each permutation. They therefore provide upper bounds
when TMC truncation is enabled. Both methods use the same
fixed number $B$ of permutations and the same truncation
rule. However, their prefix utilities and full-coalition
utilities can differ, so they can evaluate different numbers
of prefixes before truncation. These differences also
contribute to the observed wall-clock runtimes.

\begin{itemize}

  \item \textbf{Direct factorization.}
  Evaluating every prefix with a direct solve costs
  $$
  \mathcal O\!\left(
    B\bigl(
      |N|^4
      +|C||N|^3
      +|C||T||N|^2
    \bigr)
  \right)
  $$
  for FreeShap and
  $\mathcal O\!\left(
    B\bigl(
      |N|^2q^2
      +|N|q^3
      +|C||T||N|q
      +|T|q^3
    \bigr)
  \right)$
  for LRFShap.
  Excluding prediction and preparation costs, the leading
  direct-solve terms are $\mathcal O(B|N|^4)$ and
  $\mathcal O(B|N|^2q^2)$, respectively.
  Their ratio is $(|N|/q)^2$.
  This comparison concerns direct-solve costs; practical
  runtimes also depend on incremental updates, inverse
  refreshes, preparation, and TMC truncation.
  FreeShap incurs no additional class factor in its
  shared factorization cost.

  \item \textbf{Small coalitions.}
  When $|S|<q$, LRFShap solves a $|S|\times |S|$ system,
  so the low-rank representation does not reduce the
  system dimension at these prefixes.
  It also incurs the cost of constructing
  $\widetilde K_{SS}$ and $\widetilde K_{TS}$ from the
  features, whereas FreeShap uses the corresponding
  full-kernel blocks.
  Thus, a runtime advantage is not guaranteed for small
  coalitions.
  These prefixes contribute
  $\mathcal O(q^4+|T|q^3)$ per permutation to LRFShap.
  Since $q\le |N|$, the $q^4$ term is absorbed by
  $\mathcal O(|N|^2q^2)$.
  The $|T|q^3$ term is also absorbed when
  $|T|\lesssim |N|^2/q$; it is retained explicitly in
  the general bound above.

  \item \textbf{Incremental updates and inverse refreshes.}
  Both implementations begin caching after a coalition-size
  threshold $h$ ($h=500$ in our code).
  FreeShap additionally rebuilds its shared inverse at
  every multiple of $h$ to limit accumulated round-off.
  Each rebuild is performed once and reused across classes.

  Summing the incremental terms over a full permutation
  gives
  $\mathcal O\!\left(
    |C||N|^2(|N|+|T|)
  \right)$
  for FreeShap and
  $\mathcal O\!\left(
    |C||N|q(|N|+q+|T|)
  \right)$
  for LRFShap.
  These expressions include the label-dependent products
  and predictions; the inverse updates themselves are
  shared across classes.

  For $h\le |N|$, FreeShap's initial direct factorizations
  and periodic inverse rebuilds add
  $\mathcal O(h^4+|N|^4/h)$ per permutation.
  For LRFShap, let $s=\min\{h,q\}$.
  Its initial direct evaluations add
  $\mathcal O(h^2qs+|T|qs^2)$ for matrix construction,
  factorization, and small-coalition cross-kernel
  construction.
  The initial feature-space inverse construction is
  covered by the stated bounds.
  These startup and refresh costs must be added to the
  incremental terms, and all per-permutation costs are
  multiplied by $B$.
  Their contribution to actual runtime depends on the
  problem dimensions and the prefixes reached before
  truncation.

  \item \textbf{Preparation.}
  Feature construction is a one-time cost shared by all
  permutations and is included in our reported wall-clock
  speedups.
  For eigen features, the table gives the dense
  eigendecomposition cost
  $\mathcal O(|N|^3+|T||N|q)$.
  Our implementation uses a dense eigendecomposition for
  $|N|\le 5000$ and a partial Lanczos decomposition above
  this threshold.
  For larger problems, randomized subspace
  iteration~\citep{halko2011finding} provides an alternative
  with $\mathcal O(|N|^2q)$ decomposition cost for a fixed
  number of iterations and oversampling proportional to
  $q$, plus $\mathcal O(|T||N|q)$ for test features.

  For Nystr\"om features, dense eigendecomposition of the
  $m\times m$ landmark matrix and feature construction cost
  $\mathcal O(m^3+(|N|+|T|)mq)$.
  Keeping $m$ and $q$ distinct accounts for the case in
  which only $q<m$ positive eigenvalues are retained.

\end{itemize}

\clearpage
\section{KPS errors and fidelity scores}\label{app:kps_error_fidelity}
In this section, we report the kernel approximation error, the
prediction error, and the Data Shapley error for all models
and datasets at $n=2000$, together with the Spearman and
Pearson correlations with the baseline mentioned in
Section~\ref{sec:results_qna}.

In addition to the uniform Nystr\"om sampling used in the main
text, the figures include the leverage-score Nystr\"om
method~\citep{alaoui2015fast}.

For both Nystr\"om variants, the pseudoinverse
$K_{MM}^{\dagger}$ in Equation~\eqref{eq:nys_approx} is computed
from the eigendecomposition $K_{MM}=Q\Sigma Q^{\top}$ by
discarding the eigenvalues below $\varepsilon=10^{-8}$;
writing $Q_s$ and $\Sigma_s$ for the retained eigenvectors and
eigenvalues, the features are
$\Phi=K_{NM}Q_s\Sigma_s^{-1/2}$, so that
$\Phi\Phi^{\top}=K_{NM}K_{MM}^{\dagger}K_{MN}$ on the retained
spectrum.
Uniform Nystr\"om draws the $m$ landmarks uniformly at random.
Leverage-score Nystr\"om draws them with probability
proportional to the ridge leverage scores
$\bigl[K_{NN}(K_{NN}+\bar\gamma I)^{-1}\bigr]_{ii}$ with
$\bar\gamma=\gamma n$, computed exactly from the
eigendecomposition of $K_{NN}$; no importance reweighting is
applied, and the same landmark seed is used for both variants.
The two Nystr\"om variants behave similarly on all datasets.

The fidelity scores point in the same direction as the errors:
the eigen decomposition achieves higher correlations than
both Nystr\"om variants at every rank, and in particular its
higher Spearman correlation means that it better preserves the
ranking of valuable data.
  
\begin{figure}[!ht]
      \centering      \includegraphics[width=1\textwidth]{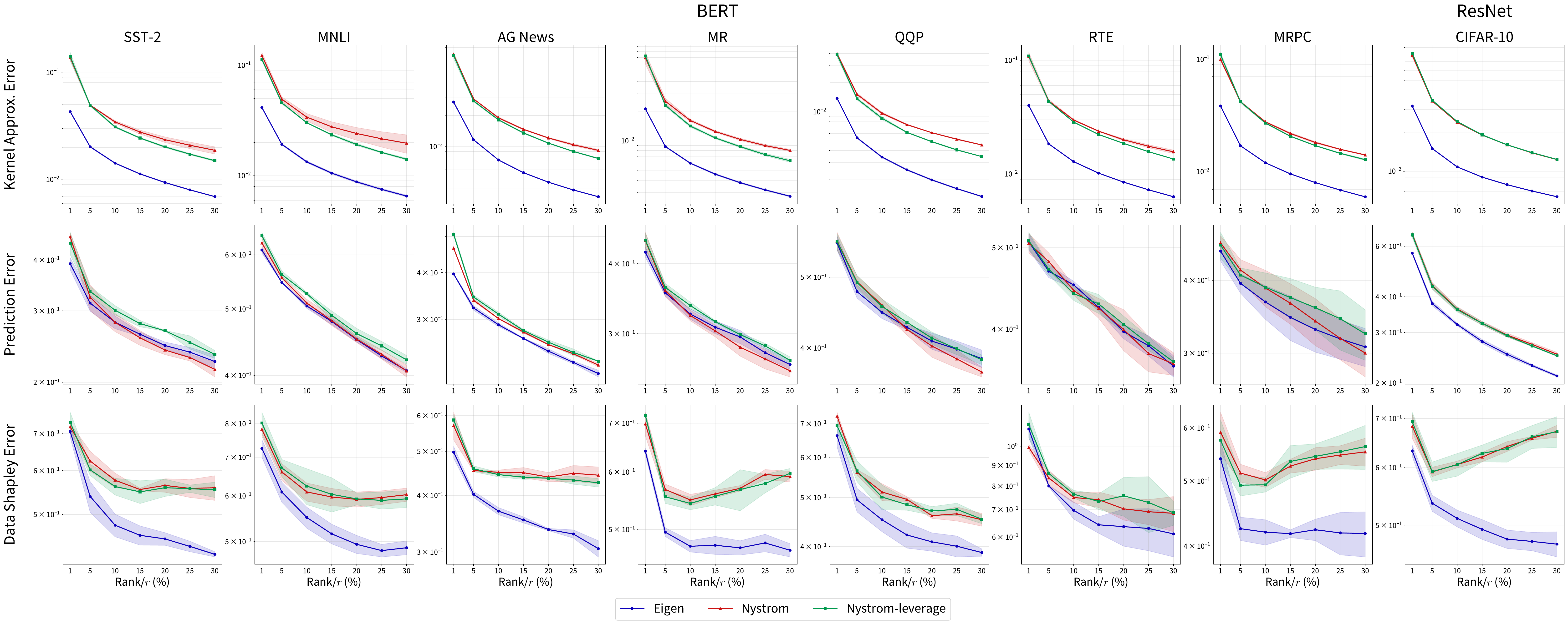}
      \caption{Kernel approximation error, prediction error, and data Shapley errors for BERT and ResNet-18.}
      \label{kps_error_3x8_bert_resnet}
  \end{figure}

\begin{figure}[!ht]
      \centering      \includegraphics[width=1\textwidth]{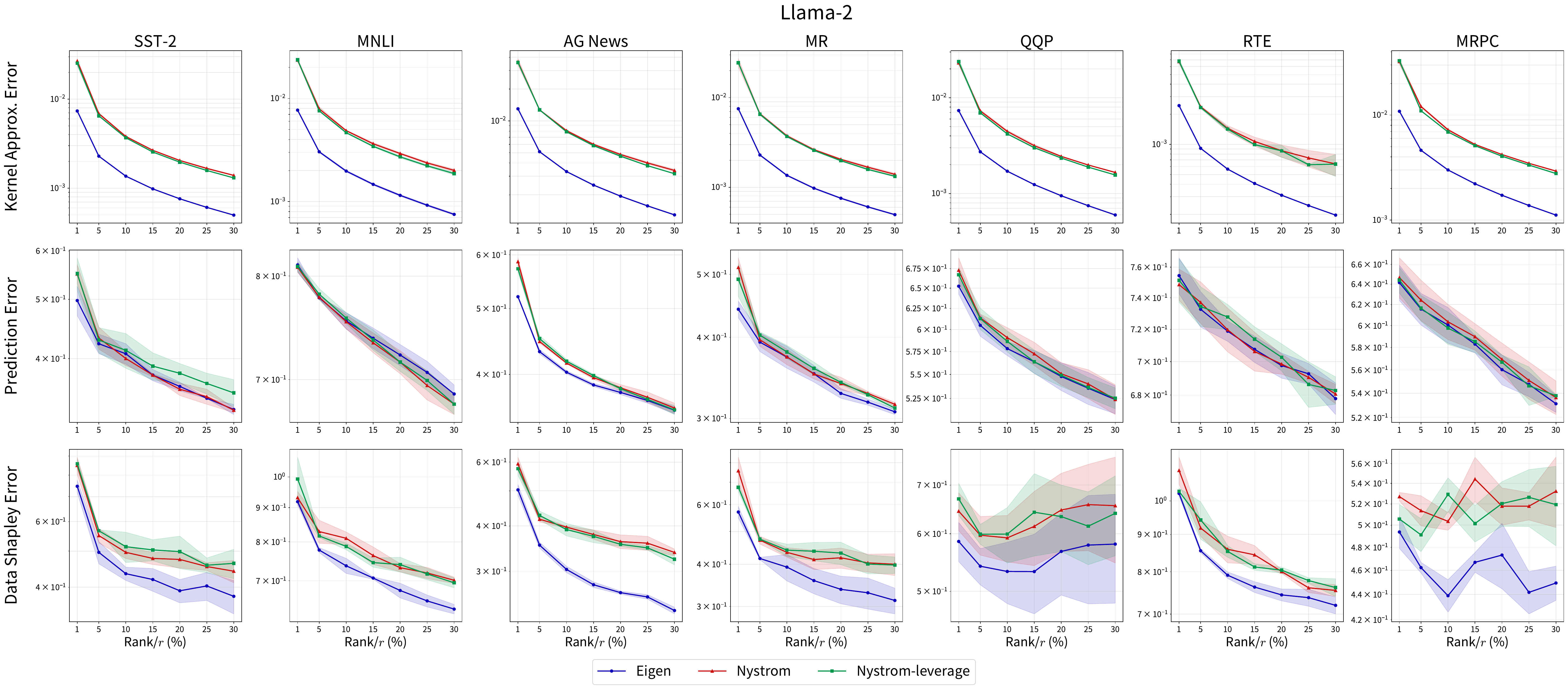}
      \caption{Kernel approximation error, prediction error, and data Shapley errors for Llama-2.}
      \label{kps_error_3x7_llama}
  \end{figure}

\begin{figure}[!ht]
      \centering      \includegraphics[width=1\textwidth]{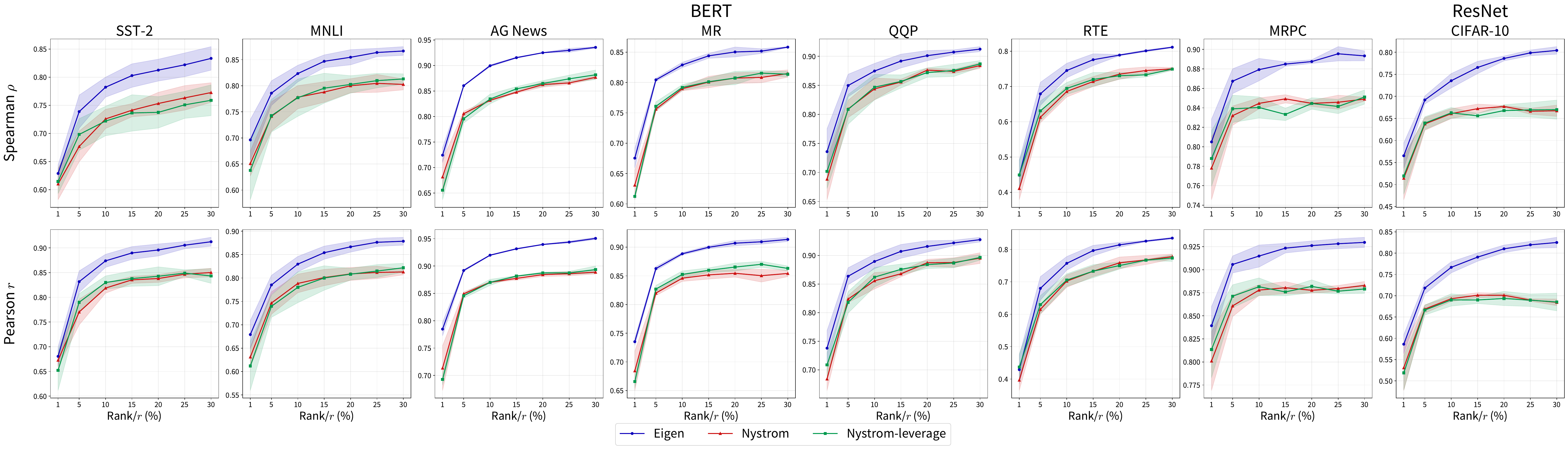}
      \caption{the Spearman and Pearson correlations for BERT and ResNet-18.}
      \label{fidelity_2x8_bert_resnet}
  \end{figure}

\begin{figure}[!ht]
      \centering      \includegraphics[width=1\textwidth]{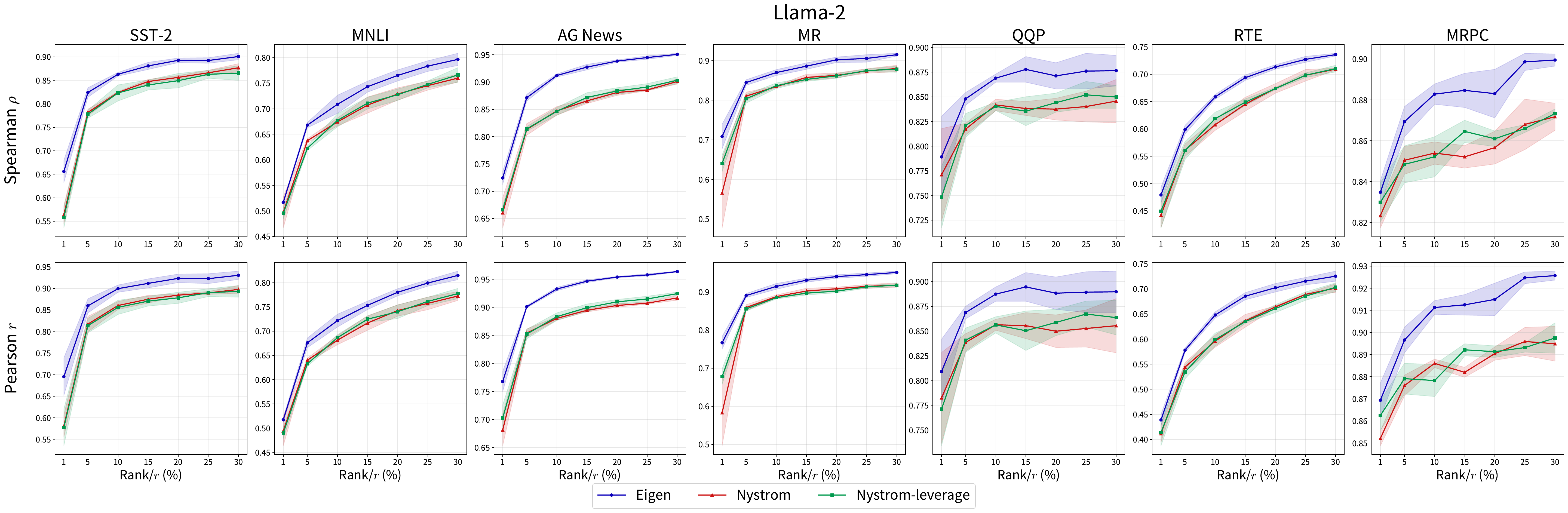}
      \caption{the Spearman and Pearson correlations for Llama-2.}
      \label{.pdf}
  \end{figure}

\clearpage
\section{Sensitivity of the Nystr\"om Method to the Choice of Landmarks}\label{app:nystrom_landmark}

\begin{figure}[!ht]
      \centering
      \includegraphics[width=1\textwidth]{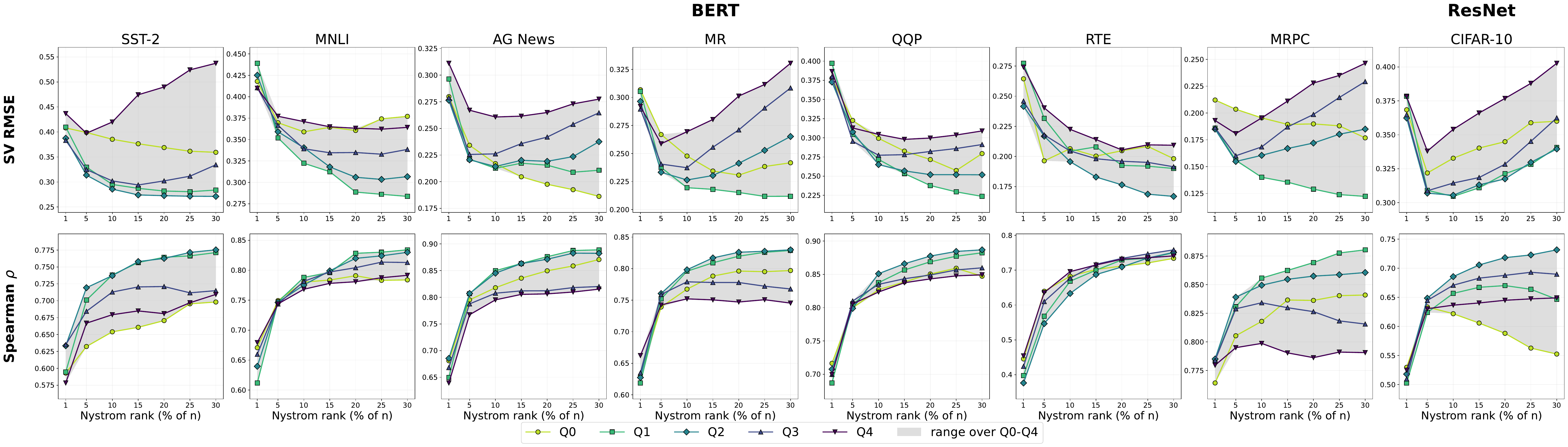}
      \caption{Kernel relative error, root mean square error (RMSE) of data Shapley with respect to the baseline, and Spearman correlation for the Nystr\"om landmark settings Q0--Q4. The model is BERT and ResNet-18. With increasing rank, the gap across settings shrinks for the kernel error but widens for RMSE and Spearman correlation.
      }
      \label{fig:rmse_spearman_2x8_bert_resnet}
  \end{figure}

\begin{figure}[!ht]
      \centering
      \includegraphics[width=1\textwidth]{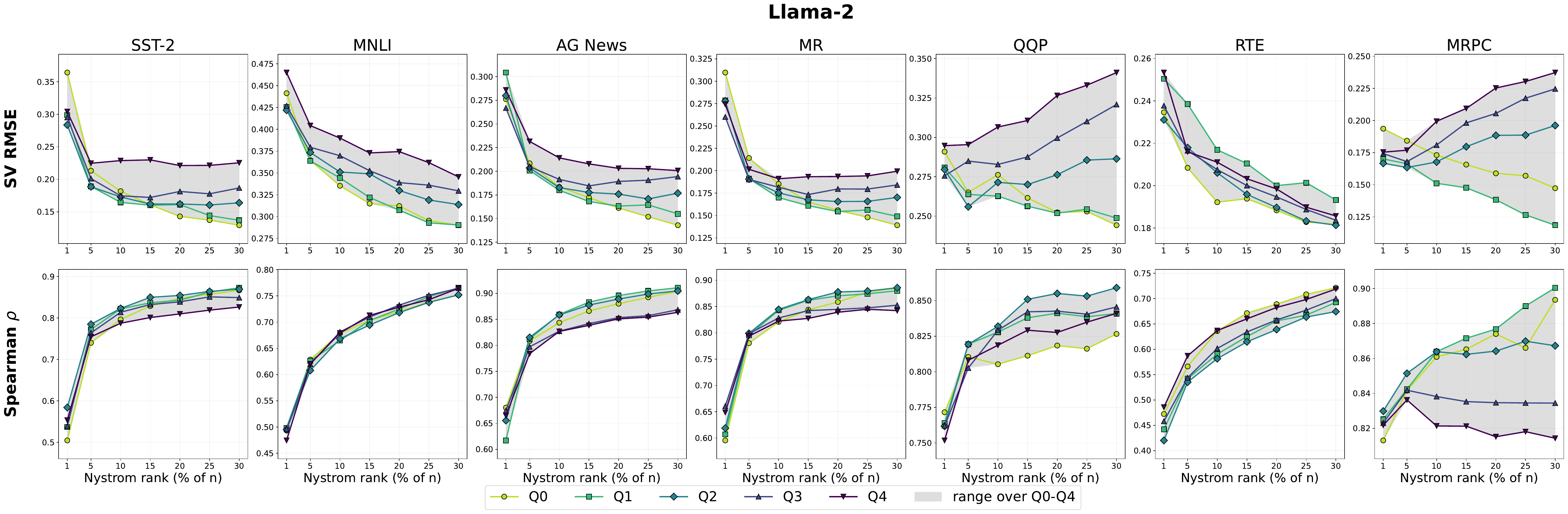}
      \caption{Kernel relative error, root mean square error (RMSE) of data Shapley with respect to the baseline, and Spearman correlation for the Nystr\"om landmark settings Q0--Q4. The model is Llama-2. With increasing rank, the gap across settings shrinks for the kernel error but widens for RMSE and Spearman correlation.
      }
      \label{fig:rmse_spearman_2x7_llama}
  \end{figure}

We show that the Data Shapley error of the Nystr\"om method depends heavily on the choice of landmarks, and that this dependence grows with the rank.
For a dataset with $n=2000$ samples and a model, we first compute Shapley values with the baseline.
At each rank $r$, we construct five landmark sets of equal size along this ordering. Q0 and Q4 contain the $r$ samples with the smallest and largest baseline Shapley values, respectively, while Q1, Q2, and Q3 contain $r$ consecutive samples centered at the 25th, 50th, and 75th percentiles, respectively.
We then compute Shapley values for each of the Q0--Q4 settings.
Figure~\ref{fig:rmse_spearman_2x8_bert_resnet} shows that, as $r$ increases, the variation across Q0--Q4 decreases for the kernel approximation error but increases for the RMSE and Spearman correlation.

\clearpage
\section{Experimental details}\label{app:exp_details}
The models which are used in the experiments are BERT (\texttt{bert-base-uncased};~\citealp{devlin2019bert}), Llama-2 (\texttt{llama-2-7b-hf};~\citealp{touvron2023llama}), and ResNet-18 (\texttt{resnet-18-pretrained};~\citealp{he2016deep}).
Table~\ref{tab:settings} lists the settings shared by all experiments, which follow FreeShap, and Table~\ref{tab:datasets} summarizes the datasets.

\begin{table}[ht]
\centering\scriptsize
\caption{Experimental settings. The values of $\gamma$ and $\varepsilon$ showed the best performance and fidelity. For each dataset, both methods are timed on the same device.}
\label{tab:settings}
\begin{tabular}{ll}
\toprule
Shapley estimator & TMC sampling~\citep{ghorbani2019data}, replacing the Monte Carlo loop in Algorithm~\ref{alg:lr_freeshap} \\
Permutations / tolerance & 500 / 0.05 \\
eNTK approximation & first-logit approximation~\citep{wei2022more} (Section~2), for both methods \\
Regularization $\gamma$ / $\varepsilon$ & $10^{-2}$ / $10^{-8}$ \\
Seeds & 3 (all results are averaged) \\
GPU & NVIDIA GeForce RTX 3090 (CIFAR-10); NVIDIA RTX A6000 (all other datasets) \\
\bottomrule
\end{tabular}
\end{table}

The construction of the Nystr\"om features, including the pseudoinverse cutoff and the leverage-score sampling, is described in Appendix~\ref{app:kps_error_fidelity}.

\begin{table}[ht]
\centering\scriptsize
\caption{Datasets. $n$ is the number of training samples used, $|C|$ the number of classes, $|T|$ the size of the test set $D_T$ (the official validation split), and $|E|$ the size of the evaluation dataset (a pre-defined test set for MR and a held-out dataset from the train dataset for other datasets).}
\label{tab:datasets}
\begin{tabular}{llrrrrl}
\toprule
Dataset & Task & $|C|$ & $n$ & $|T|$ & $|E|$ & Models \\\midrule
AG News~\citep{zhang2015character}  & topic classification & 4  & 5000 & 1000 & 1000 & BERT, Llama-2 \\
MNLI~\citep{mnli}     & NLI                  & 3  & 5000 & 1000 & 1000 & BERT, Llama-2 \\
MR~\citep{pang2005seeing}       & sentiment            & 2  & 5000 & 1000 & 1000 & BERT, Llama-2 \\
MRPC~\citep{mrpc}     & paraphrase           & 2  & 3000 & 408 & 668 & BERT, Llama-2 \\
QQP~\citep{wang2019glue}      & paraphrase           & 2  & 5000 & 1000 & 1000 & BERT, Llama-2 \\
RTE~\citep{rte}      & NLI                  & 2  & 1500 & 277 & 990 & BERT, Llama-2 \\
SST-2~\citep{sst-2}    & sentiment            & 2  & 5000 & 872 & 1000 & BERT, Llama-2 \\
CIFAR-10~\citep{krizhevsky2009learning} & image classification & 10 & 5000 & 1000 & 1000 & ResNet-18 \\
\bottomrule
\end{tabular}
\end{table}

\clearpage
\section{Computation time}\label{computation_time}

In this section, we report the computation time of FreeShap and LRFShap across all models and datasets.
In the tables, we regard LRFShap as comparable to FreeShap when its data selection score is within $0.01$ of FreeShap.

\begin{table}[!ht]
\centering
\small
\setlength{\tabcolsep}{4pt}
\renewcommand{\arraystretch}{1.1}
\caption{Wall-clock cost of data Shapley estimation of FreeShap and LRFShap, on \texttt{bert-base-uncased} at $n=5000$ ($1500$ for RTE and $3000$ for MRPC). \textbf{DS} is
the data Shapley computation time and \textbf{ED} the eigendecomposition time. \textbf{Speedup} is $\mathrm{DS}_{\text{FreeShap}}/(\mathrm{ED}_{\text{Eigen}}+\mathrm{DS}_{\text{Eigen}})$. Times are in seconds, averaged over three seeds. \textcolor{magenta}{Magenta} speedups mark the cases where LRFShap is comparable to or better than FreeShap on data selection.}
\label{tab:runtime_eigen_vs_freeshap}
\begin{tabular}{l r ccccccc}
\toprule
\multirow{3}{*}{Dataset} & FreeShap & \multicolumn{7}{c}{Eigen} \\
\cmidrule(lr){2-2} \cmidrule(lr){3-9}
& \multirow{2}{*}{DS (s)} & \multicolumn{7}{c}{ED (s) / DS (s) / \textbf{Speedup} ($\times$)} \\
\cmidrule(lr){3-9}
& & Rank 1\% & 5\% & 10\% & 15\% & 20\% & 25\% & 30\% \\
\midrule
AG News & 3513.90 & \makecell{7.01\\260.07\\\textcolor{magenta}{\textbf{13.16}}} &
\makecell{7.19\\716.93\\\textcolor{magenta}{\textbf{4.85}}} &
\makecell{7.07\\1095.03\\\textcolor{magenta}{\textbf{3.19}}} &
\makecell{7.42\\1527.65\\\textcolor{magenta}{\textbf{2.29}}} &
\makecell{7.36\\2008.99\\\textcolor{magenta}{\textbf{1.74}}} &
\makecell{7.37\\2428.07\\\textcolor{magenta}{\textbf{1.44}}} &
\makecell{7.44\\2633.04\\\textcolor{magenta}{\textbf{1.33}}} \\
\cmidrule(lr){1-9}
MNLI & 28898.75 & \makecell{7.30\\1287.57\\\textcolor{magenta}{\textbf{22.32}}} &
\makecell{7.16\\2853.85\\\textcolor{magenta}{\textbf{10.10}}} &
\makecell{7.46\\4072.05\\\textcolor{magenta}{\textbf{7.08}}} &
\makecell{7.41\\5995.83\\\textcolor{magenta}{\textbf{4.81}}} &
\makecell{7.07\\6266.81\\\textcolor{magenta}{\textbf{4.61}}} &
\makecell{7.31\\7438.42\\\textcolor{magenta}{\textbf{3.88}}} &
\makecell{7.41\\8259.17\\\textcolor{magenta}{\textbf{3.50}}} \\
\cmidrule(lr){1-9}
MR & 10515.90 & \makecell{7.45\\618.82\\\textcolor{magenta}{\textbf{16.79}}} &
\makecell{7.18\\1463.88\\\textcolor{magenta}{\textbf{7.15}}} &
\makecell{7.49\\2337.29\\\textcolor{magenta}{\textbf{4.48}}} &
\makecell{7.31\\3256.70\\\textcolor{magenta}{\textbf{3.22}}} &
\makecell{7.42\\4206.85\\\textcolor{magenta}{\textbf{2.50}}} &
\makecell{7.47\\5526.64\\\textcolor{magenta}{\textbf{1.90}}} &
\makecell{7.18\\6482.37\\\textcolor{magenta}{\textbf{1.62}}} \\
\cmidrule(lr){1-9}
MRPC & 3231.15 & \makecell{2.08\\327.45\\\textcolor{magenta}{\textbf{9.81}}} &
\makecell{2.05\\771.08\\\textcolor{magenta}{\textbf{4.18}}} &
\makecell{2.13\\1128.21\\\textcolor{magenta}{\textbf{2.86}}} &
\makecell{2.09\\1652.71\\\textcolor{magenta}{\textbf{1.95}}} &
\makecell{2.13\\2016.76\\\textcolor{magenta}{\textbf{1.60}}} &
\makecell{2.14\\1968.87\\\textcolor{magenta}{\textbf{1.64}}} &
\makecell{2.17\\2284.23\\\textcolor{magenta}{\textbf{1.41}}} \\
\cmidrule(lr){1-9}
QQP & 22666.92 & \makecell{7.41\\669.17\\\textbf{33.50}} &
\makecell{7.29\\2097.78\\\textcolor{magenta}{\textbf{10.77}}} &
\makecell{7.39\\3981.74\\\textcolor{magenta}{\textbf{5.68}}} &
\makecell{7.33\\5071.86\\\textcolor{magenta}{\textbf{4.46}}} &
\makecell{7.38\\5694.62\\\textcolor{magenta}{\textbf{3.98}}} &
\makecell{7.35\\6796.78\\\textcolor{magenta}{\textbf{3.33}}} &
\makecell{7.31\\8265.90\\\textcolor{magenta}{\textbf{2.74}}} \\
\cmidrule(lr){1-9}
RTE & 1355.60 & \makecell{0.52\\654.71\\\textbf{2.07}} &
\makecell{0.52\\565.80\\\textcolor{magenta}{\textbf{2.39}}} &
\makecell{0.51\\626.81\\\textcolor{magenta}{\textbf{2.16}}} &
\makecell{0.53\\708.65\\\textcolor{magenta}{\textbf{1.91}}} &
\makecell{0.53\\786.63\\\textcolor{magenta}{\textbf{1.72}}} &
\makecell{0.52\\852.03\\\textcolor{magenta}{\textbf{1.59}}} &
\makecell{0.53\\815.26\\\textcolor{magenta}{\textbf{1.66}}} \\
\cmidrule(lr){1-9}
SST-2 & 6824.66 & \makecell{7.24\\441.75\\\textcolor{magenta}{\textbf{15.20}}} &
\makecell{7.29\\958.94\\\textcolor{magenta}{\textbf{7.06}}} &
\makecell{7.14\\1528.11\\\textcolor{magenta}{\textbf{4.45}}} &
\makecell{7.25\\2178.87\\\textcolor{magenta}{\textbf{3.12}}} &
\makecell{7.28\\2751.13\\\textcolor{magenta}{\textbf{2.47}}} &
\makecell{7.45\\3504.77\\\textcolor{magenta}{\textbf{1.94}}} &
\makecell{7.50\\4176.16\\\textcolor{magenta}{\textbf{1.63}}} \\
\bottomrule
\end{tabular}
\end{table}
  
\begin{table}[!ht]
\centering
\small
\setlength{\tabcolsep}{4pt}
\renewcommand{\arraystretch}{1.1}
\caption{Wall-clock cost of data Shapley estimation of FreeShap and LRFShap, on \texttt{Llama-2-7b-hf} at $n=5000$ ($1500$ for RTE and
$3000$ for MRPC). \textbf{DS} is
the data Shapley computation time and \textbf{ED} the eigendecomposition time. \textbf{Speedup} is
$\mathrm{DS}_{\text{FreeShap}}/(\mathrm{ED}_{\text{Eigen}}+\mathrm{DS}_{\text{Eigen}})$. Times are in seconds, averaged over three
seeds. \textcolor{magenta}{Magenta} speedups mark the cases where LRFShap is comparable to or better than FreeShap on data selection. On MR, QQP and SST-2, TMC truncates after only a few iterations, so FreeShap itself is cheap and the speedup is correspondingly low; in these cases, however, FreeShap also attains substantially lower accuracy (see Section \ref{sec:results_qna}).}
\label{tab:runtime_eigen_vs_freeshap_llama}
\begin{tabular}{l r ccccccc}
\toprule
\multirow{3}{*}{Dataset} & FreeShap & \multicolumn{7}{c}{Eigen} \\
\cmidrule(lr){2-2} \cmidrule(lr){3-9}
& \multirow{2}{*}{DS (s)} & \multicolumn{7}{c}{ED (s) / DS (s) / \textbf{Speedup} ($\times$)} \\
\cmidrule(lr){3-9}
& & Rank 1\% & 5\% & 10\% & 15\% & 20\% & 25\% & 30\% \\
\midrule
AG News & 2698.49 & \makecell{7.25\\240.29\\\textcolor{magenta}{\textbf{10.90}}} & \makecell{7.41\\575.79\\\textcolor{magenta}{\textbf{4.63}}} &
\makecell{7.12\\878.23\\\textcolor{magenta}{\textbf{3.05}}} & \makecell{7.07\\1235.78\\\textcolor{magenta}{\textbf{2.17}}} & \makecell{7.50\\1578.36\\\textcolor{magenta}{\textbf{1.70}}} &
\makecell{7.06\\1790.97\\\textcolor{magenta}{\textbf{1.50}}} & \makecell{7.40\\1903.16\\\textcolor{magenta}{\textbf{1.41}}} \\
\cmidrule(lr){1-9}
MNLI & 18189.62 & \makecell{7.09\\2340.91\\\textbf{7.75}} & \makecell{7.08\\4458.57\\\textcolor{magenta}{\textbf{4.07}}} &
\makecell{7.27\\7654.89\\\textcolor{magenta}{\textbf{2.37}}} & \makecell{7.26\\7440.18\\\textcolor{magenta}{\textbf{2.44}}} & \makecell{7.37\\8795.92\\\textcolor{magenta}{\textbf{2.07}}} &
\makecell{7.63\\11475.75\\\textcolor{magenta}{\textbf{1.58}}} & \makecell{7.38\\11472.57\\\textcolor{magenta}{\textbf{1.58}}} \\
\cmidrule(lr){1-9}
MR & 2277.68 & \makecell{6.95\\442.25\\\textcolor{magenta}{\textbf{5.07}}} & \makecell{6.91\\1130.66\\\textcolor{magenta}{\textbf{2.00}}} &
\makecell{7.28\\1528.31\\\textcolor{magenta}{\textbf{1.48}}} & \makecell{7.33\\2084.85\\\textcolor{magenta}{\textbf{1.09}}} & \makecell{7.10\\2792.51\\\textcolor{magenta}{\textbf{0.81}}} &
\makecell{7.10\\3037.48\\\textcolor{magenta}{\textbf{0.75}}} & \makecell{7.35\\3546.07\\\textcolor{magenta}{\textbf{0.64}}} \\
\cmidrule(lr){1-9}
MRPC & 1062.20 & \makecell{2.13\\503.83\\\textcolor{magenta}{\textbf{2.10}}} & \makecell{2.11\\1195.96\\\textcolor{magenta}{\textbf{0.89}}} &
\makecell{2.08\\1398.12\\\textcolor{magenta}{\textbf{0.76}}} & \makecell{2.13\\1517.16\\\textcolor{magenta}{\textbf{0.70}}} & \makecell{2.11\\1851.80\\\textcolor{magenta}{\textbf{0.57}}} &
\makecell{2.13\\1988.92\\\textcolor{magenta}{\textbf{0.53}}} & \makecell{2.10\\2280.07\\\textcolor{magenta}{\textbf{0.47}}} \\
\cmidrule(lr){1-9}
QQP & 1269.64 & \makecell{7.17\\894.37\\\textcolor{magenta}{\textbf{1.41}}} & \makecell{7.36\\1819.15\\\textcolor{magenta}{\textbf{0.70}}} &
\makecell{7.31\\3150.27\\\textcolor{magenta}{\textbf{0.40}}} & \makecell{7.26\\4322.61\\\textcolor{magenta}{\textbf{0.29}}} & \makecell{7.38\\4679.51\\\textcolor{magenta}{\textbf{0.27}}} &
\makecell{7.30\\6216.22\\\textcolor{magenta}{\textbf{0.20}}} & \makecell{7.29\\7324.34\\\textcolor{magenta}{\textbf{0.17}}} \\
\cmidrule(lr){1-9}
RTE & 1934.86 & \makecell{0.52\\337.98\\\textcolor{magenta}{\textbf{5.72}}} & \makecell{0.53\\733.20\\\textcolor{magenta}{\textbf{2.64}}} &
\makecell{0.52\\685.37\\\textcolor{magenta}{\textbf{2.82}}} & \makecell{0.53\\747.77\\\textbf{2.59}} & \makecell{0.53\\836.38\\\textcolor{magenta}{\textbf{2.31}}} &
\makecell{0.54\\844.94\\\textcolor{magenta}{\textbf{2.29}}} & \makecell{0.53\\860.18\\\textcolor{magenta}{\textbf{2.25}}} \\
\cmidrule(lr){1-9}
SST-2 & 1536.96 & \makecell{7.17\\402.17\\\textcolor{magenta}{\textbf{3.75}}} & \makecell{7.25\\930.85\\\textcolor{magenta}{\textbf{1.64}}} &
\makecell{7.20\\1864.09\\\textbf{0.82}} & \makecell{7.33\\2260.33\\\textbf{0.68}} & \makecell{7.23\\3042.92\\\textbf{0.50}} &
\makecell{7.14\\3294.88\\\textbf{0.47}} & \makecell{7.37\\3905.43\\\textcolor{magenta}{\textbf{0.39}}} \\
\bottomrule
\end{tabular}
\end{table}

\begin{table}[!ht]
\centering
\small
\setlength{\tabcolsep}{4pt}
\renewcommand{\arraystretch}{1.1}
\caption{Wall-clock cost of data Shapley estimation of FreeShap and LRFShap, on \texttt{ResNet-18} at $n=5000$. \textbf{DS} is
the data Shapley computation time and \textbf{ED} the eigendecomposition time. \textbf{Speedup} is
$\mathrm{DS}_{\text{FreeShap}}/(\mathrm{ED}_{\text{Eigen}}+\mathrm{DS}_{\text{Eigen}})$. Times are in seconds, averaged over three
seeds. \textcolor{magenta}{Magenta} speedups mark the cases where LRFShap is comparable to or better than FreeShap on data selection.}
\label{tab:runtime_eigen_vs_freeshap_resnet}
\begin{tabular}{l r ccccccc}
\toprule
\multirow{3}{*}{Dataset} & FreeShap & \multicolumn{7}{c}{Eigen} \\
\cmidrule(lr){2-2} \cmidrule(lr){3-9}
& \multirow{2}{*}{DS (s)} & \multicolumn{7}{c}{ED (s) / DS (s) / \textbf{Speedup} ($\times$)} \\
\cmidrule(lr){3-9}
& & Rank 1\% & 5\% & 10\% & 15\% & 20\% & 25\% & 30\% \\
\midrule
CIFAR-10 & 2929.20 & \makecell{15.37\\206.08\\\textcolor{magenta}{\textbf{13.23}}} & \makecell{15.58\\569.09\\\textcolor{magenta}{\textbf{5.01}}} &
\makecell{15.82\\898.23\\\textcolor{magenta}{\textbf{3.20}}} & \makecell{15.85\\1283.69\\\textcolor{magenta}{\textbf{2.25}}} & \makecell{15.85\\1595.59\\\textcolor{magenta}{\textbf{1.82}}} &
\makecell{15.94\\1925.10\\\textcolor{magenta}{\textbf{1.51}}} & \makecell{15.46\\2469.52\\\textcolor{magenta}{\textbf{1.18}}} \\
\bottomrule
\end{tabular}
\end{table}

\clearpage
\section{Raw accuracies}\label{sec:more_experiments}

\begin{figure}[!ht]
      \centering      \includegraphics[width=1\textwidth]{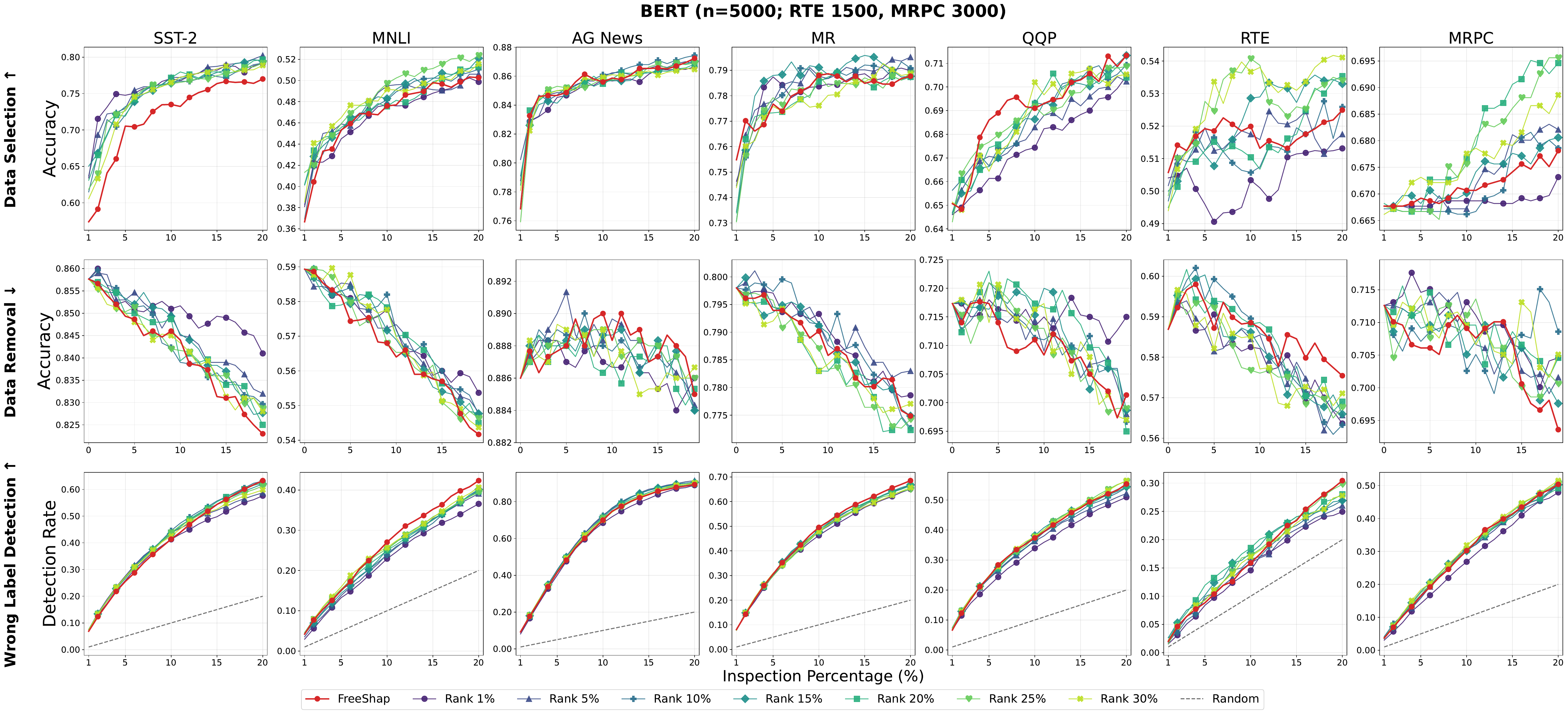}
      \caption{Data selection, removal, and wrong label detection results for 1-20\% selection percentages for \texttt{bert-base-uncased}.}
      \label{fig:sel_rem_wld_5000_bert_20}
  \end{figure}

\begin{figure}[!ht]
      \centering      \includegraphics[width=1\textwidth]{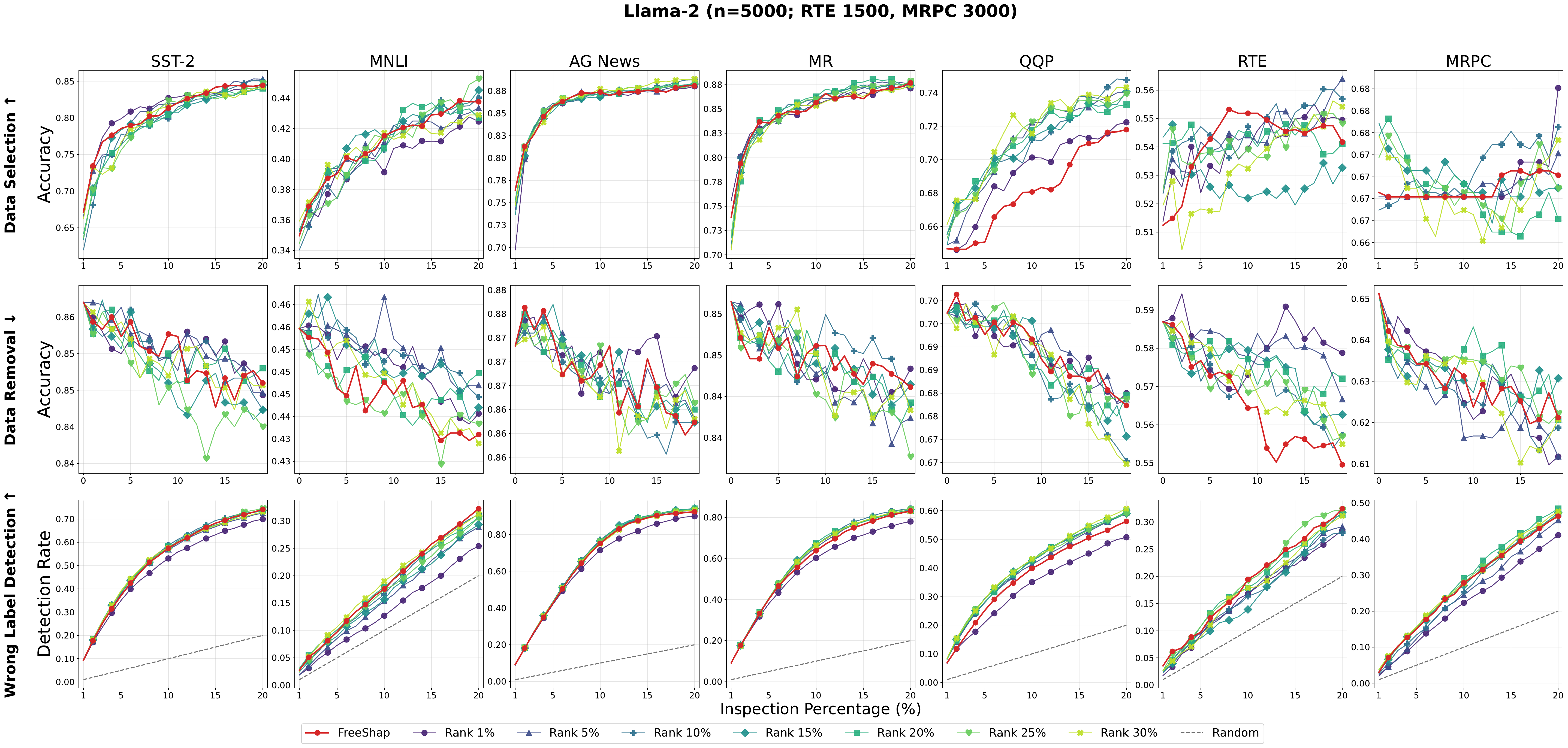}
      \caption{Data selection, removal, and wrong label detection results for 1-20\% selection percentages for \texttt{llama-2-7b-hf}.}
      \label{fig:sel_rem_wld_5000_llama_20}
  \end{figure}

\begin{figure}[!ht]
      \centering      \includegraphics[width=0.7\textwidth]{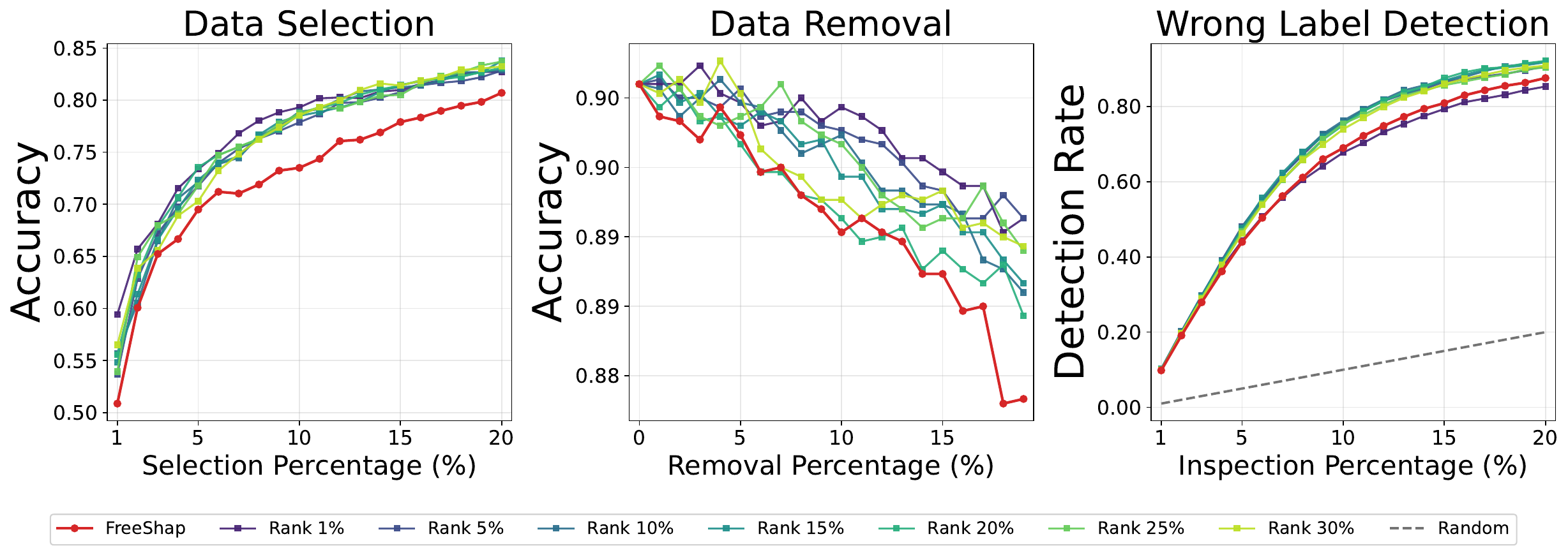}
      \caption{The results of all tasks in CIFAR-10 for \texttt{resnet-18-pretrained}.}
      \label{fig:resnet_5000}
  \end{figure}

\clearpage
\section{Mean accuracy results}\label{sec:mean_accuracy_results}

In this section, we report the mean accuracy (detection rate for wrong label detection) across three seeds for all datasets and downstream tasks.
For data selection and wrong label detection, higher is better; for data removal, lower is better, since the most valuable data are removed first. In each table, scores equal to or better than the baseline are shown in \textbf{bold}, and scores worse than the baseline by less than 0.01 are \underline{underlined}.

\begin{table}[H]
\centering
\scriptsize
\begin{tabular}{lcccccccc}
\toprule
Dataset & FreeShap & Rank 1\% & 5\% & 10\% & 15\% & 20\% & 25\% & 30\% \\
\midrule
AG News & 0.8534 & \underline{0.8503} & \textbf{0.8546} & \textbf{0.8552} & \underline{0.8519} & \underline{0.8533} & \underline{0.8522} & \underline{0.8523} \\
MNLI & 0.4684 & \underline{0.4659} & \textbf{0.4752} & \textbf{0.4747} & \textbf{0.4793} & \textbf{0.4754} & \textbf{0.4862} & \textbf{0.4801} \\
MR & 0.7804 & \textbf{0.7815} & \textbf{0.7841} & \textbf{0.7817} & \textbf{0.7854} & \underline{0.7787} & \underline{0.7787} & \underline{0.7778} \\
MRPC & 0.6715 & \underline{0.6686} & \textbf{0.6730} & \underline{0.6711} & \textbf{0.6731} & \textbf{0.6800} & \textbf{0.6783} & \textbf{0.6767} \\
QQP & 0.6908 & 0.6761 & \underline{0.6818} & \underline{0.6876} & \underline{0.6856} & \underline{0.6879} & \underline{0.6876} & \underline{0.6872} \\
RTE & 0.5173 & 0.5036 & \underline{0.5160} & \underline{0.5136} & \textbf{0.5224} & \underline{0.5156} & \textbf{0.5267} & \textbf{0.5293} \\
SST-2 & 0.7179 & \textbf{0.7599} & \textbf{0.7593} & \textbf{0.7486} & \textbf{0.7530} & \textbf{0.7521} & \textbf{0.7487} & \textbf{0.7464} \\
\bottomrule
\end{tabular}
\caption{The mean accuracy of \texttt{bert-base-uncased} when the selection percentage is 1-20\%. Higher is better.}
\label{tab:mean_accuracy_selection_bert_20}
\end{table}

\begin{table}[H]
\centering
\scriptsize
\begin{tabular}{lcccccccc}
\toprule
Dataset & FreeShap & Rank 1\% & 5\% & 10\% & 15\% & 20\% & 25\% & 30\% \\
\midrule
AG News & 0.8607 & \underline{0.8561} & \underline{0.8593} & \textbf{0.8615} & \underline{0.8588} & \textbf{0.8610} & \textbf{0.8613} & \textbf{0.8630} \\
MNLI & 0.4094 & 0.3986 & \underline{0.4068} & \underline{0.4059} & \textbf{0.4133} & \textbf{0.4095} & \underline{0.4082} & \underline{0.4072} \\
MR & 0.8469 & \textbf{0.8474} & \textbf{0.8482} & \textbf{0.8488} & \underline{0.8462} & \textbf{0.8510} & \textbf{0.8483} & \underline{0.8429} \\
MRPC & 0.6687 & \textbf{0.6691} & \underline{0.6687} & \textbf{0.6707} & \textbf{0.6698} & \underline{0.6683} & \textbf{0.6692} & \underline{0.6681} \\
QQP & 0.6819 & \textbf{0.6925} & \textbf{0.7100} & \textbf{0.7098} & \textbf{0.7088} & \textbf{0.7100} & \textbf{0.7118} & \textbf{0.7163} \\
RTE & 0.5417 & \underline{0.5403} & \textbf{0.5439} & \textbf{0.5466} & 0.5281 & \textbf{0.5423} & \underline{0.5411} & \underline{0.5342} \\
SST-2 & 0.8054 & \textbf{0.8103} & \underline{0.8046} & 0.7938 & 0.7946 & 0.7924 & 0.7940 & \underline{0.7976} \\
\bottomrule
\end{tabular}
\caption{The mean accuracy of \texttt{meta-llama/Llama-2-7b-hf} when the selection percentage is 1-20\%. Higher is better.}
\label{tab:mean_accuracy_selection_llama_20}
\end{table}

\begin{table}[H]
\centering
\scriptsize
\begin{tabular}{lcccccccc}
\toprule
Dataset & FreeShap & Rank 1\% & 5\% & 10\% & 15\% & 20\% & 25\% & 30\% \\
\midrule
CIFAR-10 & 0.7260 & \textbf{0.7699} & \textbf{0.7575} & \textbf{0.7603} & \textbf{0.7597} & \textbf{0.7636} & \textbf{0.7618} & \textbf{0.7609} \\
\bottomrule
\end{tabular}
\caption{The mean accuracy of \texttt{resnet-18-pretrained} when the selection percentage is 1-20\%. Higher is better.}
\label{tab:mean_accuracy_selection_resnet_20}
\end{table}

\begin{table}[H]
\centering
\scriptsize
\begin{tabular}{lcccccccc}
\toprule
Dataset & FreeShap & Rank 1\% & 5\% & 10\% & 15\% & 20\% & 25\% & 30\% \\
\midrule
AG News & 0.8880 & \textbf{0.8867} & \textbf{0.8877} & \textbf{0.8874} & \textbf{0.8874} & \textbf{0.8870} & \textbf{0.8879} & \textbf{0.8870} \\
MNLI & 0.5645 & \underline{0.5694} & \underline{0.5682} & \underline{0.5693} & \underline{0.5673} & \underline{0.5681} & \underline{0.5657} & \underline{0.5673} \\
MR & 0.7865 & \underline{0.7885} & \underline{0.7897} & \underline{0.7887} & \underline{0.7871} & \textbf{0.7840} & \textbf{0.7843} & \textbf{0.7845} \\
MRPC & 0.7047 & \underline{0.7095} & \underline{0.7076} & \underline{0.7075} & \underline{0.7056} & \underline{0.7080} & \underline{0.7064} & \underline{0.7081} \\
QQP & 0.7085 & \underline{0.7150} & \underline{0.7102} & \underline{0.7107} & \underline{0.7122} & \underline{0.7133} & \underline{0.7095} & \underline{0.7101} \\
RTE & 0.5860 & \textbf{0.5789} & \textbf{0.5794} & \textbf{0.5828} & \textbf{0.5809} & \textbf{0.5832} & \textbf{0.5787} & \textbf{0.5790} \\
SST-2 & 0.8394 & 0.8495 & \underline{0.8449} & \underline{0.8420} & \underline{0.8435} & \underline{0.8418} & \underline{0.8412} & \underline{0.8407} \\
\bottomrule
\end{tabular}
\caption{The mean accuracy of \texttt{bert-base-uncased} when the removal percentage is 1-20\%. Lower is better.}
\label{tab:mean_accuracy_removal_bert_20}
\end{table}

\begin{table}[H]
\centering
\scriptsize
\begin{tabular}{lcccccccc}
\toprule
Dataset & FreeShap & Rank 1\% & 5\% & 10\% & 15\% & 20\% & 25\% & 30\% \\
\midrule
AG News & 0.8684 & \underline{0.8701} & \underline{0.8684} & \textbf{0.8671} & \textbf{0.8683} & \underline{0.8685} & \textbf{0.8681} & \textbf{0.8669} \\
MNLI & 0.4438 & \underline{0.4505} & \underline{0.4537} & \underline{0.4537} & \underline{0.4520} & \underline{0.4472} & \textbf{0.4425} & \underline{0.4470} \\
MR & 0.8492 & \underline{0.8493} & \textbf{0.8470} & \underline{0.8500} & \textbf{0.8490} & \textbf{0.8476} & \textbf{0.8465} & \textbf{0.8481} \\
MRPC & 0.6291 & \textbf{0.6278} & \textbf{0.6239} & \textbf{0.6261} & \underline{0.6305} & \underline{0.6321} & \underline{0.6317} & \textbf{0.6267} \\
QQP & 0.6884 & \textbf{0.6882} & \underline{0.6887} & \textbf{0.6838} & \textbf{0.6862} & \textbf{0.6871} & \textbf{0.6862} & \textbf{0.6838} \\
RTE & 0.5636 & 0.5804 & 0.5795 & \underline{0.5692} & \underline{0.5730} & 0.5747 & \underline{0.5697} & \underline{0.5701} \\
SST-2 & 0.8494 & \underline{0.8499} & \underline{0.8505} & \underline{0.8495} & \textbf{0.8468} & \underline{0.8494} & \textbf{0.8448} & \textbf{0.8489} \\
\bottomrule
\end{tabular}
\caption{The mean accuracy of \texttt{meta-llama/Llama-2-7b-hf} when the removal percentage is 1-20\%. Lower is better.}
\label{tab:mean_accuracy_removal_llama_20}
\end{table}

\begin{table}[H]
\centering
\scriptsize
\begin{tabular}{lcccccccc}
\toprule
Dataset & FreeShap & Rank 1\% & 5\% & 10\% & 15\% & 20\% & 25\% & 30\% \\
\midrule
CIFAR-10 & 0.8904 & \underline{0.8971} & \underline{0.8964} & \underline{0.8947} & \underline{0.8944} & \underline{0.8919} & \underline{0.8954} & \underline{0.8944} \\
\bottomrule
\end{tabular}
\caption{The mean accuracy of \texttt{resnet-18-pretrained} when the removal percentage is 1-20\%. Lower is better.}
\label{tab:mean_accuracy_removal_resnet_20}
\end{table}

\begin{table}[H]
\centering
\scriptsize
\begin{tabular}{lcccccccc}
\toprule
Dataset & FreeShap & Rank 1\% & 5\% & 10\% & 15\% & 20\% & 25\% & 30\% \\
\midrule
AG News & 0.6314 & 0.6175 & \textbf{0.6491} & \textbf{0.6480} & \textbf{0.6449} & \textbf{0.6392} & \textbf{0.6361} & \textbf{0.6378} \\
MNLI & 0.2604 & 0.2221 & 0.2357 & 0.2395 & 0.2421 & 0.2448 & 0.2488 & \underline{0.2525} \\
MR & 0.4606 & 0.4378 & \underline{0.4517} & 0.4488 & \underline{0.4527} & \underline{0.4536} & 0.4454 & 0.4464 \\
MRPC & 0.2992 & 0.2724 & \underline{0.2959} & \underline{0.2964} & \textbf{0.3021} & \underline{0.2975} & \textbf{0.3027} & \textbf{0.3085} \\
QQP & 0.3627 & 0.3296 & 0.3489 & \underline{0.3557} & \textbf{0.3645} & \underline{0.3621} & \textbf{0.3678} & \textbf{0.3677} \\
RTE & 0.1689 & 0.1493 & 0.1569 & \underline{0.1624} & \textbf{0.1733} & \textbf{0.1780} & \textbf{0.1717} & 0.1664 \\
SST-2 & 0.4031 & 0.3863 & \textbf{0.4036} & \textbf{0.4189} & \textbf{0.4122} & \textbf{0.4150} & \textbf{0.4102} & \textbf{0.4038} \\
\bottomrule
\end{tabular}
\caption{The mean detection rate of \texttt{bert-base-uncased} when the inspection percentage is 1-20\%. Higher is better.}
\label{tab:mean_detection_rate_wld_bert_20}
\end{table}

\begin{table}[H]
\centering
\scriptsize
\begin{tabular}{lcccccccc}
\toprule
Dataset & FreeShap & Rank 1\% & 5\% & 10\% & 15\% & 20\% & 25\% & 30\% \\
\midrule
AG News & 0.6661 & 0.6370 & \textbf{0.6729} & \textbf{0.6756} & \textbf{0.6769} & \textbf{0.6756} & \textbf{0.6722} & \textbf{0.6683} \\
MNLI & 0.1830 & 0.1364 & 0.1597 & \underline{0.1774} & 0.1647 & \underline{0.1812} & 0.1730 & \textbf{0.1864} \\
MR & 0.5842 & 0.5500 & \textbf{0.5931} & \textbf{0.6016} & \textbf{0.5982} & \textbf{0.5990} & \textbf{0.5980} & \textbf{0.5965} \\
MRPC & 0.2720 & 0.2238 & 0.2467 & 0.2614 & \textbf{0.2746} & \textbf{0.2852} & \textbf{0.2755} & \textbf{0.2788} \\
QQP & 0.3743 & 0.3319 & \textbf{0.3965} & \textbf{0.4032} & \textbf{0.4062} & \textbf{0.4060} & \textbf{0.4063} & \textbf{0.4121} \\
RTE & 0.1888 & 0.1639 & 0.1708 & 0.1651 & 0.1638 & \underline{0.1843} & \textbf{0.1945} & 0.1766 \\
SST-2 & 0.5270 & 0.4917 & \underline{0.5209} & \textbf{0.5350} & \textbf{0.5315} & \textbf{0.5295} & \textbf{0.5317} & \textbf{0.5291} \\
\bottomrule
\end{tabular}
\caption{The mean detection rate of \texttt{meta-llama/Llama-2-7b-hf} when the inspection percentage is 1-20\%. Higher is better.}
\label{tab:mean_detection_rate_wld_llama_20}
\end{table}

\begin{table}[H]
\centering
\scriptsize
\begin{tabular}{lcccccccc}
\toprule
Dataset & FreeShap & Rank 1\% & 5\% & 10\% & 15\% & 20\% & 25\% & 30\% \\
\midrule
CIFAR-10 & 0.6255 & \underline{0.6158} & \textbf{0.6694} & \textbf{0.6742} & \textbf{0.6734} & \textbf{0.6706} & \textbf{0.6627} & \textbf{0.6614} \\
\bottomrule
\end{tabular}
\caption{The mean detection rate of \texttt{resnet-18-pretrained} when the inspection percentage is 1-20\%. Higher is better.}
\label{tab:mean_detection_rate_wld_resnet_20}
\end{table}

\clearpage
\section{TMC convergence ratios for all cases}\label{app:convergence_ratio}

\begin{figure}[!ht]
  \centering      \includegraphics[width=1\textwidth]{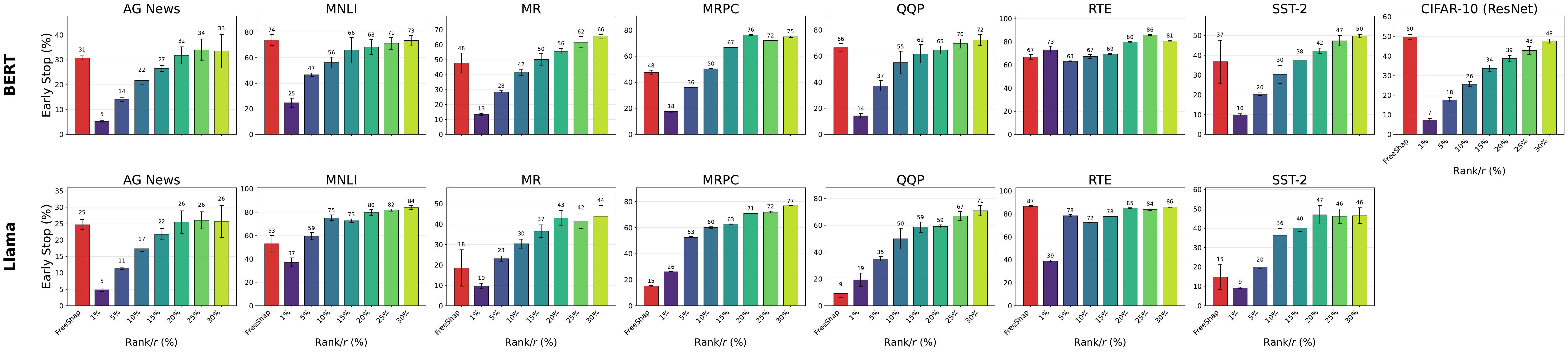}
  \caption{TMC convergence ratios for all datasets and models.}  \label{fig:early_stop_all}
\end{figure}

\end{document}